\documentclass[english,14pt]{article}
\usepackage[T1]{fontenc}
\usepackage{lmodern}   
\usepackage[utf8]{inputenc}
\usepackage{color}
\usepackage{textcomp}
\usepackage{amsmath}
\usepackage{amsthm}
\usepackage{amssymb}
\usepackage{graphicx}
\usepackage{wasysym}
\usepackage{enumitem}
\usepackage{amsfonts}
\usepackage{amsmath}
\usepackage{amssymb}

\usepackage[english]{babel}

\newcommand{\cS}{\mathcal{S}}

\newcommand{\bbR}{\mathbb{R}}

\newcommand{\transpose}{\mathsf{T}}

\usepackage{hyperref}       % hyperlinks
\usepackage{url}            % simple URL typesetting
\usepackage{booktabs}       % professional-quality tables
\usepackage{amsfonts}       % blackboard math symbols
\usepackage{nicefrac}       % compact symbols for 1/2, etc.
\usepackage{microtype}      % microtypography
\usepackage{xcolor}         % colors
\usepackage{ulem}
\usepackage{xcolor,cancel}
 
\usepackage{graphicx}
\usepackage{color}
\usepackage{physics}
\usepackage{cancel}

\usepackage{bm}

\author{Inbar Seroussi\footnote{Department of Mathematics, Weizmann Institute of Science, Rehovot, Israel}\;\;\;\;\;\;\;
	Gadi Naveh\footnote{
		Hebrew University, Racah Institute of Physics, Jerusalem, 9190401, Israel}\;\;\;\;\;\;\;
	Zohar Ringel\footnote{
		Hebrew University, Racah Institute of Physics, Jerusalem, 9190401, Israel}}
\date{}
\begin{document}
\title{Separation of Scales and a Thermodynamic Description of Feature Learning in Some CNNs}  
\maketitle
\begin{abstract}
Deep neural networks (DNNs) are powerful tools for compressing and distilling information. Their scale and complexity, often involving billions of inter-dependent  parameters, render direct microscopic analysis difficult. Under such circumstances, a common strategy is to identify slow variables that average the erratic behavior of the fast microscopic variables. Here, we identify a similar separation of scales occurring in fully trained finitely over-parameterized deep convolutional neural networks (CNNs) and fully connected networks (FCNs). Specifically, we show that DNN layers couple only through the second moment (kernels) of their activations and pre-activations. Moreover, the latter fluctuates in a nearly Gaussian manner. For infinite width DNNs, these kernels are inert, while for finite ones they adapt to the data and yield a tractable \textit{data-aware Gaussian Process}. The resulting thermodynamic theory of deep learning yields accurate predictions in various settings. In addition, it provides new ways of analyzing and understanding DNNs in general.   
\end{abstract}
\section{Introduction}
Identifying slow or relevant variables is an essential step in analyzing large-scale non-linear systems. In the context of deep neural networks (DNNs), these should be some combinations of the individual weights that are weakly fluctuating and obey a closed set of equations. One potential set of such variables is the DNNs' outputs themselves. Indeed, in the limit of infinitely over-parameterized DNNs these provide an elegant picture of deep learning \cite{Jacot2018, Matthews2018, naveh2021predicting} based on a mapping to Gaussian Processes (GPs). However, these GP limits miss out on several qualitative aspects, such as feature learning \cite{Li2015Rep,chizat2019lazy} and the fact that real-world DNNs are not nearly as over-parameterized as required for the GP description to hold \cite{Jacot2018, lee2019wide, Zhu2019, naveh2021predicting}. Obtaining a useful set of slow variables for describing deep learning at finite over-parameterization is thus an important open problem in the field.   

Several works provide guidelines for this search. Noting that GP limits can have surprisingly good performance \cite{Arora2019} and that over-parameterization is natural to deep learning \cite{Neyshabur2018, Zhang2016} we are inclined to keep some elements of the GP picture. One such element is to work in function space and study pre-activation and outputs instead of weights whose posterior distribution becomes complicated even in the GP limit \cite{Foong2020,coker2022wide}. Another element is the layer-wise composition of hidden layer kernels \cite{zavatone2021asymptotics}, using which one generates the output kernel of the GP \cite{Cho}. Such a layer-wise picture is also harmonious with the idea that DNN layers should {\it not} correlate strongly, to prevent co-adaptation \cite{Hinton2012improving}. Recently, it was shown that in some limited settings, making the GP kernel "dynamical" or flexible, so that it adapts to the dataset, can account for differences between infinite and finite DNNs \cite{LiSompolinsky2021, naveh2021predicting, aitchison2019bigger, fort2020deep,aitchison2021deepProcess}. Still, the task of finding an explicit and general set of equations describing this flexibility in DNNs remains unsolved.

In this work, we identify such slow variables and use these to derive an effective theory for deep learning capable of capturing various finite channel/width ($C/N$) effects in convolutional neural networks (CNNs) and fully connected neural networks (FCNs) such as feature learning. We argue that:  
\begin{enumerate}
\item 
For, $C,N \gg 1$ the erratic behavior of specific channels/neurons averages out and hidden layers coupled to each other only through two ``slow" variables per layer: The second moment of the pre-activations (\textit{pre-kernel}), $K^{(l)}$, and the second moment of activations ({\it post-kernel}), $Q^{(l)}$, of the $l$th layer. Furthermore, for mean square error (MSE) loss, FCNs in the so-called \textit{mean-field (MF) scaling} \cite{MeiMeanField2018} (where the last layer weights are scaled down) or CNNs with a large read-out layer fan-in behave effectively as a GP with a data-aware kernel determined by the second moment of pre-activations in the penultimate layer.
\item 
In settings where the kernels have a large density of dominant eigenvalues, the posterior (or trained) pre-activations fluctuate in a nearly Gaussian manner. Following this we use a multivariate Gaussian variational approximation for the posterior pre-activations ({\it pre-kernels}) and derive explicit equations (equations of state), for the covariance matrices governing these pre-activations. 

\item 
We identify an emergent feature learning scale (FLS) denoted by $\chi$, proportional to the train MSE times $n^2$ over $C$ (or $N$). This scale controls the difference between the finite $C,N$ output kernel ($Q_f$) and its $C,N \rightarrow \infty$ limit and in this sense reflects feature learning. Due to the $n^2$ factor, $\chi$ can be $O(1)$ or larger even for $C\gg 1$, e.g. for CNN architectures (see Fig. \ref{Fig:FCN_numerics} panel c). The same holds, with $C$ replaced by $N$, for FCNs in the \textit{MF scaling} \cite{MeiMeanField2018}. Unlike perturbation theory \cite{yaida2020non,naveh2021predicting, lee2019wide,naveh2021self}, our theory tracks all orders of $\chi$ and treats only $1/C,1/N$ perturbatively. The separation of scales between $\chi$ and $1/C,1/N$ is thus central to our analysis. Its manifestation is the fact that feature learning shifts and stretches the dynamical variables in the theory (the pre-activations) in a considerable manner yet barely spoils their Gaussianity.   
\end {enumerate}
The predictions of our approach are tested on several toy and real-world examples using direct analytical approaches and numerical solutions to the equations of state. Our analysis takes a physics viewpoint on this complex non-linear problem. Rigorous mathematical proofs are left as an open problem for future research. 

We note that there are several works showing evidence that the spectrum of the empirical weight correlation matrix show various tail effects and spikes \cite{martin2021implicit}. While in deeper layers we focus on pre-activations, the spectrum of input layer weights we obtained, is Gaussian but not independent as in Ref. \cite{coker2022wide}. Hence it can produce a variety of spectral distributions for the covariance matrix, similar to the aforementioned ones. We note a recent interesting work\cite{fischer2022decomposing} arguing that the test-loss depends only on the mean and variance of hidden activations. There, however, the setting is of a fixed trained DNN and the statistics are over the input measure rather than over the DNN parameters as in our case. 
While quantitatively different, our approach is similar in spirit to the phenomenological layer-wise Gaussian Processes put forward in a recent work \cite{aitchison2021deepProcess}. Additional approaches for finite-width include perturbative correction around the infinite width limit to leading \cite{yaida2020non,naveh2021predicting}  or higher orders \cite{Dyer2020Asymptotics, huang2020dynamics,roberts2021principles}. 
There is however mounting evidence from bounds on GP limits \cite{huang2020dynamics,Jacot2018}, numerical experiments \cite{lee2019wide,naveh2021predicting,naveh2021self,chizat2018lazy}, as well as the current work, that such perturbative expansions have slow convergence in practical regimes. 
In contrast our EoS are useful both numerically (see Sec. \ref{sec:Methods} and \ref{Sec:3FCNs}) and analytically (see Sec. \ref{subsec:CNN_2layer_solution}) and in addition allow us to model pre-activation distributions in the wild via our pre-kernels (see Sec. \ref{Sec:realworld}).

\subsection{Problem Statement}
Our general setup consists of DNNs trained on a labeled training set of size $n$, $\mathcal{D}_n=\{(\bm{x}_{\mu},y_{\mu})\}_{\mu=1}^n=\{X_n,\bm{y}\}$ with MSE loss. The input vector is $\bm{x}_{\mu}\in\mathbb{R}^d$, and $y_{\mu}$ is a scalar output. We denote vectors and tensors by boldface and use $\mu$ and $\nu$ to represent data point indices. Our training algorithm is Gradient Descent with a vanishing learning rate, along with weight decay, and additive white noise with variance $\sigma^2$ (GD+Noise \cite{naveh2021predicting}). When relaxed to its equilibrium, the resulting dynamics could also be understood in terms of Bayesian Neural Networks \cite{Welling2011} where one aims to find the minimum mean square error estimator under a Gaussian prior for the DNN weights, where each sample is corrupted by Gaussian noise with variance $\sigma^2$. 

Our theory can be applied to any finite number of convolutional, dense, or pooling layers. To illustrate its main aspects, let us focus on an $L$-layer fully connected model with width $N_l$ ($l \in [1..L-1]$),
\begin{align}
\label{Eq:3layerFCN}
f(\bm{x}) &= \sum_{j=1}^{N_L-1} w^{(L)}_{j} \phi\left(h^{(L-1)}_{j}(\bm{x}) \right)\\ \nonumber
h^{(l)}_{j}(\bm{x}) &= \sum_{i=1}^{N_{l-1}} W^{(l)}_{ji}\phi\left(h^{(l-1)}_{i}(\bm{x}) \right)\\ \nonumber
h^{(1)}_{i}(\bm{x}) &= \bm{w}^{(1)}_{i} \cdot \bm{x}
\end{align}
where $\bm{w}^{(1)}_i\in\mathbb{R}^{d}$ ($i\in[1,N_1]$), $\bm{w}^{(L)}\in\mathbb{R}^{N_{L-1}}$, $W^{(l)}\in\mathbb{R}^{N_l \times N_{l-1}}$ are the weights of the network, and the input vector $\bm{x}\in\mathbb{R}^d$ such that $d=N_0$. The activation function, $\phi:\mathbb{R}\rightarrow \mathbb{R}$, is applied element-wise. Below, we focus on error function activation, as its anti-symmetry simplifies our theory. Generalizations to ReLU, where one needs to track both kernels and averages, could be found in Supp. Mat. (5).

The main object we analyze is the equilibrium distribution of the GD+Noise algorithm in function space. Adopting physics notation, this distribution can be written as $p(\bm{f}|\mathcal{D}_{n})=e^{-\cS}/\mathcal{Z}(\mathcal{D}_{n})$, where $\mathcal{Z}(\mathcal{D}_{n})=\int e^{-\cS}$ is the partition function, and $\cS$ is the action or negative log-posterior (see also Methods). Taking a Bayesian perspective, this probability distribution can also be viewed as a posterior distribution given measurements of $y_{\mu}$ having Gaussian noise with variance $\sigma^2$ and a prior given by a finite-width random DNN. As shown in Supp. Mat. (1) our partition function is governed by the following action 
\begin{multline} 
\label{eq:S_nonLin}
\mathcal{S}=\frac{1}{2}\sum_{ i=1}^{N_1}(\bm{h}_{i}^{(1)})^\transpose\left[Q^{(1)}\right]^{-1}\bm{h}_{i}^{(1)}
+\frac{1}{2}\sum^{L-1}_{l=2}\sum_{ i=1}^{N_l}(\bm{h}_{i}^{(l)})^\transpose\left[\tilde{Q}^{(l)}(\bm{h^{(l-1)}})\right]^{-1}\bm{h}_{i}^{(l)}
\\+\frac{1}{2}\bm{f}^\transpose\left[\tilde{Q}_{f}(\bm{h}^{(L-1)})\right]^{-1}\bm{f}+\frac{1}{2\sigma^{2}}\sum_{\mu}(f_{\mu}-y_{\mu})^{2}
\end{multline}
where $f_\mu = f(\bm{x}_{\mu})$, and
\begin{align}\label{eq:sample free kernels}
\tilde{Q}_{f}(\bm{h}^{(L-1)})_{\mu\nu}=&\frac{\sigma_{L}^{2}}{N_{L-1}}\sum_{j=1}^{N_{L-1}}\phi(h_{j\mu}^{(L-1)})\phi(h_{j\nu}^{(L-1)})\\ \nonumber
\tilde{Q}^{(l+1)}(\bm{h}^{(l)})_{\mu\nu}&=\frac{\sigma_{l+1}^{2}}{N_{l}}\sum_{i=1}^{N_{l}}\phi(h_{i\mu}^{(l)})\phi(h_{i\nu}^{(l)})\\ \nonumber
Q_{\mu\nu}^{(1)}=\frac{\sigma_{1}^{2}}{d}\boldsymbol{x}_{\mu}^{T}\boldsymbol{x}_{\nu}.
\end{align}
where $\bm{h}^{(l)}_i\in\mathbb{R}^n$ and $\sigma_{l}^2$ determine the weight variances at $n=0$ (in the GD+Noise context, these are determined by the strength of the weight decay \cite{naveh2021predicting}). We comment that for rank-deficient matrices, the inverses are regularized by including a small positive regularizer to be taken to zero at the end of the computation. 

To familiarize ourselves with the action in Eq. (\ref{eq:S_nonLin}) let us see how it reproduces the standard Gaussian Processes picture at infinite channel/width \cite{naveh2021predicting, Cho}. 
Strictly speaking, this action is highly non-linear, since the $\tilde{Q}$'s matrix elements contain high powers of pre-activations and since their inverse enters the action. 
Crucially, however, the $\tilde{Q}$'s are width-averaged quantities. Thus, at $N_l \rightarrow \infty$ one may replace them by their averages.  
Furthermore, upstream dependencies, wherein $\bm{h}^{(l)}$ affects $\bm{h}^{(l-1)}$, vanish (see Supp. Mat. (1)). Roughly speaking, this is because $\bm{h}^{(l)}$ only feels the collective effect of all the neurons in $\bm{h}^{(l-1)}$ rendering its feedback on any specific neuron negligible. 

Having these two simplifications in mind, we begin a layer-by-layer downstream analysis of the DNN: As there is no upstream feedback on $\bm{h}^{(1)}$, the average of $\tilde{Q}^{(2)}(\bm{h}^{(1)})$ (denoted $Q^{(2)}$) can be carried under the Gaussian action of the input layer alone (first term in the action). Replacing $\tilde{Q}^{(2)}(\bm{h}^{(1)})$ by $Q^{(2)}$ in the second term in the action, would then imply that  $\bm{h}^{(2)}$ fluctuates in a Gaussian manner with $Q^{(2)}$ as its covariance matrix. 
Next, the average of $\tilde{Q}^{(3)}(\bm{h}^{(2)})$ (denoted $Q^{(3)}$) can be found based on the now known, Gaussian statistics of $\bm{h}^{(2)}$. Repeating this process, the final kernel ($Q^{(L)}=Q_f$) is found and is exactly the one obtained using the method introduced by Cho and Saul \cite{Cho}. Together with the MSE loss term (last term in the action), we find that the outputs ($f_{\mu}$) fluctuate in a Gaussian manner leading to the standard GP picture of infinite width trained DNNs \cite{naveh2021predicting}.  

Here, however, our focus is at large but finite width ($N_l \gg 1$). In this more complex regime, several corrections may appear:
{\bf (i)} The pre-activations' average and covariance may deviate from those of a random DNN. 
{\bf (ii)} $Q^{(l)}$, the covariance of activations in the $l-1$ layer, would not solely determine the covariance of pre-activations on the downstream layer $l$, as upstream effects between $\bm{h}^{(l+1)}$ and $\bm{h}^{(l)}$ come into play.  
{\bf (iii)} Inter-channel (or inter-neuron in the fully-connected case) and inter-layer correlations may appear. 
{\bf (iv)} The fluctuations of pre-activations may deviate from that of a Gaussian. 
A priori, all these corrections may play similarly dominant roles, thereby making analysis cumbersome.

\section{Results}

\subsection{Effective GP Description in the Feature Learning Regime}
The basic analytical insight underlying this work is that these four types of corrections scale differently with $n$, $d$, and $N_l$. This allows for a controlled mean-field treatment, which differs substantially from straightforward perturbation theory in one over the width. As shown in Supp. Mat. (1.6), corrections of type (iii) are often much smaller than those of type (i) and (ii). This holds generally for hidden layers when $N_l$'s (or channel's number for CNNs) are much larger than $1$. Considering the output layer of CNNs this requires a large fan-in and FCNs this holds when using an MF scaling \cite{MeiMeanField2018}. Turning to correction of type (iv) in the $l$'th layer, these are suppressed when the average $\tilde{Q}^{(l)}$ has a large density of dominant eigenvalues --- a situation relevant for when $n$ and the input dimension are both large relative to one. 

This leaves us with corrections of types (i) and (ii). Interestingly, following these corrections to all orders leads to a tractable mean-field picture of learning. The latter is an augmentation of the standard correspondence between GPs and DNNs at infinite width (NNGP)\cite{Cho, lee2017deep}: Pre-activations in different layers or channels/neurons remain uncorrelated and Gaussian. Correlations only appear between different data-points (and latent pixels for CNNs) within the same layer and channel/neuron. We henceforth denote the covariance of pre-activations and activations at layer $l$ (up to normalization by the variance of the weights) by $K^{(l)}$ and $Q^{(l)}$ and refer to these as {\it pre-kernel} and {\it post-kernel}, respectively. However in the NNGP \cite{lee2017deep} viewpoint, $Q^{(l)}$ is simply proportional to $K^{(l)}$ and fully determined by the upstream kernel ($Q^{(l-1)}$) whereas here $K^{(l)}$ and $Q^{(l)}$ differ and moreover depend both on the upstream and downstream kernels. 

Specifically, for the above $L$-layers FCN with $\phi=\mathrm{erf}$ activation function, we obtain (see also Fig. \ref{fig:Abstract} and Supp. Mat. (1) for derivation and extension to CNNs and any number of layers)
\begin{align}\label{Eq:3LayerEOS}
\bar{\bm{f}} &= 
Q_{f}[Q_{f} + \sigma^{2}I_{n}]^{-1}\bm{y}
\\\nonumber
[(K^{(L-1)})^{-1}]_{\mu \nu } &= [(Q^{(L-1)})^{-1}]_{\mu\nu} - \frac{1}{N_{L-1}}\Tr{A^{(L)} \frac{\partial Q_f}{\partial [K^{(L-1)}]_{\mu \nu}}}  
\\\nonumber
[[K^{(l-1)}]^{-1}]_{\mu\nu} &=[[Q^{(l-1)}]^{-1}]_{\mu\nu}+\frac{2N_l}{N_{l-1}}\frac{\partial D_{\mathrm{KL}}(K^{(l)}||Q^{(l)})\,\,}{\partial [K^{(l-1)}]_{\mu\nu}} \,\,\, \text{for all $l\in[2,L-1]$}  \\ \nonumber %\label{eq:K_{l-1}}
[\Sigma^{-1}]_{ss'} & =\frac{d}{\sigma_{1}^{2}}\delta_{ss'}+\frac{2N_{2}}{N_{1}}\frac{\partial D_{\mathrm{KL}}(K^{(2)}||Q^{(2)})\,\,}{\partial\Sigma_{ss'}}  \\ \nonumber
A^{(L)} &= \sigma^{-4} (\bm{y}-\bar{\bm{f}})(\bm{y}-\bar{\bm{f}})^{\transpose}- \left[Q_{f}+\sigma^{2}I_{n}\right]^{-1} 
\end{align}
where $[Q_f]_{\mu \nu} =\sigma_{L}^2 G(K^{(L-1)})_{\mu \nu}$, $ [Q^{(l)}]_{\mu \nu } = \sigma_l^2  G(K^{(l-1)})_{\mu \nu }$ and  
$G(K)_{\mu \nu}= \frac{2}{\pi} \sin^{-1}\left(\frac{2 K_{\mu\nu}}{\sqrt{1+2K_{\mu \nu}}\sqrt{1+2K_{\mu \nu}}} \right)$ \cite{williams1996computing} for a matrix $K \in \mathbb{R}^{n \times n}$ (equivalent expressions are known for several common activation functions, such as ReLU \cite{Cho}).
Also, $\bar{\bm{f}}$ is the average DNN output and $D_{\mathrm{KL}}(K || Q)$ is the KL-divergence between two Gaussians with covariance matrices $K$ and $Q$. The input layer post-kernel is $K^{(1)} = X_n \Sigma X_n^{\transpose}$, where $\Sigma$ is the covariance matrix of input layer weights.

As their lack of dependence on width suggests, the first equation together with the definitions of the \textit{post-kernels} $Q^{(l)},Q_f$ are already present in the strict GP limit ($N_l \rightarrow \infty$). They are, respectively, the GP inference formula and standard kernel recursive equations of random DNNs \cite{Cho} with $\erf$ activation. 
The remaining equations are, to the best of our knowledge, novel and follow the changes to the {\it pre-kernels} and  {\it post-kernels} at finite $N_l$. These could be solved analytically in some simple cases (see subsection \ref{subsec:CNN_2layer_solution} for the case of two-layer CNN). We note that for non-anti-symmetric activation, one will also need to track the mean of each layer's pre-activation (see Supp. Mat. (5)).

To get a qualitative impression of their role, one can consider the case where the penultimate layer ($l = L-1$) is linear, in which case $Q_f = \sigma_L^2 K^{(L-1)}$. Consequently, $\frac{\partial [Q_f]_{\mu',\nu'}}{\partial [K^{(L-1)}]_{\mu,\nu}} =\sigma_{L}^2 \delta_{\mu \mu'} \delta_{\nu \nu'}$, (where $\delta_{\mu\nu}$, with double index refers here to the Kronecker delta) and thus the second equation simplifies to 
\begin{align}
\label{Eq:Qf_LinPen}
\sigma_L^2 Q_f^{-1} &= (Q^{(L-1)})^{-1} - \frac{\sigma_L^2}{N_{L-1}}\left(\bm{\delta}\bm{\delta}^\transpose -[Q_f+\sigma^2I_n]^{-1}\right),
\end{align}
where $\bm{\delta}=(\bm{y}-\bar{\bm{f}})/\sigma^2$. We note in passing that even for a non-linear penultimate layer, a similar term will arise from the expansion of $Q_{f}$ in $K^{(L-1)}$ to linear order.  From the above form, several insights can be drawn.

First, we argue that the above equation implies that the trained DNN is more susceptible to changes along $\bm{\delta}$ than the DNN at $N_{L-1} \rightarrow \infty$. Noting how $Q_f^{-1}$ enters the action (Eq. \ref{eq:S_nonLin}), it controls the stiffness associated with fluctuations in ${\bm f}$. Hence $Q_f^{-1}$ makes fluctuations in the direction of $\bm{\delta}$ more likely than they are according to $(Q^{(L-1)})^{-1}$. Since $\bm{\delta}$ measures the discrepancy in train predictions, this effect reduces the discrepancy by making the DNN more responsive in these directions than it is at $N_{L-1}\rightarrow \infty$. 
The second term, proportional to $\left[Q_{f}+\sigma^{2}I_{n}\right]^{-1}$ amounts to a negligible reduction in fluctuations along eigenvectors of $Q_f$ corresponding to eigenvalues which are larger than $\sigma^2$. 

Using Eq. (\ref{Eq:Qf_LinPen}) one can also identify the aforementioned emergent feature learning scale (or FLS) namely, $\chi= N_{L-1}^{-1}\bm{\delta}^{\transpose} Q^{(L-1)} \bm{\delta}$. This scale represents the magnitude of the leading term when one Taylor expands $Q_f$ in $1/N_{L-1}$. When $\chi=O(1)$ or larger there is a significant change in the eigenvalues of $Q_f$ compared to $Q^{(L-1)}$ which indicates feature learning. On the other hand, when this quantity is small, we are closer to the GP regime (see Supp. Mat. (1.6)). To asses the scaling $\chi$, one can consider the common situation where $\bm{\delta}$ has some non-negligible overlap with dominant eigenvectors of $Q^{(L-1)}$ whose eigenvalues are on the scale $\lambda$. Here we find $\chi \approx \lambda \cdot \text{MSE}/\sigma^2 \cdot n/N_{L-1}$, where $\text{MSE}$ denotes the mean train MSE which enters here via $||\bm{\delta}||^2/n$. Due to its explicit $n$ dependency, and for $\lambda=O(n)$ at large $n$ \cite{Cohen2019}  --- $\chi$ maybe $O(1)$ even at very large $N_{L-1}$ and/or when the average MSE is rather small. 

Figure \ref{Fig:FCN_numerics} panel {\bf (c)} shows the value of $N_{L-1}$ (or $C_{L-1}$) at which $\chi=1$ (i.e. $N_{L-1}$ or $C_{L-1}$ at which feature learning becomes a dominant effect) as a function of $n$ for several DNNs we study. The scale separation, demonstrated there by the fact that $\chi$ can be $O(1)$ in regions where $1/N_{l}$'s is negligible, is central to our analytical approach. 

This scale $\chi$ is also the reason that naive perturbation theory in $1/N_l$ fails at large $n$ \cite{naveh2021predicting, lee2019wide, naveh2021self, Dyer2020Asymptotics}, as it treats $\chi$ and $O(1/N_{L-1})$ on the same footing, since they both have a single negative power of $N_{L-1}$. In contrast, our EoS treat the FLS non-perturbatively. 

Last we stress that the EoS provide us with a concrete effective GP description for the entire DNN as well as its hidden layers. A priori one would expect that the normality of pre-activations, a large $C,N$ trait, will be lost at finite $C,N$. Yet we find that pre-activations remain Gaussian and accommodate strong feature learning effects while maintaining accurate predictions. This unexpectedly simple behavior opens various reverse engineering possibilities wherein one infers the effective kernels from experiments and uses their spectrum and eigenvectors to rationalize about the DNN (see also Fig. \ref{Fig:EOS3}). 

\subsection{Numerical Demonstration: 3 Layer FCN}
\label{Sec:3FCNs}
\begin{figure}[h!]
\vspace*{-0.1in}
\begin{center}
\includegraphics[height=3.05cm,trim={2cm 22.5cm 1.2cm  0cm},clip]{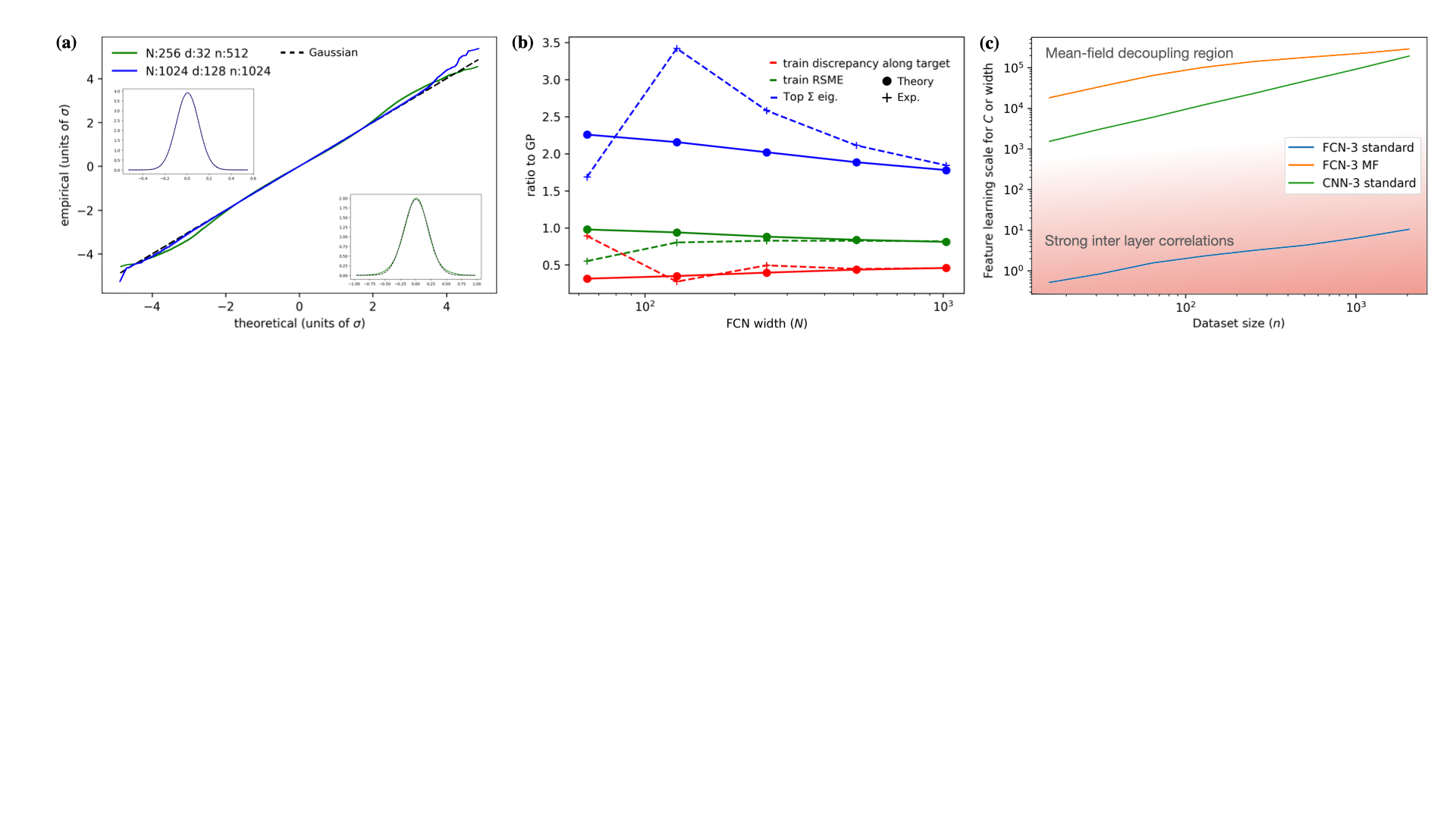}% height=4.28cm, scale = 0.18,
\end{center}
\caption{{\bf Theory versus experiment.} 
Panels {\bf (a)} and {\bf (b)} compare fully trained 3-layer FCNs  in the MF scaling with our theoretical predictions. Panel {\bf (a)} studies the most strongly fluctuating weight mode in the first layer. A QQ-plot is shown, of its empirical distribution against a normal one, along with histograms together with Gaussian fits (insets). As the scale of the problem increases, these fluctuations become more and more Gaussian. 
Panel {\bf (b)} shows the experimental values and EoS predictions each normalized by their values at $N_l \rightarrow \infty$ with fixed $\sigma_l$ (i.e. the GP value). Here we used $n=1024$ and $d=128$.
Panel {\bf (c)} plots the $N$ (or channel number $C$) at which the feature learning scale ($\chi$) is $1$. This dictates the crossover between weak and strong feature learning in our theory. The color gradient provides a qualitative separation between large and small $N,C$ and thus correlates with the adequateness of our mean-field decoupling for the hidden layers. Evidently, FCNs with standard scaling ($[Q_{f}]_{\mu \mu}=O(1)$ at $n=0$) require $N_l=O(1)$ for appreciable feature learning. However, FCN with an MF scaling, as well as CNNs in standard scaling, can exhibit strong feature learning well within the regime of our mean-field decoupling. 
\label{Fig:FCN_numerics}}
\end{figure}

Next, we test the agreement between the above results and statements and actual trained DNNs, starting from the 3-layer FCN defined in Eq. (\ref{Eq:3layerFCN}) with $L=3$. We focus here on a student-teacher setting with $n=512$ or $1024$ training data points drawn from iid Gaussian distributions with unit variance along each input dimension. The target was generated by a randomly drawn teacher FCN of the same type only with $N_1=N_2=1$. The student was trained using an analog scaling to the MF scaling \cite{MeiMeanField2018}, wherein the output layer weights are scaled down by a factor of $1/\sqrt{N_l}$. Whereas for the CNNs discussed below, this choice of scaling was not required, for FCNs we found it necessary for getting any appreciable feature learning at $N_1=N_2=N \gg 1$ (see Fig. \ref{Fig:FCN_numerics} panel {\bf (c)}).

As described in the methodology section \ref{sec:Methods}, we trained $20$ FCNs using the GD+Noise algorithm until they reached equilibrium. We use these trained FCNs to calculate various average quantities under our partition function (Eq. (\ref{eq:S_nonLin})). Specifically, we focused on: 
{\bf (i)} The normalized train-loss on the scale of $\sigma^2$, namely $\text{MSE}/(\sigma^2\sum_{\mu} y^2_{\mu})$ 
{\bf (ii)} The eigenvalues ($\lambda_{i}, i \in [1..d]$) of the average $\Sigma$, where we average over neurons, training seeds, and training time (the latter within the equilibrium region). 
{\bf (iii)} The normalized overlap ($\alpha$) between the discrepancy in prediction on the training set times the target, namely $\alpha = \sigma^{-2} \sum_{\mu=1}^n (\bar{f}_{\mu}-y_{\mu}) y_{\mu}/(\sum_{\mu} y^2_{\mu})$. We then used a JAX-based \cite{jax2018github} numerical solver for the EoS and compared it with the experiment. 

As the results of Fig. \ref{Fig:FCN_numerics} panel {\bf (b)} show, the predictions of our EoS for all these three quantities converged well as we increased $N$. Furthermore, they do so in a region where they differ considerably from their associated GP limit. Indeed, as shown in the Supp. Mat. (3) the top $\Sigma$ eigenvalue came out 2-3 times larger than it is in the GP limit. The associated eigenvector corresponded to the first layer weights of the teacher ($\bm{w}^*$). The rest of the eigenvalues remained at their GP limit values. Put together, this is a clear sign of strong feature learning. 

Notably, however, this notion of feature learning does not involve compression. Indeed, since $\Sigma$ has the same variance as in the GP limit for directions perpendicular to $\bm{w}^*$, it does not compress the input by projecting it solely on the label relevant direction ($\bm{w}^*$). Instead, it exaggerates the fluctuation of student weights along, $\bm{w}^*$ thereby making it statistically more likely that $h^{(1)}_{c \mu}$ and $h^{(1)}_{c \nu}$ with opposite sign of $\bm{w}^* \cdot \bm{x}_{\mu}$ and $\bm{w}^* \cdot \bm{x}_{\nu}$ will be further apart in the space of pre-activations. %Notwithstanding, knowing $\Sigma$ one can now compress the data by projecting it on $\Sigma$'s leading eigenvectors (\cite{Brend}) however this is not what the DNN is doing. 

Next, we study how $\chi$ behaves as a function of $n$ and $C$ (or $N$) for different architectures. Fig. \ref{Fig:FCN_numerics} shows the value of $N$ (or $C$) at which $\chi=1$. As $\chi$ contains a single inverse power of $C$ at ten times this value, $\chi$ would be $0.1$ and thus indicate only minor feature learning effects in our EoS. As $N,C$ diminish from this latter value, our EoS yield increasingly stronger feature learning effects. We find that both for CNNs in the standard scaling and for FCNs with MF scaling, the crossover to feature learning happens well within the validity region of our mean-field decoupling (i.e. large $N$ or $C$). In contrast, FCN with standard scaling shows this crossover when $N=O(1)$, which is outside the scope of our theory. In this aspect, we comment that there is evidence that FCNs with standard scale are inferior to those with mean-field scaling \cite{Yu2020} and perform similarly to GPs \cite{lee2017deep}.

\subsection{Analytical Solution of the EoS - Two Layer CNN \label{subsec:CNN_2layer_solution}} 
Having tested our EoS numerically, we turn to show they lend themselves, in simple settings, to a fully analytical calculation. Amongst other things, this will flesh out the non-perturbative nature of our results. To this end, we consider a simple non-linear CNN with 2 layers. Though bounds have been derived \cite{brutzkus2021optimization, brutzkus2017globally}, we are not aware of any analytical predictions for the performance of finite non-linear 2-layer DNNs, let alone CNNs. It is therefore a natural first application of our approach. Specifically, we consider 
\begin{align}
    f(\bm{x}) &= \sum_{i=1}^N \sum_{c=1}^{C} a_{ic} \erf\left(\bm{w}_c \cdot \bm{x}_{i}\right) 
\end{align}
where $\bm{x} \in \bbR^{d}$ with $d = NS$ and $\bm{w}_c, \bm{x}_{i} \in \bbR^{S}$. The vector $\bm{x}_{i}$ is given by the $iS,..,(i+1)S-1$ coordinates of $\bm{x}$. 
The dataset consists of $\{ \bm{x}_{\mu} \}_{\mu=1}^{n}$ i.i.d. samples, each sample $\bm{x}_\mu$ is a centered Gaussian vector with covariance $I_d$. We choose a linear target of the form $y_{\mu} = \sum_{i} a^*_i (\bm{w}^* \cdot \bm{x}_{\mu,i})$ where $a^*_i \sim {\mathcal N}(0,1/N)$ and $w^*_s \sim {\mathcal N}(0,1/S)$. This choice is not crucial but does simplify considerably the GP inference part of the computation. We train this DNN using GD+noise and tune weight-decay and gradient noise such that, without any data, $a_{ic} \sim {\mathcal N}(0,\sigma_\text{a}^{2}(NC)^{-1})$ and $[\bm{w}_c]_s \sim {\mathcal N}(0,\sigma_\text{w}^2 S^{-1})$.

\begin{figure}[h!]
\vspace*{-0.1in}
\begin{center}
\includegraphics[height=4cm,trim={0cm 15.8cm 4cm 3cm},clip]{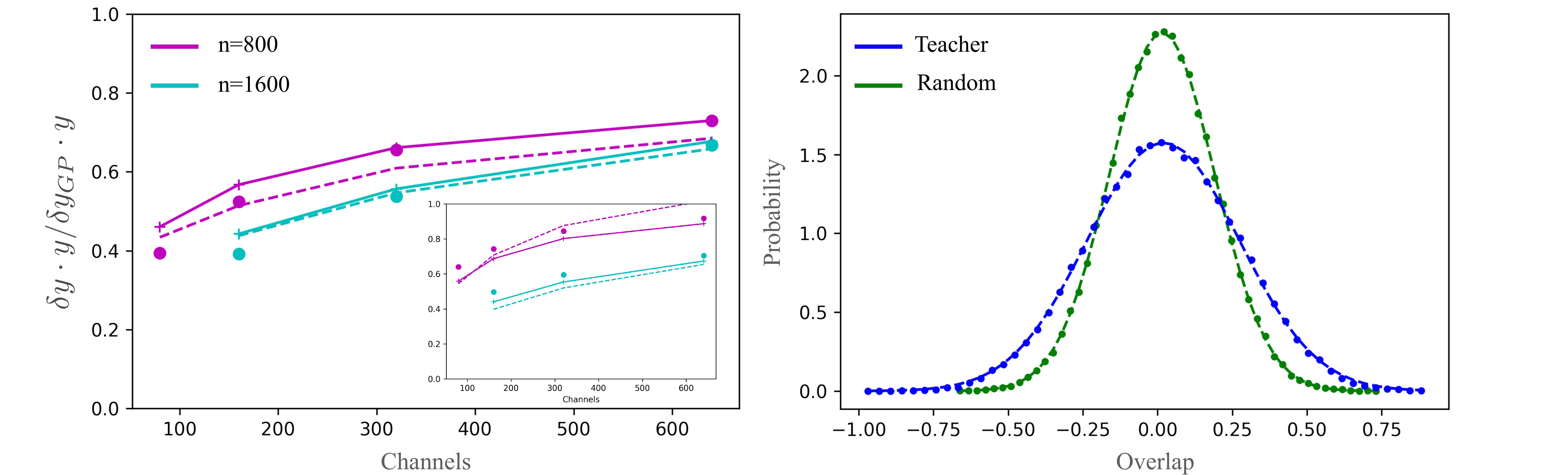}
\end{center}
\caption{{\bf Theory and experiment comparison for the 2-layer non-linear CNN.} 
{\it Left.} Discrepancy measured along target direction normalized by that of the corresponding GP ($\alpha/\alpha_{C\rightarrow \infty}$). Dots denote empirical values for the test-set, and dashed and solid lines are theoretical predictions via an approximate-analytical and exact-numerical solution of the equations of state, respectively. Train-set comparisons are shown in the inset. 
{\it Right.} Statistics of $\bm{w} \cdot \bm{w}^*/|\bm{w}^*|$ of the trained CNN (blue) compared with $\bm{w}$ projected on a normalized random vector (green) for the $n=1600$ experiment with $C=640$. Dashed lines show fit to a Gaussian. 
\label{Fig:2Layer}}
\end{figure}

The equations of state are given by (See Supp. Mat. (3))
\begin{align}
\label{eq:Qf2Layer}
\bar{\bm{f}}&= 
 Q_{f}[\sigma^{2}I_{n}+Q_{f}]^{-1}\bm{y}
\\\nonumber
[Q_f]_{\mu \nu}&= \frac{ \sigma_\text{a}^2}{N} \sum_i G(X_n\Sigma X_n^{\transpose})_{\mu i,\nu i} \\\nonumber
[\Sigma^{-1}]_{ss'} &= \frac{S}{\sigma_\text{w}^2} \delta_{ss'} - \frac{1}{C} \Tr{(\bm{\delta}\bm{\delta}^\transpose-{K_f}^{-1}) \frac{\partial Q_f}{\partial \Sigma_{ss'}})}
\end{align}
Here we denote the discrepancy from the target by $\bm{\delta}=(\bm{y}-\bar{\bm{f}})/\sigma^2$.
% K_f = Q_f + \sigma^2 I_n$.
The above equations for $\Sigma_{ss'}$ and ${\delta}_{\mu}$ could be solved numerically and compared with DNN training experiments. The results are shown in Fig. (\ref{Fig:2Layer}) in solid lines and match empirical values well.  

To obtain fully analytical results, we proceed with several approximations for large $n$. 
First, we approximate the spectrum of the matrix $[Q_f]_{\mu \nu}$ based on its continuum kernel version $Q_f(\bm{x},\bm{x}')$. This is closely related to the equivalent kernel \cite{Rasmussen2005} approximation, which we adopt here along with its leading order correction \cite{Cohen2019}. Similarly, we use large $n$ to replace the double summation $\sum_{\nu \mu} {\delta}_{\mu} {\delta}_{\nu} [Q_f]_{\mu \nu}$ by two integrals over the measure from which $\bm{x}_{\mu}$ are drawn ($d\mu$). See Supp. Mat. (3.1) for further details and a discussion of the fully connected case ($N=1$). 

The latter approximation also fleshes out the importance of the FLS ($\chi$). Technically, the two summations provide for the $n^2$ scaling and $\delta_{\mu}$ is related to the MSE via $t_{\mu} = (y_\mu-\bar{f}_\mu )/\sigma^2$ (see also Supp. Mat. (3)). 
The FLS controls the deviation between the \textit{pre-kernel} and \textit{post-kernel} of the penultimate layer with our EoS. Hence, in particular, it implies deviations from GP where these are not equal.

Following our approximations, the equations acquire the full rotation symmetry of the data-set measure, which amount to an independent orthogonal transformation of each $\bm{x}_{i}$. Furthermore, as shown in Supp. Mat. (3), at large $S$, $\bar{f}(\bm{x})$ (the continuum function representing $f_{\mu}$) is linear given a linear target, $y(\bm{x})$,  regardless of $\Sigma$ and hence so is ${\delta}(\bm{x})$.
The above symmetry then implies that ${\delta}(\bm{x})$ is only a function of $\bm{w}^* \cdot \bm{x}_i$ and furthermore takes the simple form ${\delta}(\bm{x}) = \alpha y(\bm{x})$. The quantity $\alpha$ thus measures the overlap between the discrepancy in predictions (${\bm t}$) and the target. Following this, the EoS are reduced to a non-linear equation in a single variable $\alpha$ 
\begin{align}
\label{Eq:AlphaNonLin}
    \sigma^2 \alpha &= 1 - \frac{q_{\mathrm{train}} S \lambda_{\infty} l_*}{S \lambda_{\infty} l_* + \sigma^2/n} \\ \nonumber 
    l_* &= \frac{1}{S}[1 - \chi_2]^{-1} \\ \nonumber 
    \chi_2 &= C^{-1} \alpha^2 n^2 \lambda_{\infty}
\end{align}
where $\lambda_{\infty}$ is the dominant eigenvalue of $Q_f(\bm{x}, \bm{x}')$ associated with a linear function in the limit $C\rightarrow \infty$, $l_*$ is the eigenvalue of $\Sigma$ associated with $\bm{w}^*$ (the remaining eigenvalues  are inert and equal to $1/S$), $\chi_2$ is the FLS, and we assumed $||\bm{a}^*||^2=||\bm{w}^*||^2=\sigma_\text{a}^2=\sigma_\text{w}^2=1$ (see Supp. Mat. (3). for more generic expressions). The quantity $q_{\mathrm{train}}$ is exactly one in the equivalent kernel limit, and its perturbative correction (in $1/n$) can be found in Supp. Mat. (3). 

Solving the above equation for $\alpha$, one obtains $l_*$ and hence $\Sigma$ and also $Q_f$ (via Eq. (\ref{eq:Qf2Layer})). Using the obtained $Q_f$ one can calculate the DNN's predictions on the test-set. The effect of the FLS is evident in the second equation where it controls the deviations from the GP limit. Here we also recall that $\alpha^2$ is the train MSE over $\sigma^4$, thus $\chi_2$ as defined above contains the MSE factor mentioned in the introduction.    

To test the theoretical predictions, we trained two such CNNs, with $n=\{800,1600\};S=64;N=20$ and varying channel number. Fig. \ref{Fig:2Layer}, left panel, shows the empirical test-set values for $\alpha$ (dots) compared with a numerical solution of the equations of state (solid lines) and their analytical solution (dashed lines). For the latter, we obtained $\Sigma$ analytically and performed the resulting GP inference with $Q_f$ numerically. The inset tracks the train-set results which, in this case, are fully analytical and involve no numerical GP inference. Both predictions match empirical values quite well, even in the regime where test root MSE is roughly half that of a Gaussian Process ($C \rightarrow \infty$). The right panel shows the input layer weights, dotted with $\bm{w}^*/|\bm{w}^*|$ and with a normalized random vector. These remain Gaussian up to minor statistical noise. Further details can be found in the methods section.

To emphasize the non-perturbative nature of our Eq. (\ref{Eq:AlphaNonLin}), let us assume for the sake of negation that they agree with first order perturbation theory in $1/C$ (as in Refs.  \cite{naveh2021predicting, lee2019wide, naveh2021self, yaida2020non}). If so, we may replace $\alpha$ in the above expression for $\chi_2$ by its GP value, as it  already contains one negative power of $C$ and hence receives no further corrections at that order. Numerics show this value $\alpha_{GP}=0.558$ for $n=1600$. Plugging this in, one obtains $l_*=\frac{2}{S}[1-633.2/C]^{-1}$. Clearly, this logic leads to a contradiction unless $C\gg 633.2$. In contrast, our theory provides highly accurate predictions for $n=1600,C=320$ and $C=640$ well away from where $\frac{2}{S}[1-633.2/C]^{-1}$ admits a perturbation theory in $1/C$. In Supp. Mat. (6.2) we report additional results on $l_*$ over its GP value.  

\subsection{Extensions to Deeper CNNs and Subsets of Real-World Data-sets}
\label{Sec:realworld}
For truly deep CNNs and real-world datasets, obtaining fully analytical predictions for DNN performance is a challenging task even in the $C\rightarrow \infty$ limit. Still, the EoS could be solved numerically and compared with experimental values. Furthermore, the quantities which underlie them could be examined and reasoned upon. We do so here in two richer settings, a 3-layer CNN trained with a teacher CNN and the Myrtle-5 CNN \cite{shankar2020neural} trained on a subset of CIFAR-10.

\begin{figure}[h!]
\vspace*{-0.1in}
\begin{center}
\includegraphics[
trim={0cm 3cm 0cm 2cm},clip,scale=0.4]{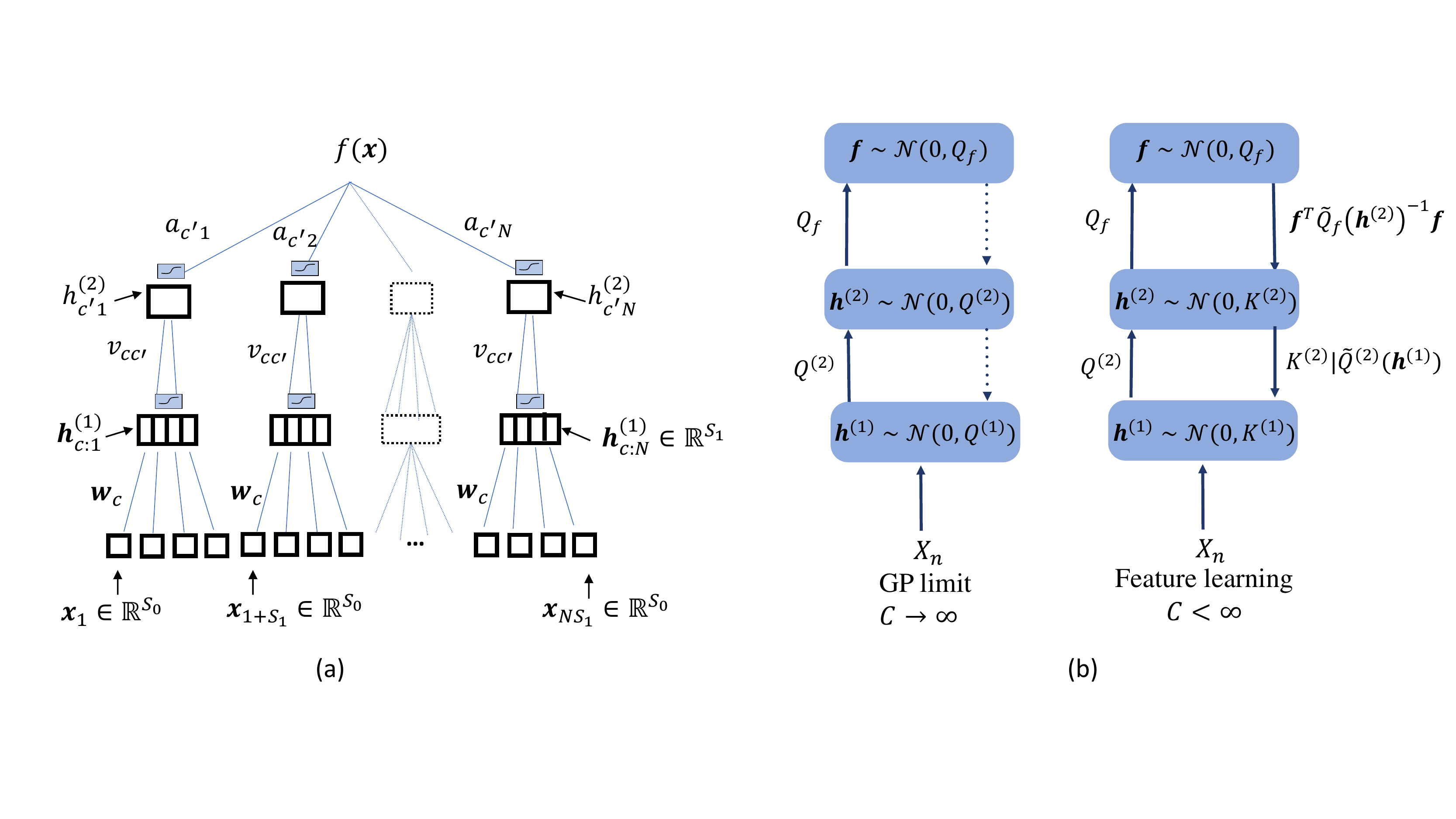}
\end{center}
% \vspace*{-0.15in}
\caption{{\bf (a)} An illustration of a 3 layers CNN architecture. 
{\bf (b)} Learning as described by the effective theory, for $C_1=C_2=C$. 
{\it Left:} for $C\rightarrow \infty$, weights of the first layer ($\bm{w}_c$), pre-activations ($\bm{h}^{(2)}$), and the output ($\bm{f}$) fluctuate according to a Gaussian distribution with \textit{post-kernels} $\sigma^2_w I_{S_0}/S_0, Q^{(2)}$, $Q_f$, respectively. 
{\it Right:} for $C \gg 1$ but finite, each layer receives an effective Gaussian action from the previous layer with kernels $Q^{(2)},Q_f$ and "back-propagation" terms which try to align $\tilde{Q}^{(2)}(\bm{h}^{(1)}), \tilde{Q}_f(\bm{h}^{(2)})$ with the second moment of $\bm{h}^{(2)}$, and $\bm{f}$ respectively. At large $n$ and input fan-in, these combine to yield an approximately Gaussian distribution for the weights and pre-activation with \textit{pre-kernels} $\Sigma, K^{(2)}$, and the DNN outputs ($\bm{f}$), which amount to a Gaussian Process Regression (GPR) with $Q_f$.     
\label{fig:Abstract}}
\end{figure}

Our first setting extends that of the previous subsection by having an extra activated layer and a non-linear target function. Specifically, we consider a student CNN defined by  
\begin{align}
\label{Eq:3layerCNN}
    f(\bm{x}) &= \sum_{j=0}^{N-1} \sum_{c'=1}^{C_2} a_{c'j} \phi\left(h^{(2)}_{c' j}(\bm{x}) \right) \\ \nonumber 
    h^{(2)}_{c' j}(\bm{x}) &= \sum_{i=0}^{S_1-1}\sum_{c=1}^{C_1} v_{c' c i}\phi\left(h^{(1)}_{c ji}(\bm{x}) \right) \\ \nonumber 
    h^{(1)}_{c ji}(\bm{x}) &= \bm{w}_{c} \cdot \bm{x}_{i+j S_1}
\end{align}
where $\bm{w}_c,\bm{x}_{i+jS_1}\in\mathbb{R}^{S_0}$, $a\in\mathbb{R}^{C_2\times N}$, $v\in\mathbb{R}^{C_2 \times C_1 \times S_1}$, and the input vector $\bm{x}\in\mathbb{R}^d$ with $d=N S_1 S_0$, and the activation function, $\phi:\mathbb{R}\rightarrow \mathbb{R}$, is applied element-wise. See Fig. \ref{fig:Abstract} for illustration. Similarly to before, the regression target ($y_{\mu}$) is generated by a random teacher CNN ($y_{\mu} = f^*(\bm{x}_{\mu})$) having the same architecture as the student, only with $C_1=C_2=1$. In addition, we chose  $S_0=50,S_1=30,N=2,C_1=C_2=100$ for the student. Further details are found in the methods section \ref{sec:Methods}.

\begin{figure}[h!]
\vspace*{-0.1in}
\begin{center}
\includegraphics[height=3.8cm,trim={3.5cm 29.18cm 43.5cm 2.87cm},clip]{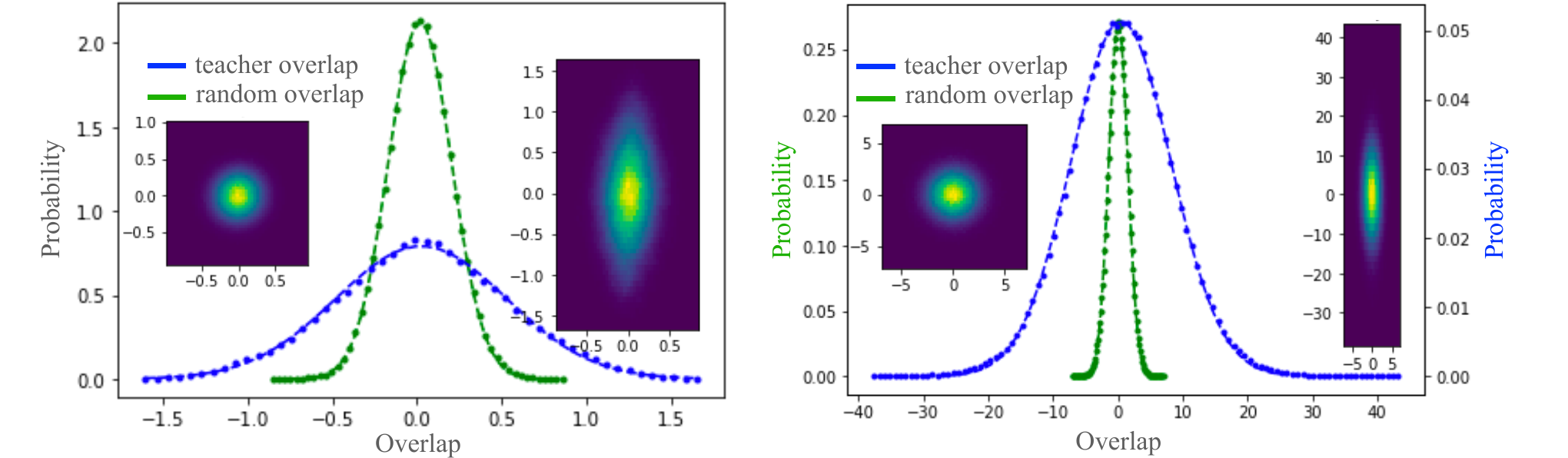}
\end{center}
% \vspace*{-0.15in}
\caption{{\bf Pre-activation statistics in the 3-layer student-teacher setting.} 
{\it Left.} Histogram of student input weight vector, dotted with the normalized teacher weight (blue) and normalized random weight (green). 
{\it Right.} Histogram of student hidden layer pre-activations, dotted with the normalized teacher pre-activations (blue) and normalized pre-activations of a random teacher (green). Dots are empirical values and dashed lines are Gaussian fits.  Insets: 2d histograms along the same vectors before (left) and after (right) training. Within our framework, these variances are determined by $\bm{v
}^{\transpose} K^{(l)} \bm{v}$ with $\bm{v}$ being either random unit vector or $\bm{h}^{(l)}$ of the single channel teacher. 
Remarkably, despite strong changes to the kernels and various non-linearity in the action, the pre-activation is almost perfectly Gaussian. 
\label{Fig:3LayerGauss}}
\end{figure}

As the first test of our theory, we examine the fluctuations of pre-activations in the input and middle layers of the trained student CNN and check their normality. Specifically, for the input weights $\bm{w}_{c}$ we obtain the histogram (over channels, equilibrium samples, and seeds) of $\bm{w}_c \cdot \bm{w}^*/|\bm{w}^*|$, where $\bm{w}^*$ is the teacher input weight and the histogram of $\bm{w}_c \cdot \bm{w}_r/|\bm{w}_r|$ where $\bm{w}_r$ is a random vector. Teacher overlap has a variance of $0.254$ here, whereas random overlap variance, averaged over choices of $\bm{w}_r$'s, was $0.039$ with an std of $0.0043$. For the hidden layer, we obtain the histogram of $\bm{h}^{(2)}_{c'} \cdot \bm{h}^{(2),*}/|\bm{h}^{(1),*}|$ where $\bm{h}^{(2),*}$ are the teacher's pre-activations as well as $\bm{h}^{(2)}_{c'} \cdot \bm{h}^{(2)}_r/|\bm{h}^{(2)}_r|$ where $\bm{h}^{(2)}_r$ is the pre-activation of a different randomly chosen teacher. Teacher overlap variance here was, $64.4$ whereas average student variance was, $2.3$ with an std of $0.12$. Fig. \ref{Fig:3LayerGauss} shows the associated histograms along with their fit to a Gaussian. 
The large and consistent differences in the variance of the fluctuations between teacher directions and random directions show that we are deep in the feature learning regime. Remarkably, the fluctuations remain almost perfectly Gaussian. The larger variance along teacher directions implies that by drawing DNNs from the trained DNN ensemble and diagonalizing their empirical covariance matrices, one is more likely to find dominant eigenvalues along these teacher directions.

\begin{figure}[h!]
\vspace*{-0.1in}
\begin{center}
\includegraphics[height=4.5cm,trim={0cm 14cm 0cm 1cm},clip]{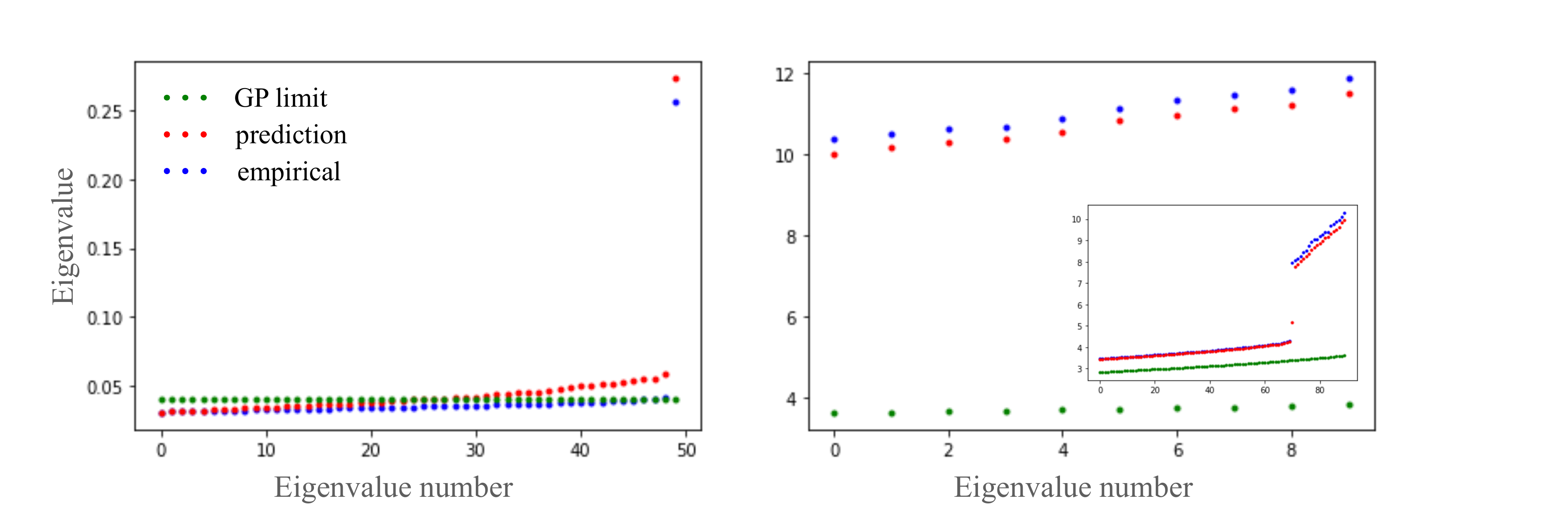}
\end{center}
% \vspace*{-0.15in}
\caption{{\bf Theoretical, empirical, and GP spectrum of kernels for the 3-layer non-linear student-teacher CNN setting.} 
{\it Left:} the leading 50 eigenvalues making up the spectrum of $\Sigma$ for the empirical $\Sigma$ (blue), for the predicted $\Sigma$ based on the equations of state (red), and for $C \rightarrow \infty$ (green). 
{\it Right:} the leading 10 eigenvalues of the {\it post- kernel} of the hidden layer ($Q^{(2)}$) again for the empirical, predicted, and GP {\it post kernels}. The inset shows the next 90 trailing eigenvalues and reveals a gap separating the leading $S_1$ eigenvalues from the rest. The dimension of $Q^{(2)}$ here is $Nn=2000$, we focus on large eigenvalues as these dominate the predictions. The outlier in the left panel is aligned with the teacher weight vector $\bm{w}^*$. The 30 (or $N$) outliers on the right panel are again a feature learning effect and represent the linear feature proportional to $\bm{w}^*$ in each of the $N$ latent pixels of the hidden layer.  
\label{Fig:EOS3}}
\end{figure}

We turn to verify the EoS and rationalize on the behavior \textit{pre-kernels}. To this end, we average the empirical pre-activations, over channels and training seeds, to obtain an estimator for the  \textit{pre-kernel} and  \textit{post-kernel} of $h^{(2)}$ (i.e. $Q^{(2)}$ and $K^{(2)}$) and that of the input weights ($\Sigma$). We then obtain $\partial D_{\text{KL}}(K^{(2)}||Q^{(2)})/ \partial \Sigma$ analytically using the 3rd equation from Eqs. \ref{Eq:3LayerEOS}, plugging in the empirical $\Sigma$. Finally, we compare the empirical $\Sigma$ with that obtained from the last equation from Eqs. (\ref{Eq:3LayerEOS}). Fig. \ref{Fig:EOS3} left panel plots the eigenvalues of $\Sigma$, our predictions ($\Sigma_{\text{pred}}$), and the \textit{post-kernel} of the input layer which is simply $\sigma_\text{w}^{-2} S_0 I_{S_0}$, showing a good match between the first two. 
Figure (\ref{Fig:EOS3}) right panel plots the eigenvalues of $Q^{(2)}$ as predicted from $K^{(2)}$, compared with its empirical value. 

\begin{figure}[h!]
\vspace*{-0.1in}
\begin{center}
\includegraphics[height=7.5cm,clip]{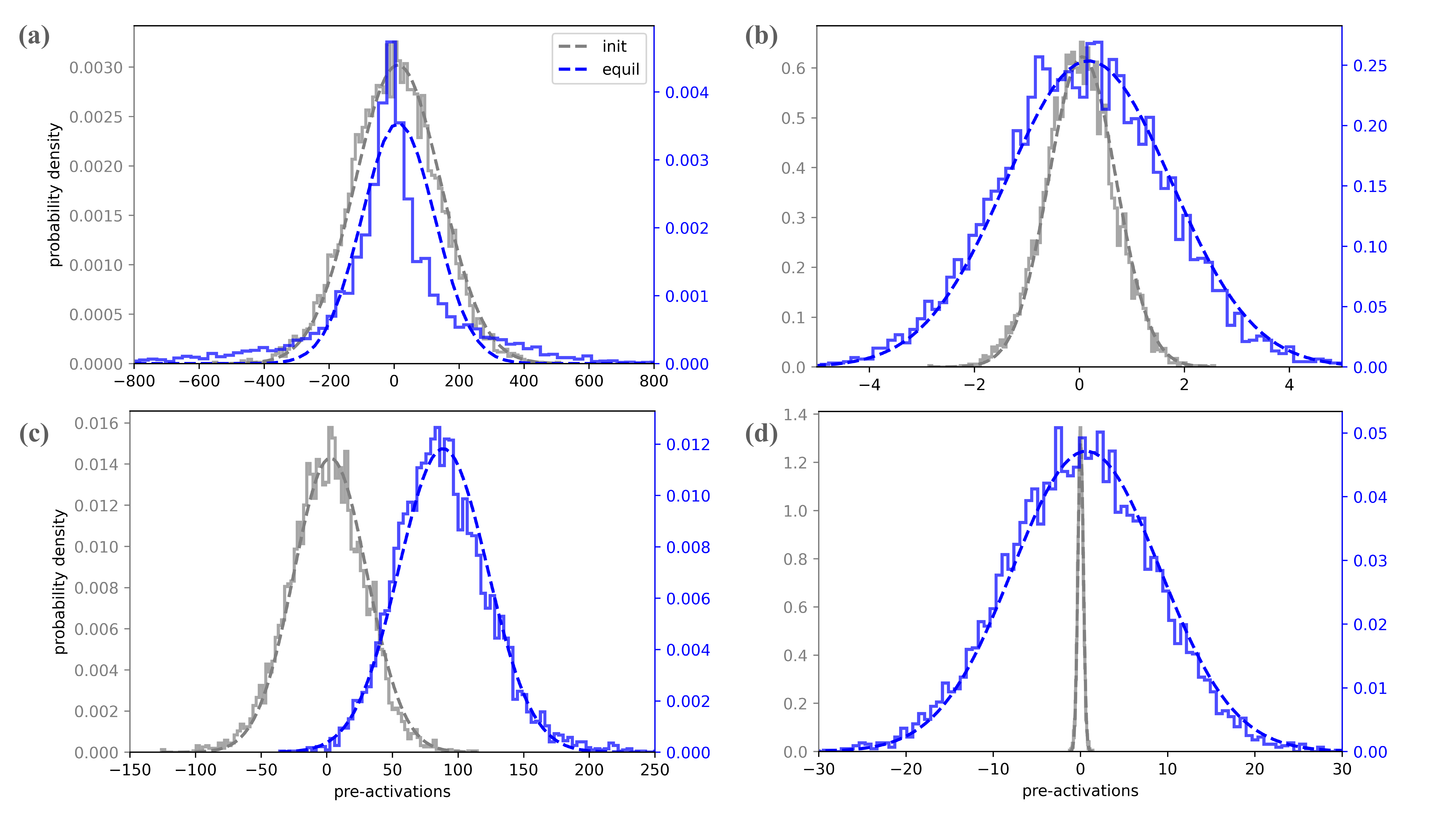}
\end{center}
\caption{{\bf Pre-activation statistics for the myrtle-5 CNN with ReLU activation and $C=256$, trained on $n=2048$ CIFAR-10 images.} 
Panels {\bf (a)}+{\bf (b)}: Statistics of pre-activations of untrained (grey) and trained (blue) Myrtle-5 nets, in the 2nd layer. The histograms are aggregations over channels and initialization seeds of the pre-activations, projected on various eigenvectors of the corresponding covariance matrix.
Panel {\bf (a)} shows a projection on the 1st (leading) eigenvector, while panel {\bf (b)} is the same for the 10th eigenvector. 
Panels {\bf (c)}+{\bf (d)} are the same as {\bf (a)}+{\bf (b)} but for the 4th layer. 
The fits to a Gaussian (dashed lines) provide an accuracy measure for our variational Gaussian approximation. 
Notice that the equilibrium distribution, even when it is near Gaussian, can differ from the initial one in either or both its mean and variance, a signature of feature learning.  
Generally, we find that Gaussianity increases with the depth of the layer (i.e. downstream layers are more Gaussian than upstream layers), and with the index of the eigenvector, we project on (e.g. here projection on the 10th eigenvector is more Gaussian than on the 1st eigenvector). 
More details can be found in Supp. Mat. (6.3), where we also show that correlations are small across different layers and across channels within each layer, thus justifying neglecting these in our theory.  
\label{Fig:Myrtle5}}
\end{figure}

Next, we trained the myrtle-5 CNN\cite{shankar2020neural}, capable of good performance and containing both pooling layers and ReLU activations, with $C=256$ on a subset of CIFAR-10 ($n=2048$). As shown in Fig. \ref{Fig:Myrtle5}, pre-activations show a strong deviation of trained DNNs from non-trained DNNs or DNNs at infinite channel/width, and at the same time show quite a good fit to Gaussian in most cases. This opens the possibility of reverse engineering the pre-kernels governing this trained network and using them to rationalize about the DNN, for instance by identifying their dominant eigenvectors. 

The 2nd layer (as well as the input layer, (see Supp. Mat. (6.3))) show deviations from Gaussianity in the leading eigenvalue. This is expected since the kernels of these layers show quite a dilute dominant spectrum whereas VGA requires a contribution from many adjacent modes (see Supp. Mat. (1.3)). 
Interestingly, despite this non-Gaussianity in the leading eigenvalue of layers 1 and 2, Gaussainity is restored in the downstream layers 3 and 4. 
Correlations across layers and across channels within the same layer are very weak (largely on the order of $10^{-3}$) and fully consistent with the mean-field decoupling underlying this work. Further technical details are found in Supp. Mat. (6.3).

\section{Discussion}
In this work, we presented what is, to the best of our knowledge, a novel mean-field framework for analyzing finite deep non-linear neural networks in the feature learning regime. Central to our analysis was a series of mean-field approximations, revealing that pre-activations are weakly correlated between layers and follow a Gaussian distribution within each layer with a pre-kernel $K^{(l)}$. Using the latter together with the post-kernel $Q^{(l)}$ induced by the upstream layer, explicit equations-of-state (EoS) governing the statistics of the hidden layers were given. These enabled us to derive, for the first time, analytical predictions for the performance of non-linear CNNs and deep non-linear FCNs in the feature learning regime. We further note that our EoS generalize straightforwardly to combined CNN-FCN architectures, pooling layers, and models with multiple outputs. Our theory can also be viewed from a Bayesian perspective, the GP process represented by the equation we find is a good approximation to the true posterior distribution generated by a large but finite width Bayesian neural network.  

Various aspects of this work invite further study. Empirically, it would be interesting to better characterize the scope of models for which Langevin dynamics (or potentially ensemble-averaged NTK dynamics) leads to Gaussian pre-activations and overall GP-like behavior. Probing the ``feature-learning-load`` of each layer, by experimentally measuring the differences between the kernels $Q^{(l)}$ and $K^{(l)}$, may also provide insights on generalization, transfer learning, and pruning, thus complementing other diagnostic tools suggested recently \cite{martin2021predicting}. For instance, transferring a layer with a small feature learning load may provide little benefit, and pruning a channel having a large overlap with a leading eigenvalue of $K^{(l)}-Q^{(l)}$ may be harmful. 

From the theory side, it is desirable to develop analytical techniques for solving the EoS as well as guarantees regarding the existence and uniqueness of solutions. In particular, exploring the possibility of spontaneous symmetry breaking of internal symmetries such as weight inversion. Providing a mathematical underpinning for the approximations involved here may lend itself to developing performance bounds on the Langevin algorithm and Bayesian neural networks\cite{seroussi2022lower,coker2022wide}. Similarly, one can consider using the empirical effective kernel ($Q_f$) as a starting point to develop GP-based bounds \cite{Seeger2002} on performance. Last, it is interesting to explore the approach to equilibrium of the training dynamics and adapt the approximations carried here to the NTK setting \cite{Jacot2018, Dyer2020Asymptotics}.

\newpage
\section{Methods \label{sec:Methods}}
\subsection{Mean Field Action}
Here we present the main ingredients of our theory, leading to the EoS we find. Further details can be found in the Supp. Mat. 

\subsubsection{Decoupling of Layers and Neurons} 
First, we provide decoupling of Eq. (\ref{eq:S_nonLin}) into layer-wise neuron-wise terms, wherein each of the terms depends on the upstream and downstream layers only through channel-averaged second moments of activations and pre-activations ({\it pre-kernel} and {\it post-kernel}). Further details are found in Supp. Mat. (1).  

Consider the non-linear terms in the action \ref{eq:S_nonLin} which couple the different layers. This coupling is mediated through the channel/width-averaged quantities: indeed $\bm{h}^{(1)}$ depends on $\bm{h}^{(2)}$ through the channel/width averaged square term in $\bm{h}^{(2)}$, $\bm{h}^{(2)}$ depends on $\bm{h}^{(1)}$ through the average of $\phi(\bm{h}^{(1)})\phi(\bm{h}^{(1)})$, and $\bm{h}^{(3)}$ depends on $\bm{h}^{(2)}$ through the average of $\phi(\bm{h}^{(2)})\phi(\bm{h}^{(2)})$ and so forth. 
For $N_l\gg 1$ we expect these to be weakly fluctuating and well approximated by their mean-field values. This behavior propagates till the output layer, and in particular implies that the outputs $\bm{f}$ fluctuate in a Gaussian manner, as previously conjectured \cite{naveh2021self}. 
As for the dependency of $\bm{h}^{(L-1)}$ on the $\bm{f}$ variables, it is not through a channel/width averaged quantity. However, we find that in various scenarios, such as FCNs with MF scaling or CNNs with large $N$, the fluctuations of $\bm{f}$ are suppressed enabling us to replace $\bm{f}$ by its average (see Supp. Mat. (1.7) and (2.2)). 
Following this, we obtain our mean-field action,
\begin{equation}
\mathcal{S}_{\mathrm{MF}}=\sum^{L-1}_{l=1}\mathcal{S}^{(l)}_{\mathrm{MF}}+\mathcal{S}_{f,\mathrm{MF}}.\label{eq:pi_MF}
\end{equation}
This allows us to define the $\langle...\rangle_{\mathrm{MF}}$ with respect to the distribution: $\pi_{\mathrm{MF}}(\{\bm{h}^{(l)}\}^{L-1}_{l=1},\bm{f})=e^{-\mathcal{S}_{\mathrm{MF}}}/\mathcal{Z}_{\mathrm{MF}}$,
and the partition function $\mathcal{Z}_{\mathrm{MF}} = \int\prod_{l=1}^{L-1}d\bm{h}^{(l)}d\bm{f} \pi_{\mathrm{MF}}(\{\bm{h}^{(l)}\}^{L-1}_{l=1},\bm{f})$.
Such that
\begin{equation}
    \mathcal{S}_{f,\mathrm{MF}}=\frac{1}{2\sigma^{2}}\sum_{\mu}(f_{\mu}-y_{\mu})^{2}+\sum_{\mu \nu} \frac{1}{2}f_{\mu}[Q_f]^{-1}_{\mu \nu}f_{\nu}\label{eq:S_{f}}
\end{equation}
\begin{equation}
    \mathcal{S}^{(l)}_{\mathrm{MF}} =\frac{1}{2} N_{l+1}\sum_{\mu\nu }A^{(l+1)}_{\mu\nu}\tilde{Q}_{\mu\nu}^{(l+1)}(\bm{h}^{(l)})+\sum_{\mu \nu j} \frac{1}{2}h_{\mu j}^{(l)}[Q^{(l)}]_{\mu\nu}^{-1}h_{\nu j}^{(l)}\,\,\,\,\,\, \text{for } l\in[1,L-1]  
\label{eq:S_{l}}
\end{equation}
where 
\begin{equation}
A^{(L)} = \sigma^{-4} (\bm{y}-\bar{\bm{f}})(\bm{y}-\bar{\bm{f}})^{\transpose} - \left[Q_{f}+\sigma^{2}I_{n}\right]^{-1},
\end{equation} 
and 
\begin{equation}
A^{(l)} =  [Q^{(l)}]^{-1}\left(I_{n}-\langle\bm{h}_{j}^{(l)}(\bm{h}_{j}^{(l)})^{T}\rangle_{\mathrm{MF}}[Q^{(l)}]^{-1}\right).
\end{equation} 
The {\it post-kernels} are defined self-consistently as $Q_f=\langle\tilde{Q}_f\rangle_{\mathrm{MF}}$, and $Q^{(l)}=\langle\tilde{Q}^{(l)}\rangle_{\mathrm{MF}}$ for $l\in[1,L-1]$. 

Notably, any coupling between the different layers is only through static mean-field quantities, namely the pre-kernels and-post kernels. In addition, all neuron-neuron couplings (and similarly, channel-channel couplings for CNNs) have been removed.

\subsubsection{Intra-Layer Decoupling} 
Despite the simplified inter-layer coupling and intra-layer neuron coupling, the mean-field actions are still non-quadratic for all layers but the output layer. This non-linearity couples all the $\bm{h}^{(l)}$ variables for the same neuron (channel in the CNN case) in a way that is roughly all-to-all in the data-point index. In atomic and nuclear physics, similar circumstances are well described by self-consistent Hartree-Fock approximations \cite{Levit1999, Fischer1977, bonneau2004, Pfannkuche1993}. In our setting, this approximation is directly analogous to a variational Gaussian approximation (VGA). In Supp. mat. (4) we argue that in the typical case where the diagonal of $K^{(l)}$ is much larger than the off-diagonal elements, the VGA is well controlled. Technically, we do so by showing, order by order in perturbation theory, that the diagrams accounted for by the VGA approximation dominate all other perturbation theory diagrams. In Supp. Mat. (3) we also establish this using different means for $S_0 \gg 1$ for the specific case of two-layer CNN with a single activated layer. We further comment that the VGA is exact for deep linear DNNs. 

Accordingly, we now look for the Gaussian distribution, governed by a kernel $K^{(l)}$ which is the closest to the above non-quadratic action. In models with many hidden layers, this leads to the following "inverse kernel shift" behavior for $1 < l < L - 2$  
\begin{align}
\label{MFEqs_1}
[[K^{(l-1)}]^{-1}]_{\mu\nu} & =[[Q^{(l-1)}]^{-1}]_{\mu\nu}+\frac{2N_l}{N_{l-1}}\frac{\partial D_{\mathrm{KL}}(K^{(l)}||Q^{(l)})\,\,}{\partial [K^{(l-1)}]_{\mu\nu}}    
\end{align}
where $l$ denotes a layer index and $D_{\mathrm{KL}}(A||B)$ is the Kullback–Leibler (KL) divergence between two multivariate Gaussians with covariance matrices $A$ and $B$. As shown in Supp. Mat. (1), for antisymmetric activation functions, the derivative of the KL-divergence is explicitly given by
\begin{equation}
\Tr[[Q^{(l)}]^{-1}(K^{(l)}-Q^{(l)})[Q^{(l)}]^{-1} (\partial Q^{(l)}/\partial K^{(l-1)}_{i\mu,j\nu})].
\end{equation} 
For non-anti-symmetric ones, see Supp. Mat. (5).

\subsection{Experimental Details}
{\bf Hyperparameters.} For the 2-layer CNN experiments, we used $S=64$, $N=20$, and varying channel number. The training parameters (noise and weight-decay) were tuned such that $\sigma^2 = 0.1$ and weight variance of $2.0$ over fan-in, for both layers at $n=0$. The target was drawn once for all experiments using i.i.d. Gaussian centered random $a^*_i$ and $w^*_s$ with variances $1/N$ and $1/S$ respectively.  

For the 3-layer CNN experiments, we took $S_1=50,S_0=30,N=2$. The training parameters (noise and weight-decay) were scaled such that $\sigma^2 = 0.005$ and weight variance of $2.0$ over fan-in for the inputs and hidden layer with no training data  (at initialization). The weight variance of the read-out layer was $15$ over the fan-in. The target was drawn again once for all experiments from a teacher CNN with $C=1$. 

For all the myrtle-5 experiments, we used $n=2048, C=256$ and ReLU activation. The training parameters (noise and weight-decay) were scaled such that $\sigma^2 = 0.005$ and weight variance of $2.0$ over fan-in for all layers with no training data (at initialization). 

For all the FCN experiments, we used equal width ($N_1=N_2$) and weight decay corresponding to variance, $\sigma_\text{w}^2=\sigma_\text{a}^2=2$ (with no training data) in the regular scaling. For the MF scaling, we took $\sigma_\text{a}^2=2/N_2$. The target was drawn again once for all experiments from a teacher CNN with $N_1=N_2=1$. Specifically when calculating the emergent scale, we used $\sigma_\text{a}^2=2/256$ independent of $N_2$.   

{\bf Equilibrium sampling.} To obtain weakly correlated samples from the equilibrium distribution of the trained CNNs we used the following procedure. For the 2 and 3-layer CNNs, we used an adaptive learning rate scheduler: For the first $100$ epochs we used a learning rate $lr_0/10$, then we crank up the learning rate to $lr_0$. As of epoch $5000$, every $1000$ epoch we estimate the fluctuations of the train-loss and check for spikes - events in which the train-loss was 5-times larger than the standard deviation in the past $500$ epochs. If a spike is observed, the learning rate is reduced by a factor of $0.7$. This continues until $50000$ epochs pass without any events. Then the learning rate is reduced again by a factor of two and remains fixed. Samples from these final stages were treated as equilibrium samples. We further checked that {\bf (i)} different initialization seeds trained with this protocol reached the same train-loss statistics. {\bf (ii)} No further reduction in train-loss occurred after the final learning rate reduction. For several runs, we also verified that increasing the last reduction of learning rate by an additional factor of $2$ did not have any appreciable effect on the loss. The initial $lr_0$ was $\sim 1e-4$ (w.r.t. a standard mean reduction MSE loss) and the final learning rate was typically $\sim 1e-5$. The runs terminated at epoch $300000$.

For the myrtle-5 CNN trained on CIFAR-10, we first ran several runs for $300000$ epochs using the above procedure and examined those that reached the lowest train-loss. We then generated a fixed scheduler based on those more successful instances, running up to $4e5$ epochs. We again verified that further lowering the final learning rate has no appreciable effect on the training loss and that different seeds reach similar final train-loss. This ensures that we are indeed sampling from a valid equilibrium distribution.   

For the 3-layer CNN and Myrtle-5, we found that auto-correlation times of pre-activations change considerably between the layers. While the read-out layer typically had an auto-correlation time of the order of $10^3$ epochs (at the lowest learning rates) the auto-correlation times for the input layers could reach $\sim 10^6$ or larger values. To overcome this issue, when analyzing pre-activations of these deeper DNNs we took an ensemble containing $98$ and $234$ different initialization seeds for the 3-layer CNN and Myrtle-5 respectively. 

For the 3-layer FCN we used a fixed scheduler which starts at $1/2$ the maximal stable learning rate and reduces the learning rate by factors of $2$ at $100,1e5,1e6,3e6$ epochs and by a factor of $4$ at $4e6,5e6$ epochs (factor of $128$ in total). Equilibrium sampling was done between $6e6-7e6$ epochs.  

{\bf Numerical solution of the equations of state.} For the 2-layer CNN, the equations of state were solved using Newton-Krylov method\cite{knoll2004jacobian}, which does not require explicit gradients. To facilitate convergence, we adopted an annealing procedure: For $C \sim 1000$, we obtain the solution using a GP initial value ($x_0$) for $\Sigma$. The optimization outcome was then used as $x_0$ for the next lower value of $C$. Using 12 CPU cores, this optimization took several hours. After obtaining $\Sigma_{ss'}$ as a function of $C$, the resulting kernel $[Q_f]_{\mu \nu}$ was used in standard GP inference to obtain $f$ on the test-set. For the 3-layer FCN we used a more efficient JAX-based code to generate the kernels and kernel derivatives involved in the EoS, but otherwise followed the same procedure. Optimization took between several minutes to a few hours on one Titan-X GPU, depending on parameters.
   
\bibliographystyle{plain}

\newpage 
%\documentclass{article}
%\usepackage[utf8]{inputenc}
%\usepackage[T1]{fontenc}
%\usepackage{lmodern}
%\usepackage{color}
%\usepackage{bm}
%\usepackage{amsmath}
%\usepackage{cancel}
%\usepackage{graphicx}
%\usepackage{microtype}
%\usepackage[unicode=true,
%bookmarks=false,
%breaklinks=false,pdfborder={0 0 1},backref=section,colorlinks=false]
%{hyperref}
%% \usepackage{hyperref}
%\makeatletter
%%%%%%%%%%%%%%%%%%%%%%%%%%%%%%% 
%\usepackage{url}% simple URL typesetting
%\usepackage{booktabs}% professional-quality tables
%\usepackage{amsfonts}% blackboard math symbols
%\usepackage{nicefrac}% compact symbols for 1/2, etc.
%\usepackage{xcolor}% colors
%\usepackage[normalem]{ulem}
%\usepackage{color}
%\usepackage{physics}
%\newcommand{\clr}{\color{red}}
%\usepackage{bm}
%\usepackage{comment}
%\date{\vspace{-5ex}}
%%-------------------
%\input{header.tex}
%%\input{ArxivV2.tex}
%-------------------
\section*{Supplementary Information - Separation of Scales and a Thermodynamic
	Description of Feature Learning in Some CNNs}

\makeatother

%\begin{document}
\maketitle
\tableofcontents
\section{Derivation of the Mean-field Equations for a Fully Connected Network}

In this section, we derive the equations of state for deep fully connected NNs with a finite number of layers, $L$. In Sec. \ref{Sec:3layerCNN} we provide a sketch of the derivation for CNN architecture. The analysis can also be extended to other architectures, including pooling layers and skip connections. 

\subsection{Model Definition and Main Players}

The model is composed of a $L$ layer NN having $L-1$ activated hidden
layers, and one linear readout layer. Specifically, we consider 
\begin{align}
f(\bm{x})&=\sum_{j=0}^{N_{L-1}-1}w^{(L)}_{j}\phi\left(h_{j}^{(L-1)}(\bm{x})\right)\\
h_{j}^{(l+1)}(\bm{x})&=\sum_{i=0}^{N_{l}-1}W^{(l+1)}_{ji}\phi\left(h_{i}^{(l)}(\bm{x})\right)\\
h_{i}^{(1)}(\bm{x})&=\sum^{d-1}_{j=0}{W}^{(1)}_{ij}{x}_j\label{eq:fcn_model}
\end{align}
where $l\in[1,L-1]$, the weights of the network are $\bm{w}^{(L)}\in\mathbb{R}^{N_{L-1}}$, ${W}^{(l)}\in\mathbb{R}^{N_{l-1}\times N_{l}}$, such that $N_0=d$ and $N_L=1$,
and $\bm{x}\in\mathbb{R}^{d}$ is
the input vector. Bold letters denote vector or tensor quantities. The function $\phi:\mathbb{R}\rightarrow\mathbb{R}$
is the activation function applied element-wise. In subsection \ref{eq:EquationsOfState_FCN}, when deriving the equation of states, we take $\phi=\mathrm{erf}$ for concreteness, other activation functions can also be considered. See section \ref{sec:VGA_ReLU} for extension to ReLU. The training data is denoted by $\mathcal{D}_{n}=\left\{ \bm{x}_{\mu},y_{\mu}\right\} _{\mu=0}^{n}$
and we use a square loss $\mathcal{L}=\sum_{\mu=1}^{n}\left(y_{\mu}-f(\bm{x}_{\mu})\right)^{2}$. The feature matrix, $X_{n}$, represents all the input samples, $\{\bm{x}_{\mu}\}_{\mu=1}^{n}$.

Our main object of interest is the following equilibrium distribution of the Langevin dynamics algorithm with noise strength $\sigma^{2}$ and weight decay written in function space (i.e. in terms of the DNNs outputs) 
\begin{align}
p(\bm{\bm{f}}|\mathcal{D}_{n}) & \propto p(\bm{f}|X_{n})\mathrm{exp}\left(-\frac{\mathcal{L}}{2\sigma^{2}}\right),\label{eq:p(f|D)_NN}
\end{align}
In practice, we sample from this distribution using Gradient descent (GD), at small learning rates, together with weight decay and noise on each weight derivative. The weight decay parameters are  $\sigma_{l}^{2}/N_{l-1}$ for layer $l$.
The first term on the r.h.s. is given by 
\begin{align}
p(\bm{f}|X_{n})=\left\langle \prod_{\mu}\delta\left[f(\bm{x}_{\mu})-\sum_{j=0}^{N_{L-1}-1}w^{(L)}_{j}\phi\left(h_{j}^{(L-1)}(\bm{x}_{\mu})\right)\right]\right\rangle_{\{W^{(l)}\}^{L-1}_{l=1}, \bm{w}^{(L)}}\label{eq:p(f|X)}
\end{align}
where $\bm{f}=(f_{1},...,f_{n})$ is viewed now as a random variable
following the NN outputs on all different training points. The average
$\langle...\rangle_{\bm{w}}$ is over the weights $\bm{w}$ of the network at equilibrium. The weights' distribution at equilibrium can be obtained explicitly. 
At $N_{l}\rightarrow\infty$, for $l\in[1,L]$
these distributions (Eq. (\ref{eq:p(f|X)}) and Eq. (\ref{eq:p(f|D)_NN})) tend to a GP, however our interest here is at finite $N_{l}$.

Eq. (\ref{eq:p(f|D)_NN}) and Eq. (\ref{eq:p(f|X)}) can also be understood from a Bayesian perspective. Eq. (\ref{eq:p(f|D)_NN}) can be viewed as the posterior distribution assuming each sample is generated by a neural network model as in Eq. (\ref{eq:fcn_model}) and is corrupted by an additive i.i.d. Gaussian noise with variance $\sigma^2$. The prior distribution over the weights of the network is taken to be Gaussian with variance $\sigma_{l}^{2}/N_{l-1}$.  

To obtain a more explicitly "layer-wise" representation of Eq. \ref{eq:p(f|X)}, we next condition over the pre-activations ($\bm{h}^{(l)}$) of each layer
using Bayes' formula, we obtain the following Markov
representation of Eq. (\ref{eq:p(f|X)})
\begin{align}
p(\bm{f}|X_{n}) &=\int p(\boldsymbol{f}|\bm{h}^{(L-1)},X_{n})\prod^{L-2}_{l=1}p(\bm{h}^{(l+1)}|\bm{h}^{(l)},X_{n})d\bm{h}^{(l)}d\bm{h}^{(L-1)}.\label{eq:Markovian stracture of P}
\end{align}
The hidden layers probabilities above are defined as follows:
\begin{align}
p(\bm{f}|\bm{h}^{(L-1)},X_{n})&=\left\langle \prod_{\mu}\delta\left(f_{\mu}-\sum_{j=0}^{N_{L}-1}w^{(L)}_{j}\phi\left(h_{j\mu}^{(L-1)}\right)\right)\right\rangle _{\bm{w}^{(L)}},\\
p(\bm{h}^{(l+1)}|\bm{h}^{(l)},X_{n})&=\left\langle \prod_{\mu j}\delta\left(h_{j\mu}^{(l+1)}-\sum_{i=0}^{N_{l}-1}W^{(l+1)}_{ji}\phi\left(h_{i\mu}^{(l)}\right)\right)\right\rangle _{W^{(l+1)}},\\
p(\bm{h}^{(1)}|X_{n})&=\left\langle \prod_{\mu i}\delta\left(h_{i\mu}^{(1)}-\sum^{d}_{j=0}W^{(1)}_{ij}{x}_{\mu,j}\right)\right\rangle_{W^{(1)}},
\end{align}
where we write for short $h_{i\mu}^{(l)}=h_{i}^{(l)}(\boldsymbol{x}_{\mu})$
for $l\in\{1,...,L-1\}$, the latter being the random variables describing argument of the activation function (pre-activation) of the $l$th layer at neuron $i$, on the $\bm{x}_{\mu}$
data-point. Later we will use this Markov structure of the distribution to decouple the different layers.

We continue our analysis by using the Fourier identity, which replaces
the above delta functions by auxiliary fields, $\bm{t},\{\bm{m}^{(l)}\}^{L-1}_{l=1}$.
To this end, the probability distribution over the network output given the input data can be written as follows.
\begin{align}
    p(\bm{f}|X_{n}) &\propto \int d\bm{t}\prod^{L-1}_{l=1}d\bm{m}^{(l)}d\bm{h}^{(l)}\exp(-\mathcal{S}\left(\bm{f},\bm{t},\{\bm{m}^{(l)},\bm{h}^{(l)}\}^{L-1}_{l=1}\right)) \label{eq:action_with_auxilery}
\end{align}
where we adopt here physics notation, and define,  $\mathcal{S}\left(\bm{f},\bm{t},\{\bm{m}^{(l)},\bm{h}^{(l)}\}^{L-1}_{l=1}\right)$, as the action associated with this distribution. Collecting all terms, the
action is defined as follows 
\begin{align}	\label{Eq:LLayerAction}
\cS &=\sum^{L-1}_{l=1}\mathcal{S}^{(l)}+\mathcal{S}_f\\ \nonumber
\mathcal{S}^{(1)}&=\frac{1}{2}\sum_{\mu\nu i}m_{\mu i}^{(1)}m_{\nu i}^{(1)}Q_{\mu\nu}^{(1)}-i\sum_{\mu j}m_{\mu j}^{(1)}h_{\mu j}^{(1)}\\ \nonumber
\mathcal{S}^{(l)} &= \frac{1}{2}\sum_{\mu\nu i}m_{\mu i}^{(l)}m_{\nu i}^{(l)}\tilde{Q}_{\mu\nu}^{(l)}(\bm{h}^{(l-1)})-i\sum_{\mu j}m_{\mu j}^{(l)}h_{\mu j}^{(l)} \\ \nonumber
\mathcal{S}_f &= -i\sum_{\mu}t_{\mu}f_{\mu}+\frac{1}{2}\sum_{\mu\nu}t_{\mu}t_{\nu}\left[\tilde{Q}_{f}(\bm{h}^{(L-1)})\right]_{\mu\nu}+\frac{1}{2\sigma^{2}}\sum_{\mu}(f_{\mu}-y_{\mu})^{2}
\end{align} 
To obtain the above expression, we performed Gaussian integration
over the weights of all layers. In addition, we define the following matrices: 
\begin{align}
\tilde{Q}_{f}(\boldsymbol{h}^{(L-1)})_{\mu\nu}=&\frac{\sigma_{L}^{2}}{N_{L-1}}\sum_{j=0}^{N_{L-1}-1}\phi(h_{j\mu}^{(L-1)})\phi(h_{j\nu}^{(L-1)})\\
\tilde{Q}^{(l+1)}(\bm{h}^{(l)})_{\mu\nu}=&\frac{\sigma_{l+1}^{2}}{N_{l}}\sum_{i=0}^{N_{l}-1}\phi(h_{i\mu}^{(l)})\phi(h_{i\nu}^{(l)}),\\
Q_{\mu\nu}^{(1)}=&\frac{\sigma_{1}^{2}}{d}\boldsymbol{x}_{\mu}^{\transpose}\boldsymbol{x}_{\nu}.\label{eq:sample free kernels}
\end{align}	
Averages of these matrices within our mean-field theory defined below are referred to henceforth as \textit{post-kernels}. 
The name \textit{post-kernel} represents the fact that at large width, these matrices concentrate around the covariance matrix of the post activation. At infinite width, these are the GP kernels, but for finite width, we show that they concentrate around a different average, which is defined self-consistently via our mean-field theory.   
The action in Eq. (\ref{Eq:LLayerAction}) is associated with the partition function $\mathcal{Z}=\int e^{-\mathcal{S}}d\bm{f}d\bm{t}\Pi^{L-1}_{l=1}d\bm{h}^{(l)}d\bm{m}^{(l)}$.
We comment that by integrating over the auxiliary fields, $\bm{t},\bm{m}^{(l)}$,
and using the following identity $\int e^{-\bm{x}^{\transpose
}A\bm{x}/2+\bm{J}\bm{x}}d^{n}\bm{x}=\sqrt{\frac{(2\pi)^{n}}{\mathrm{det}A}}e^{\bm{J}^{\transpose}A^{-1}\bm{J}/2}$,
where $\boldsymbol{J},\boldsymbol{x}\in\mathbb{R}^{n}$, and $A\in\mathbb{R}^{n\times n}$,
one obtains the equivalent form cited in the main text, containing
only the pre-activations and the outputs 
\begin{multline}
\mathcal{S}=\frac{1}{2}\sum_{\mu\nu i}h_{i\mu}^{(1)}\left[Q^{(1)}\right]_{\mu\nu}^{-1}h_{i\nu}^{(1)}
+\frac{1}{2}\sum^{L-1}_{l=2}\sum_{\mu\nu i}h_{i\mu}^{(l)}\left[\tilde{Q}^{(l)}\right]_{\mu\nu}^{-1}h_{i\nu}^{(l)}\\
+\frac{1}{2}f_{\mu}\left[\tilde{Q}_{f}(\bm{h}^{(L-1)})\right]_{\mu\nu}^{-1}f_{\nu}+\frac{1}{2\sigma^{2}}\sum_{\mu}(f_{\mu}-y_{\mu})^{2}\label{eq:action}
\end{multline}
In the following section, we derive the inter-layer mean-field decoupling. In this context, the form of the action in the presence of the auxiliary fields (Eq. (\ref{Eq:LLayerAction})) turns out to be useful.

\subsection{Mean-Field Decoupling\label{subsec:MF_decoupling_FCN}}
Next, we capitalize on the fact that the coupling between layers in Eq. \ref{Eq:LLayerAction} is only through width-index averaged quantities, specifically the width-index average of
$\bm{m}^{(l)}$ and $\tilde{Q}^{(l)}(\bm{h}^{(l-1)})$. Under such circumstances, it is natural to consider a mean-field type approximation where one replaces
\begin{align}\label{eq:MF_main_idea}
    \sum_{\mu\nu i}m_{\mu i}^{(l)}m_{\nu i}^{(l)}\tilde{Q}_{\mu\nu}^{(l)}(\bm{h}^{(l-1)}) &\rightarrow \sum_{\mu\nu i}\langle m_{\mu i}^{(l)}m_{\nu i}^{(l)}\rangle_{\mathrm{MF}} \tilde{Q}_{\mu\nu}^{(l)}(\bm{h}^{(l-1)}) \\\nonumber &+\sum_{\mu\nu i} m_{\mu i}^{(l)}m_{\nu i}^{(l)} \langle \tilde{Q}_{\mu\nu}^{(l)}(\bm{h}^{(l-1)}) \rangle_{\mathrm{MF}}
\end{align}
where $\langle...\rangle_\mathrm{MF}$ is the average over a mean-field distribution which we construct via the above approximation in a self-consistent manner. Importantly, following such a replacement, the fluctuations of the different layers as well as the different neurons within each layer become independent. This constitutes a major simplification and sets the stage for the final approximation we carry - a variational Gaussian approximation (VGA), to estimate the individual partition function of each neuron. 
This also allows us to derive the equation of states of the GP process we find. 
However, before turning to the VGA, the three subsections below provide a detailed derivation of the mean-field decoupling together with an estimate of the leading order correction to the mean-field action.  

\subsubsection{Mean-field Decoupling - Hidden Layers}
Our first step is to understand the dependence of $\bm{h}^{(l)}$ on $\bm{h}^{(l-1)}$ for all $l\in[2,L-1]$. Following the mean-field idea introduced in Eq. (\ref{eq:MF_main_idea}) 
we first subtract average quantities  and rewrite the action of the $j$th neuron of the $l$th layer such that $\mathcal{S}^{(l)} = \sum^{N_l-1}_{j=0}\mathcal{S}_j^{(l)}$:
\begin{multline}
    \mathcal{S}_j^{(l)} = \sum_{\mu\nu }\langle m_{\mu j}^{(l)}m_{\nu j}^{(l)}\rangle_{\mathrm{MF}} \tilde{Q}_{\mu\nu}^{(l)}(\bm{h}^{(l-1)})  +\sum_{\mu\nu} m_{\mu j}^{(l)}m_{\nu j}^{(l)} \langle \tilde{Q}_{\mu\nu}^{(l)}(\bm{h}^{(l-1)}) \rangle_{\mathrm{MF}}\\ +i\sum_{\mu}m_{\mu j}^{(l)}h_{\mu j}^{(l)}+\frac{1}{2}\sum_{\mu\nu }\Delta [\bm{m}^{(l)}(\bm{m}^{(l)})^\transpose]_{\mu\nu}\Delta\tilde{Q}_{\mu\nu}^{(l)}(\bm{h}^{(l-1)})+\mathrm{const} \label{eq:S_j}
\end{multline}
where $\Delta [\bm{m}^{(l)}(\bm{m}^{(l)})^\transpose]_{\mu \nu} = \frac{1}{N_l}\sum_{j}{m}_{\mu j}^{(l)}{m}_{\nu j}^{(l)} - \frac{1}{N_l}\sum_{j}\langle{m}_{\mu j}^{(l)}{m}_{\nu j}^{(l)}\rangle_{\mathrm{MF}}$, and 
$\Delta\tilde{Q}_{\mu\nu}^{(l)}(\bm{h}^{(l-1)}) = \tilde{Q}_{\mu\nu}^{(l)}(\bm{h}^{(l-1)}) - \langle \tilde{Q}_{\mu\nu}^{(l)}(\bm{h}^{(l-1)}) \rangle_{\mathrm{MF}}$. Note that, in the mean-field decoupling $\frac{1}{N_l}\sum_{j}\langle{m}_{\mu j}^{(l)}{m}_{\nu j}^{(l)}\rangle_{\mathrm{MF}}=\langle{m}_{\mu j}^{(l)}{m}_{\nu j}^{(l)}\rangle_{\mathrm{MF}}$ for any $j\in[0,N_l-1]$. The last term in Eq. (\ref{eq:S_j}) (discarding the $const$) is the fluctuations around the mean-field average. Since the sum over all neurons in the layer concentrates around its average for large width by the law of large numbers, the order of this term is $1/N_l$ smaller than all the other terms in the action. We, therefore, neglect this term. This leaves us with a Gaussian distribution over the auxiliary field, $\bm{m}^{(l)}$ which we can now integrate over and obtain: 
\begin{multline}
\int\prod_{j=0}^{N_{l}-1} d\bm{m}_{j}^{(l)}\exp\left(-{\sum_{\mu\nu }\langle m_{\mu j}^{(l)}m_{\nu i}^{(l)}\rangle_{\mathrm{MF}} \tilde{Q}_{\mu\nu}^{(l)}(\bm{h}^{(l-1)})  -\sum_{\mu\nu} m_{\mu j}^{(l)}m_{\nu i}^{(l)} \langle \tilde{Q}_{\mu\nu}^{(l)}(\bm{h}^{(l-1)}) \rangle_{\mathrm{MF}}-i\sum_{\mu}m_{\mu j}^{(l)}h_{\mu j}^{(l)}}\right)
\\  
=\exp\left( -\frac{1}{2} \sum_{\mu\nu,j}h_{\mu j}^{(l)}[Q^{(l)}]_{\mu\nu}^{-1}h_{\nu j}^{(l)}-\frac{1}{2}\sum_{\mu\nu,j} \langle m_{\mu j}^{(l)}m_{\nu j}^{(l)}\rangle_{\mathrm{MF}}\tilde{Q}_{\mu\nu}^{(l)}(\bm{h}^{(l-1)})\right)\cdot
\label{eq:integra_m_ll}
\end{multline}
In the second transition, we used the previously defined  \textit{post-kernel} $Q^{(l)} = \langle \tilde{Q}_{\mu\nu}^{(l)}(\bm{h}^{(l-1)}) \rangle_{\mathrm{MF}}$. 

Following Eq. (\ref{eq:integra_m_ll}) for $l$ and $l-1$, we gather all the $h^{(l-1)}$ dependent terms and obtain the mean-field action of the $l-1$th layer: 
\begin{equation}
\mathcal{S}^{(l-1)}_{\mathrm{MF}} =\frac{1}{2} N_{l}\sum_{\mu\nu }A^{(l)}_{\mu\nu}\tilde{Q}_{\mu\nu}^{(l)}(\bm{h}^{(l-1)})+\frac{1}{2}h_{\mu j}^{(l-1)}[Q^{(l-1)}]_{\mu\nu}^{-1}h_{\nu j}^{(l-1)}  
\label{eq:S_{l}}
\end{equation}
where we denote by $A^{(l)}_{\mu\nu} = \langle {m}_{\mu j}^{(l)}{m}_{\nu j}^{(l)}\rangle_{\mathrm{MF}}$, which due to the symmetry of the problem do not depend on the specific neuron $j$. 
The above mean-field action is valid for $l\in [2,L-1]$.  

To calculate $A^{(l)} = \langle\bm{m}^{(l)}(\bm{m}^{(l)})^{\transpose}\rangle_{\mathrm{MF}}$ for $l\in[2,L-1]$,
we use standard multivariate Gaussian results  (see subsection \ref{subsec:Auxiliary-field-correlation_mmT}) this yields the following result: 
\begin{align}
A^{(l)}&=[Q^{(l)}]^{-1}\left(I_{n}-\langle\bm{h}_{j}^{(l)}(\bm{h}_{j}^{(l)})^{\transpose}\rangle_{\mathrm{MF}}[Q^{(l)}]^{-1}\right) \label{eq:Al}
\end{align}
where $\langle \bm{h}_{j}^{(l)}(\bm{h}_{j}^{(l)})^{\transpose }\rangle_{\mathrm{MF}}$ is determined self-consistently. 

\subsubsection{Mean-Field Decoupling - Output Layer}
We next consider the coupling between the final/output layer and the penultimate layer. As done previously in the analysis of the hidden layers, we rewrite
the action of the top layer as
\begin{multline}
    \mathcal{S}_f = i\sum_{\mu}t_{\mu}f_{\mu}+\frac{1}{2\sigma^{2}}\sum_{\mu}(f_{\mu}-y_{\mu})^{2}+\frac{1}{2}\sum_{\mu\nu}\langle t_{\mu}t_{\nu}\rangle_{\mathrm{MF}}\left[\tilde{Q}_{f}(\bm{h}^{(L-1)})\right]_{\mu\nu}\\+\frac{1}{2}\sum_{\mu\nu} t_{\mu}t_{\nu}\left[\langle\tilde{Q}_{f}(\bm{h}^{(L-1)})\rangle_{\mathrm{MF}}\right]_{\mu\nu}+\frac{1}{2}\sum_{\mu\nu}\Delta[ t_{\mu}t_{\nu}]\left[\Delta \tilde{Q}_{f}(\bm{h}^{(L-1)})\right]_{\mu\nu} +\mathrm{const} 
    \label{eq:S_f}
\end{multline}
where we define $\Delta[ t_{\mu}t_{\nu}] = t_{\mu}t_{\nu} - \langle t_{\mu}t_{\nu} \rangle_{\mathrm{MF}}$ and $\Delta \tilde{Q}_{f}(\bm{h}^{(L-1)}) = \tilde{Q}_{f}(\bm{h}^{(L-1)})- \langle \tilde{Q}_{f}(\bm{h}^{(L-1)}) \rangle_{\mathrm{MF}}$.Unlike the hidden layers, here there is no averaging of many neurons, and therefore a-priori it is not clear whether the fluctuating term is small. Interestingly, we show in subsection \ref{subsec:fluctuationMF} that this term leads to $O(1/N_{L-1})$ corrections in observable quantities. Following this analysis, we focus on the remaining terms, we can now integrate over the auxiliary field $\bm{t}$ and obtain: 
\begin{multline}
\int d\bm{t}\exp\left(-\sum_{\mu}i t_{\mu}f_{\mu}-\frac{1}{2}\sum_{\mu\nu}\langle t_{\mu}t_{\nu}\rangle_{\mathrm{MF}}\left[\tilde{Q}_{f}(\bm{h}^{(L-1)})\right]_{\mu\nu}-\frac{1}{2}\sum_{\mu\nu} t_{\mu}t_{\nu}\left[\langle\tilde{Q}_{f}(\bm{h}^{(L-1)})\rangle_{\mathrm{MF}}\right]_{\mu\nu}\right)
\\
= \exp\left(-\frac{1}{2}\sum_{\mu\nu}f_{\mu}[Q_f]^{-1}_{\mu \nu}f_{\nu}-\frac{1}{2}\sum_{\mu\nu}\langle t_{\mu}t_{\nu}\rangle_{\mathrm{MF}}[\tilde{Q}_{f}(\bm{h}^{(L-1)})]_{\mu\nu}\right),
\label{eq:cumulant_expansion_f_t}
\end{multline}
where in the second transition we define the output \textit{post-kernel} as 
$Q_f = \langle\tilde{Q}_{f}(\bm{h}^{(L-1)})\rangle_{\mathrm{MF}}$.

Our mean-field decoupling is therefore valid at $N_{L-1} \gg 1$ but finite. We stress that even in this regime, feature learning effects can still be of order, $1$ as these are controlled by an emergent scale ($\chi$) containing positive powers of $n$. See Fig. 1(c) in the main text.  

For the $L-1$ layer, we obtain from Eq. (\ref{eq:S_{l}}) the following action:
\begin{align}
\mathcal{S}^{(L-1)}_{\mathrm{MF}}&=
\frac{1}{2}\sum_{\mu\nu }A^{(L)}_{\mu\nu}\tilde{Q}_{f}(\bm{h}^{(L-1)})+\sum_{j\mu\nu} \frac{1}{2}h_{\mu j}^{(L-1)}[Q^{(L-1)}]_{\mu\nu}^{-1}h_{\nu j}^{(L-1)},\label{eq:S_{L-1}}
\end{align} 
where we denote by $A^{(L)}_{\mu\nu} = \langle t_{\mu}t_{\nu}\rangle_{\mathrm{MF}}$.  Notably, the fluctuations
of the hidden layer (i.e. the $\bm{h}^{(L-1)},\bm{m}^{(L-1)}$ variables)
are now decoupled both from the input layer and the output layer.
The dependency on $\bm{h}^{(L-1)}$ is only through, $Q^{(L-1)}$ and on the $\bm{m}^{(L-1)}$
only through $\langle t_{\mu}t_{\nu}\rangle_{\mathrm{MF}}$. A tedious yet straightforward calculation similar to that carried in Sec. \ref{subsec:Auxiliary-field-correlation_mmT} reveals that 
\begin{align}
A^{(L)} &= \langle\bm{t}\bm{t} ^{\transpose}\rangle_{\mathrm{MF}}=-\bm{\delta}\bm{\delta}^{\transpose}+\left[Q_{f}+\sigma^{2}I_{n}\right]^{-1}, 
\end{align}
where the mean field average of $\bm{t}$ is then, 
\begin{equation}
i\bar{\bm{t}}=\bm{\delta}=\frac{\boldsymbol{y}-\bar{\bm{f}}}{\sigma^{2}}=\left[Q_{f}+\sigma^{2}I_{n}\right]^{-1}\bm{y}.
\end{equation}
We can also now identify the last layer of the mean-field action:
\begin{equation}
\mathcal{S}_{f,\mathrm{MF}}=\frac{1}{2\sigma^{2}}\sum_{\mu}(f_{\mu}-y_{\mu})^{2}+\frac{1}{2}f_{\mu}[Q_f]^{-1}_{\mu \nu}f_{\nu}\label{eq:S_{f}}
\end{equation}

Combining all the layers Eq. (\ref{eq:S_{l}}), Eq. (\ref{eq:S_{L-1}}), and Eq. (\ref{eq:S_{f}}), we can write the mean-field action: 
\begin{align}
\mathcal{S}_{\mathrm{MF}}=\sum^{L-1}_{l=1}\mathcal{S}^{(l)}_{\mathrm{MF}}+\mathcal{S}_{f,\mathrm{MF}}.\label{eq:pi_MF}
\end{align}
This allows us to define the $\langle...\rangle_{\mathrm{MF}}$ with respect to the distribution:
\begin{equation}
\pi_{\mathrm{MF}}(\{\bm{h}^{(l)}\}^{L-1}_{l=1},\bm{f})=e^{-\mathcal{S}_{\mathrm{MF}}}/\mathcal{Z}_{\mathrm{MF}}
\end{equation}
and the partition function $\mathcal{Z}_{\mathrm{MF}} = \int\prod_{l=1}^{L-1}d\bm{h}^{(l)}d\bm{f} \pi_{\mathrm{MF}}(\{\bm{h}^{(l)}\}^{L-1}_{l=1},\bm{f})$. We note that though this distribution is decoupled among layers and neurons in each layer, it is parameterized by mean-field average quantities. These quantities are defined self consistently and, unlike the infinite width case, create both downstream and upstream dependencies between layers. These dependencies will be fleshed out in the next subsection.

\subsection{Variational Gaussian approximation\label{eq:EquationsOfState_FCN}}

Despite reducing the full DNN into a product of decoupled partition functions per neuron and layer, the resulting actions for all but the top layer, are still non-Gaussian. Following the justifications discussed in the main text,
we approximate the latter using the variational Gaussian approximation (VGA), in which we search for the closest (in the KL divergence sense) Gaussian distribution, with general covariance matrix $K^{(l)}$, to the non-Gaussian mean-field distribution of the $l$th layer. 
We denote the optimal covariance matrix found for each layer $K^{(l)}$ as the \textit{pre-kernel}, since it is the approximated covariance of the pre-activation $\bm{h}^{(l)}$. In this section, we also assume for simplicity an antisymmetric activation function, such as $\phi=\mathrm{erf}$.
Due to $\phi$'s anti-symmetry, the DNN acquires an internal symmetry, making each pre-activation as likely as its negative. 
At large enough $N_{l}$, we do not expect spontaneous symmetry
breaking of this internal symmetry, thus one may consider a simplified version of the VGA which involves only centered Gaussian. 
We note that if the activation function is not antisymmetric as in the case of ReLU, a parameter for the mean should be added to the VGA (see Sec \ref{sec:VGA_ReLU} for more details). 

For the first layer, one is free to choose either pre-activations or the weights themselves as the variables, as these two are linear functions of one another. Here we will use the weights, $W^{(1)}_{i:}$, (the $i$th row of the matrix $W^{(1)}$) and denote the covariance matrix of these weights as $\Sigma$. The \textit{pre-kernel} matrix of the input layer pre-activations is then given by $K^{(1)} = X_n\Sigma X^{\transpose}_n$.

In the next two subsections, we perform the VGA to find the effective Gaussian kernel of the $l$th hidden layer (i.e. the {\it pre-kernel} $K^{(l)}$).  
Note that we apply the VGA for each layer $l$ where our reference distribution is the mean-field distribution we found in Sec. \ref{subsec:MF_decoupling_FCN}. The $l$th layer mean-field distribution is defined as $ e^{-\mathcal{S}^{(l)}_{\mathrm{MF}}}\propto \prod_j \pi^{(l)}_{j}$ where the distribution for each neuron $j$ and layer $l$ is $\pi^{(l)}_{j}$ and $\mathcal{S}^{(l)}_{\mathrm{MF}}$
is defined in Eq. (\ref{eq:S_{l}}), Eq. (\ref{eq:S_{L-1}}), and Eq. (\ref{eq:S_{f}}). 
We begin with the hidden layers, turn to discuss the input and final layer, and then finally gather all the terms and report our main result which is the Equations of State (EoS)

\subsubsection{$l\in[2,L-1]$}
The KL divergence between the above distribution and a Gaussian distribution for all  $j\in[0,N_{l}-1]$ with covariance $K^{(l)}$ is then
\begin{multline}
    D(\mathcal{N}(0,K^{(l)})|\pi_{j}^{(l)})=-\left\langle\frac{1}{2}\sum_{\mu\nu}h_{j\mu}^{(l)}\left[Q^{(l)}\right]_{\mu\nu}^{-1}h_{j\nu}^{(l)}\right\rangle_{K^{(l)}}\\
    -\frac{\sigma_{l+1}^{2}}{2N_{l}}\sum_{\mu\nu}A^{(l+1)}_{\mu\nu}\left\langle\phi(h_{j\mu}^{(l)})\phi(h_{j\nu}^{(l)})\right\rangle_{K^{(l)}}+\frac{1}{2}\log\det(K^{(l)})+\text{Const}.\label{eq:D_kl_l}
\end{multline}
where $\langle ... \rangle_C$ indicate average with respect to a centered Gaussian distribution with covariance matrix $C$, and recall that $A^{(l)} =\langle \bm{m}^{(l)}(\bm{m}^{(l)})^{\transpose} \rangle_{\mathrm{MF}}$.  
To find the \textit{pre-kernels} $K^{(l)}$, we take the derivative with respect to $K^{(l)}$ and set it to zero, in order to find the closest Gaussian distribution:
\begin{multline}
    \partial_{K^{(l)}}D(\mathcal{N}(0,K^{(l)})|\pi_{j}^{(l)})\\
    =-\frac{1}{2}[Q^{(l)}]^{-1}-\frac{N_{l+1}}{2N_{l}}\sum_{\mu\nu}A^{(l+1)}_{\mu\nu}\partial_{K^{(l)}}Q^{(l+1)}_{\mu\nu}+\frac{1}{2}[K^{(l)}]^{-1}=0\label{eq:div D for_K}
\end{multline}
In the first equality, since the optimal Gaussian distribution found is the closest to the mean-field one, the expectation of the \textit{post-kernel} $\tilde{Q}^{(l)}$ with respect to the Gaussian distribution is a good approximation of the mean-field expectation. This is also validated using perturbation theory in section \ref{Sec:validityVGA}. Therefore, we have that, 
\begin{multline} \sigma_{l+1}^{2}\left\langle\phi(h_{j\mu}^{(l)})\phi(h_{j\nu}^{(l)})\right\rangle_{K^{(l)}} = \sigma_{l+1}^{2}\left\langle\phi(h_{0\mu}^{(l)})\phi(h_{0\nu}^{(l)})\right\rangle_{K^{(l)}} \\ \approx\left\langle \tilde{Q}^{(l+1)}_{\mu\nu}(\boldsymbol{h}^{(l)})\right\rangle_{\mathrm{MF}}=Q^{(l+1)}_{\mu\nu}\label{eq:MF_VGA_expectation}
\end{multline}
where for the last layer $\tilde{Q}^{(L)}=\tilde{Q}_{f}$, such that $Q^{(L)}=Q_{f}$. We note here that if $Q^{(l)}$ is not full rank, then $K^{(l)}$ has the same support (vector space associated with non-zero eigenvalues) as $Q^{(l)}$. 
We can now replace in Eq. (\ref{eq:Al})  $\langle\bm{h}_j^{(l)}(\bm{h}_j^{(l)})^{\transpose{}}\rangle_{\mathrm{MF}}$ by $K^{(l)}$ such that for $l\in[2,L-1]$: 
\begin{align}
A^{(l)} \approx [Q^{(l)}]^{-1}-[Q^{(l)}]^{-1}K^{(l)}[Q^{(l)}]^{-1}.
\end{align}
Note that, this is indeed the optimal distribution since taking the second derivative yields $-[K^{(l)}]^{-2}/2$ and since $K^{(l)}$ is positive definite as a covariance matrix $-[K^{(l)}]^{-2}/2$ is negative definite, therefore it is a global minimum.   

\subsubsection{$l=1$}
We turn to find the \textit{pre-kernel} of the input layer. Here, we find it more
convenient to work with the covariance matrix of the weights rather than the pre-activation, as it is a more compact object having fewer indices. 
We follow the same procedure and minimize the KL divergence for all $i\in[0,N_{1})$ to find
the closest Gaussian distribution with covariance $\Sigma$, 
\begin{multline*}
D(\mathcal{N}(0,\Sigma)|\pi_{i}^{(1)})=-\left\langle \frac{d\left\Vert W^{(1)}_{i:}\right\Vert ^{2}}{2\sigma_{1}^{2}}\right\rangle _{\Sigma}\\
-\sigma_{2}^{2}\frac{N_{2}}{2N_{1}}\sum_{\mu\nu}A^{(2)}_{\mu\nu}\left\langle \phi(h_{i\mu}^{(1)})\phi(h_{i\nu}^{(1)})\right\rangle _{\Sigma}+\frac{1}{2}\log\det(\Sigma)+const\\
=-\frac{d}{2\sigma_{1}^{2}}\mathrm{Tr}\left(\Sigma\right)-\frac{N_{2}}{2N_{1}}\sum_{\mu\nu}A^{(2)}_{\mu\nu}\left\langle \phi(h_{i\mu}^{(1)})\phi(h_{i\nu}^{(1)})\right\rangle _{\Sigma}+\frac{1}{2}\log\det(\Sigma)+const,
\end{multline*}
taking the derivative with respect to $\Sigma$ 
\begin{equation}
\frac{\partial D(\mathcal{N}(0,\Sigma)|\pi_{i}^{(1)})}{\partial\Sigma}=-\frac{d}{2\sigma_{1}^{2}}I_{d}-\frac{N_{2}}{2N_{1}}\sum_{\mu\nu}A^{(2)}_{\mu\nu}\left[\partial_{\Sigma}Q^{(1)}\right]_{\mu\nu}+\frac{1}{2}\Sigma^{-1}=0\label{eq:div D=00003D00003D0 for_Sigma}.
\end{equation}
where again we replace the expectation within the Gaussian optimal distribution with the mean-field expectation. By the same reasoning, we can replace in Eq. (\ref{eq:Al}) the term $\langle\bm{h}_j^{(1)}(\bm{h}_j^{(1)})^{\transpose}\rangle_{\mathrm{MF}}$ by $K^{(2)}$, and use the result in Sec. \ref{subsec:Auxiliary-field-correlation_mmT}, to obtain:
\begin{align}
A^{(2)}\approx[Q^{(2)}]^{-1}-[Q^{(2)}]^{-1}K^{(2)}[Q^{(2)}]^{-1}.
\end{align}
Plugging in the above expression for $A^{(2)}$ and
combining with Eq. (\ref{eq:div D for_K}) we have that:
\[
-\frac{d}{\sigma_{1}^{2}}I_{d}-\frac{N_{2}}{N_{1}}\sum_{\mu\nu}\left([Q^{(2)}]^{-1}-[Q^{(2)}]^{-1}K^{(2)}[Q^{(2)}]^{-1}\right)_{\mu\nu}\left(\partial_{\Sigma}Q^{(2)}\right)_{\mu\nu}+\Sigma^{-1}=0.
\]

We note that following the VGA, one can use results of Ref. \cite{williams1997computing} to obtain explicit expressions for the \textit{post-kernels} $Q_f,Q^{(l)}$.
Using the definition of the \textit{pre-kernels} in Eq. (\ref{eq:MF_VGA_expectation}) we obtain for $\phi=\mathrm{erf}$, 
for all $l\in[1,L-1]$
\begin{align}
[Q^{(l+1)}]_{\mu\nu} & =\sigma_{l+1}^{2}\frac{2}{\pi}\sin^{-1}\left(\frac{2[K^{(l)}]_{\mu\nu}}{\sqrt{1+2[K^{(l)}]_{\mu\mu}}\sqrt{1+2[K^{(l)}]_{\nu\nu}}}\right),
\end{align}
where $K^{(1)}=X_n\Sigma X^{\transpose}_n$ and $Q^{(L)}=Q_f$.

\subsection{Equations of State (EoS)}
Collecting
all the results above, we obtain the following closed set of equations determining all \textit{pre-kernel} and \textit{post-kernel} as well as the average output of the Langevin algorithm, $\bar{\bm{f}}$: 
namely 
\begin{align}
\bar{\bm{f}} &= 
Q_{f}[\sigma^{2}I_{n}+Q_{f}]^{-1}\bm{y}\\{}
[Q^{(l)}]_{\mu\nu} & =\sigma_{l}^{2}\frac{2}{\pi}\sin^{-1}\left(\frac{2[K^{(l-1)}]_{\mu\nu}}{\sqrt{1+2[K^{(l-1)}]_{\mu,\mu}}\sqrt{1+2[K^{(l-1)}]_{\nu,\nu}}}\right)\nonumber \\{}
\nonumber\\{}
[[K^{(l-1)}]^{-1}]_{\mu\nu} & =[[Q^{(l-1)}]^{-1}]_{\mu\nu}+\frac{N_l}{N_{l-1}}\mathrm{Tr}\left\{A^{(l)}\frac{\partial Q^{(l)}}{\partial[K^{(l-1)}]_{\mu\nu}}\right\} \,\,\, \text{for all $l\in[2,L]$} \label{eq:K_{l-1}} \\ 
[\Sigma^{-1}]_{ss'} & =\frac{d}{\sigma_{1}^{2}}\delta_{ss'}+\frac{N_{2}}{N_{1}}\mathrm{Tr}\left[A^{(2)}\partial_{\Sigma_{ss'}}Q^{(2)}\right]  \label{eq:Sigma}\\ 
A^{(l)} &= [Q^{(l)}]^{-1}-[Q^{(l)}]^{-1}K^{(l)}[Q^{(l)}]^{-1} \,\,\, \text{for all $l\in[2,L-1]$}\nonumber\\ 
A^{(L)} &= -(\bm{y}-\bar{\bm{f}})(\bm{y}-\bar{\bm{f}})^{\transpose}\sigma^{-4}+\left[Q_{f}+\sigma^{2}I_{n}\right]^{-1}\nonumber
\end{align}
where $N_L=1$, $Q^{(L)}=Q_f$, and $K^{(1)} = X_n\Sigma X^{\transpose}_n$.  
While not explicitly apparent, the above expression does converge
to the GP limit as $N=N_{l}\rightarrow\infty$ for all $l\in[2,L-1]$.
Indeed, for very large $N$,  $K^{(L-1)}=Q^{(L-1)}+O(1/N)$. Consequently, the term $[[Q^{(l)}]^{-1}(K^{(l)}-Q^{(l)})[Q^{(l)}]^{-1}]$ is $O(1/N)$. The term on the r.h.s. thus vanishes as $1/N$.

We note that the second term in Eq. (\ref{eq:K_{l-1}}) and Eq. (\ref{eq:Sigma}) has a more
profound meaning in terms of the information transfer between the
\textit{pre-kernel} and \textit{post-kernel}. Looking at the KL divergence between two centered
multidimensional Gaussian with kernel $K^{(l)}$ and $Q^{(l)}$ of
the same dimension $m=Nn$ 
\begin{equation}
D_{\mathrm{KL}}(K^{(l)}||Q^{(l)})=\frac{1}{2}\left(\mathrm{Tr}\left([Q^{(l)}]^{-1}K^{(l)}\right)-m+\ln\left(\frac{\det Q^{(l)}}{\det K^{(l)}}\right)\right).
\end{equation}
Taking the derivative with respect to $K^{(l-1)}$ for $l\in[3,L-1]$ and with respect to $\Sigma$ for $l=2$, we
then obtain that: 
\begin{multline}
\partial_{K^{(l-1)}}D_{\mathrm{KL}}(K^{(l)}||Q^{(l)})\\=\frac{1}{2}\mathrm{Tr}\left(-[Q^{(l)}]^{-1}\partial_{K^{(l-1)}}Q^{(l)}[Q^{(l)}]^{-1}K^{(l)}+[Q^{(l)}]^{-1}\partial_{K^{(l-1)}}Q^{(l)}\right)\\
=\frac{1}{2}\mathrm{Tr}\left[\left([Q^{(l)}]^{-1}(Q^{(l)}-K^{(l)})[Q^{(l)}]^{-1}\right)\partial_{K^{(l-1)}}Q^{(l)}\right]
=\frac{1}{2}\mathrm{Tr}\left[A^{(l)}\partial_{K^{(l-1)}}Q^{(l)}\right],
\end{multline}
where in the second transition we rearranged terms and used the cyclical
property of the trace. Substituting this relation leads to Eq. (4) 
provided in the main text.

\subsubsection{Extension to test points}

The main focus of this work was on predicting the behavior of finite DNNs on the training set. Nonetheless the results could be extended to include test points in the following straightforward manner: Add an additional ``training point'' ($f_*$) whose MSE term looks like $(f_*-y_*)^2/2\tilde{\sigma}^2$ where $\tilde{\sigma}^2$ is some constant (rather than $(f_{\mu}-y_{\mu})^2/2\sigma^2$), repeat the derivation, and then once the EoS are obtained take $1/\tilde{\sigma}^2$ to zero. Alternatively, as done in Sec. \ref{App:2layer}, one can use large $n$ considerations to extend the kernel found by the EoS at finite $n$ to a continuum kernel. A more comprehensive analytical and numerical study of generalization and the EoS is left for future work.  

\subsubsection{Auxiliary field correlation
%	 $\langle\bm{m}^{(l)}(\bm{m}^{(l)})^{\transpose}\rangle_{\mathrm{MF}}$
\label{subsec:Auxiliary-field-correlation_mmT}}

To evaluate the second moment correlation of $\bm{m}^{(l)}$ for $l\in[2,L-1]$ under
the mean-field action, we follow a standard field theory technique.
We define the partition function and introduce a source field $J\in\mathbb{R}^{N_{l}\times n}$
, 
\begin{align}
\mathcal{Z}_{\boldsymbol{J}}[X_{n};A^{(l+1)},Q^{(l)}]=\prod_{j}\mathcal{Z}_{J_{j}}
\end{align}
where
\begin{align}
\mathcal{Z}_{J_{j}} = \int e^{-\frac{\sigma_{l+1}^{2}}{N_{l}}\sum_{\mu\nu}A_{\mu\nu}^{(l+1)}\phi(h_{j\mu}^{(l)})\phi(h_{j\nu}^{(l)})-\frac{1}{2}(\boldsymbol{J}_{j}+\boldsymbol{h}_{j}^{(l)})^{\transpose}[Q^{(l)}]^{-1}(\boldsymbol{J}_{j}+\boldsymbol{h}_{j}^{(l)})}d\boldsymbol{h}_{j}^{(l)}.\label{eq:ZJmm}
\end{align}
The second moment correlation then can be found by taking twice the
derivative of the following free entropy and taking the source field
to zero: 

\begin{equation}
\langle\bm{m}^{(l)}(\bm{m}^{(l)})^{\transpose}\rangle_{\mathrm{MF}}=-\frac{1}{N_{l}}\sum_{j}\triangle_{\boldsymbol{J}_{j}}\log\left(\mathcal{Z}_{\boldsymbol{J}}\right)_{\bm{J}=0}\label{eq:mm_DeltaZ},
\end{equation}
the vectors $\boldsymbol{h}_{j}^{(l)},\boldsymbol{J}_{j}\in\mathbb{R}^{n}$
refer to the $j$th column of, $\bm{h}^{(l)},\bm{J}$ respectively. Note that,
this partition function is decoupled in the neurons: 
\begin{multline}
\triangle_{J}\log\left(\mathcal{Z}_{\bm{J}}\right)=\sum_{kj}\nabla_{J_{j}}^{2}\log\left(\mathcal{Z}_{J_{k}}\right)
=\sum_{j}\nabla_{J_{j}}\left(\mathcal{Z}_{J_{j}}^{-1}\nabla_{J_{j}}\mathcal{Z}_{J_{j}}\right)\\
=-\frac{1}{2}\sum_{j}\nabla_{J_{j}}\left([Q^{(l)}]^{-1}(\boldsymbol{J}_{j}+\langle\boldsymbol{h}_{j}^{(l)}\rangle_{\mathrm{MF}})+(\boldsymbol{J}_{j}+\langle\boldsymbol{h}_{j}^{(l)}\rangle_{\mathrm{MF}})^{\transpose}[Q^{(l)}]^{-1}\right)\\
=-\frac{1}{2}\sum_{j}\left([Q^{(l)}]^{-1}\left(I_{n}+\nabla_{J_{j}}\langle\boldsymbol{h}_{j}^{(l)}\rangle_{\mathrm{MF},J}\right)+\left(I_{n}+\nabla_{J_{j}}\langle\boldsymbol{h}_{j}^{(l)}\rangle_{\mathrm{MF},J}\right)^{\transpose}Q^{(l)}]^{-1}\right)\label{eq:Hesssian_logZ},
\end{multline}
where we define
\begin{multline*}
\nabla_{J_{j}}\langle\boldsymbol{h}_{j}^{(l)}\rangle_{\mathrm{MF},J}|_{\bm{J}=0}\\=\nabla_{J_{j}}\left(\mathcal{Z}_{J_{j}}^{-1}\int e^{-\frac{\sigma_{l+1}^{2}}{2N_{l}}\sum_{\mu\nu}A_{\mu\nu}^{(l+1)}\phi\left(h_{j\mu}^{(l)}\right)\phi\left(h_{j\nu}^{(l)}\right)-\frac{1}{2}(\boldsymbol{J}_{j}+\boldsymbol{h}_{j}^{(l)})^{\transpose}[Q^{(l)}]^{-1}(\boldsymbol{J}_{j}+\boldsymbol{h}_{j}^{(l)})}\boldsymbol{h}_{j}^{(l)}d\boldsymbol{h}_{j}^{(l)}\right)|_{\bm{J}=0}\\
=-\langle\boldsymbol{h}_{j}^{(l)}\rangle_{\mathrm{MF}}\mathcal{Z}_{j}^{-1}\nabla_{J_{j}}\mathcal{Z}_{J_{j}}|_{J=0}-\left(\langle\bm{h}_{j}^{(l)}(\boldsymbol{h}_{j}^{(l)})^{\transpose}\rangle_{\mathrm{MF}}[Q^{(1\l)}]^{-1}\right)\\
=\langle\boldsymbol{h}_{j}^{(l)}\rangle_{\mathrm{MF}}\langle(\boldsymbol{h}_{j}^{(l)})^{\transpose}\rangle_{\mathrm{MF}}-\left(\langle\boldsymbol{h}_{j}^{(l)}(\boldsymbol{h}_{j}^{(l)})^{\transpose}\rangle_{\mathrm{MF}}[Q^{(l)}]^{-1}\right).
\end{multline*}
Note that, by symmetry for antisymmetric activation function: $\langle\boldsymbol{h}_{j}^{(l)}\rangle_{\mathrm{MF}} = \langle\boldsymbol{h}_{j}^{(l)}\rangle_{\mathrm{MF},\bm{J}=0}=0$, therefore, 
\[
\nabla_{\boldsymbol{J}_{j}}\langle\boldsymbol{h}_{j}^{(l)}\rangle_{\mathrm{MF},J}|_{\bm{J}=0}=-\langle\boldsymbol{h}_{j}^{(l)}(\boldsymbol{h}_{j}^{(l)})^{\transpose}\rangle_{\mathrm{MF}}[Q^{(l)}]^{-1}.
\]
Plugging this expression back in Eq. (\ref{eq:Hesssian_logZ}), and
using the fact that all matrices are symmetric, we have that 
\[
\triangle_{{\bm J}}\log\left(\mathcal{Z}_{\boldsymbol{J}}\right)=N_{l}[Q^{(l)}]^{-1}\left(\langle\boldsymbol{h}_{j}^{(l)}(\boldsymbol{h}_{j}^{(l)})^{\transpose}\rangle_{\mathrm{MF}}[Q^{(l)}]^{-1}-I_{n}\right)
\]
Plugging back in Eq. (\ref{eq:mm_DeltaZ}), we have that 
\[
\langle\bm{m}^{(l)}(\bm{m}^{(l)})^{\transpose}\rangle_{\mathrm{MF}}=[Q^{(l)}]^{-1}\left(I_{n}-\langle\boldsymbol{h}_{j}^{(l)}(\boldsymbol{h}_{j}^{(l)})^{\transpose}\rangle_{\mathrm{MF}}[Q^{(l)}]^{-1}\right).
\]

\subsection{An Emergent Scale \label{subsec:emergentscale}}
Here we identify a quantity ($\chi$) whose scale characterizes the amount of feature learning in the DNN and its deviations from the GP limit. More specifically, when $\chi$ becomes comparable to $1$, strong feature learning effects appear and perturbation theory in $1/N_{l}$ becomes impractical. Since $\chi$ would consist of a non-trivial combination of factors, we refer to it as an emergent scale.          
To define this scale, we work within our EoS. and ask when $Q_f$ changes in a noticeable manner as we lower $N_l$ from the $N_l \rightarrow \infty$ at fixed $\sigma_l^2$ (the GP limit). More technically, we next solve the EoS using perturbation theory in $N^{-1}_l$ and estimate the magnitude of the leading $O(1/N_l)$ terms we obtain. 

For simplicity, we focus on a setting where the penultimate layer is linear, due to this linearity we obtain $Q_{f}=\sigma_{L}^{2}K^{(L-1)}$ and so: 

\[
[Q_{f}]_{\mu\nu}^{-1}=\sigma_{L}^{-2}[Q^{(L-1)}]_{\mu\nu}^{-1}-\frac{1}{N_{L-1}}\left(\delta_{\mu}
\delta_{\nu}-\left[Q_{f}+\sigma^{2}I_{n}\right]_{\mu\nu}^{-1}\right)
\]
Following the aforementioned perturbation theory in $1/N_l$, we perform a first-order Taylor expansion of $Q_{f}$ yielding 
\begin{multline}
Q_{f}=\sigma_{L}^{2}\left([Q^{(L-1)}]^{-1}-\frac{\sigma_{L}^{2}}{N_{L-1}}\left(Q^{(L-1)}\boldsymbol{\delta}\boldsymbol{\delta}^{\transpose}-Q^{(L-1)}\left[Q_{f}+\sigma^{2}I_{n}\right]^{-1}\right)\right)^{-1}\\
=\sigma_{L}^{2}Q^{(L-1)}+\frac{\sigma_{L}^{2}}{N_{L-1}}Q^{(L-1)}\boldsymbol{\delta}\boldsymbol{\delta}^{\transpose}Q^{(L-1)} - \frac{\sigma_{L}^{2}}{N_{L-1}}Q^{(L-1)}\left[Q_{f}+\sigma^{2}I_{n}\right]^{-1}Q^{(L-1)}+O(N_{L-1}^{-2})\label{eq:Taylor_expansion}
\end{multline}
To evaluate the magnitude of the first term we multiply by $\boldsymbol{\delta}$
from both sides to obtain 
\begin{equation}
\boldsymbol{\delta}^{\transpose}Q_{f}\boldsymbol{\delta}=\sigma_{L}^{2} \bm{\delta}^{\transpose}Q^{(L-1)}\boldsymbol{\delta}+\frac{\sigma_{L}^{2}}{N_{L-1}}\boldsymbol{\delta}^{\transpose}Q^{(L-1)}\left(\boldsymbol{\delta}\boldsymbol{\delta}^{\transpose}-\left[Q_{f}+\sigma^{2}I_{n}\right]^{-1}\right)Q^{(L-1)}\boldsymbol{\delta}+O(N_{L-1}^{-2})
\end{equation}
We define $\chi=\frac{1}{N_{L-1}}\boldsymbol{\delta}^{\transpose}Q^{(L-1)}\boldsymbol{\delta}$,
and find: 

\begin{equation}
\frac{1}{N_{L-1}\sigma_L^{2}}\boldsymbol{\delta}^{\transpose}Q_{f}\boldsymbol{\delta}= \chi+ \chi^{2}+\frac{1}{N_{L-1}^2}\boldsymbol{\delta}^{\transpose}Q^{(L-1)}\left[Q_{f}+\sigma^{2}I_{n}\right]^{-1}Q^{(L-1)}\boldsymbol{\delta}+O(N_{L-1}^{-3})
\end{equation}

As we argue below, the last term on the r.h.s is small compared to the first two. Putting it aside and recalling that the first term on the right-hand side is the zeroth order term in $1/N_{l}$, we find that $\chi$ controls the ratio between the zeroth order contribute and the first order perturbative correction. Hence, once $\chi$ becomes order $1$, first-order perturbation in $1/N_l$ becomes inaccurate. However, it can be further checked that a second-order perturbation will contain a $\chi^3$ contribution, and hence this is not just a problem in first-order perturbation theory. Rather, it is that low order perturbation theory becomes inaccurate. 

Next, we argue that having non-negligible $\chi$ also implies feature learning, in the sense that $Q_f$ changes from its $N_{l} \rightarrow \infty$ value.  Indeed, the quantity we are examining  ($\boldsymbol{\delta}^{\transpose}Q_{f}\boldsymbol{\delta}$) involves both the discrepancy ($\boldsymbol{\delta}$) and the kernel $Q_f$. Thus, a-priory may change just because $\boldsymbol{\delta}$ changes. However, $\boldsymbol{\delta}$ is a function of $Q_f$ via $\boldsymbol{\delta} = [Q_f + \sigma^2 I]^{-1}\boldsymbol{y}$. Thus, it cannot undergo any change if $Q_f$ remains inert. Thus, we conclude that a change to $\boldsymbol{\delta}^{\transpose}Q_{f}\boldsymbol{\delta}$ must come from a change in $Q_f$ and hence, by our definition, from feature learning. This combined with the previous paragraph also shows that strong feature learning effects are beyond the practical reach of straightforward perturbation theory. 

Following some assumptions on the support of $\boldsymbol{\delta}$, the emergent scale can be presented in a more explicit manner. Indeed, assuming $\boldsymbol{\delta}$ has support mainly on $Q^{(L-1)}$ leading eigenvalues which are order $\lambda_{\max}$, we can estimate $\chi \sim n^2 \mathrm{MSE}\lambda_{\max}/N_{L-1}$, where we also use $\left\Vert \boldsymbol{\delta}\right\Vert ^{2}=n\mathrm{MSE}$. Last we note that $\chi$ is also a relevant scale for nonlinear penultimate layers since even in that case $Q_f$ can be expanded in $K^{(L-1)}$ and the linear term in this expansion will yield a correction to $\boldsymbol{\delta}^{\transpose} Q_f \boldsymbol{\delta}$ proportional to the emergent scale.  

We turn to estimate the magnitude of the term we neglected namely 
\begin{align}
\frac{1}{N_{L-1}^2}\boldsymbol{\delta}^{\transpose}Q^{(L-1)}\left[Q_{f}+\sigma^{2}I_{n}\right]^{-1}Q^{(L-1)}\boldsymbol{\delta}
\end{align}
Noting $Q_f = \sigma_L^2 K^{(L-1)}$ and that, following our EoS., $K^{(L-1)}=Q^{(L-1)} + O(1/N_{l})$, we can rewrite the above term up to $1/N_l^2$ corrections as 
\begin{align}
\frac{1}{N_{L-1}^2}\boldsymbol{\delta}^{\transpose}Q^{(L-1)}\left[\sigma_L^2 Q^{(L-1)}+\sigma^{2}I_{n}\right]^{-1}Q^{(L-1)}\boldsymbol{\delta}
\end{align}
following this one can establish that 
\begin{align}
\frac{1}{N_{L-1}^2}\boldsymbol{\delta}^{\transpose}Q^{(L-1)}\left[\sigma_L^2 Q^{(L-1)}+\sigma^{2}I_{n}\right]^{-1}Q^{(L-1)}\boldsymbol{\delta} \leq \frac{1}{N_{L-1}^2 \sigma_L^2} \boldsymbol{\delta}^{\transpose}Q^{(L-1)}\boldsymbol{\delta}=\frac{\chi}{N_{L-1}}
\end{align}
which is indeed negligible compared to $\chi$ at large $N_{L-1}$. 

\subsection{Estimating Corrections to Mean-field Results \label{subsec:fluctuationMF}}
In the mean-field derivation, when focusing on the two last layers, we neglected the term 
\begin{align}
\frac{1}{2}\sum_{\nu \mu} \Delta [t t]_{\mu \nu} [\Delta\tilde{Q}_{f}(\bm{h}^{(L-1)})]_{\mu\nu}
\end{align}
where $\Delta[t t]_{\mu \nu} = t_{\mu} t_{\nu} - \langle t_{\mu} t_{\nu}\rangle_{\mathrm{MF}}$ and $\Delta\tilde{Q}_{f}(\bm{h}^{(L-1)}) = \tilde{Q}_{f}(\bm{h}^{(L-1)})- {Q}_{f}$. However, unlike in the case of hidden layers, where the above two "$\Delta$" terms were both clearly a sum over many independent random variables (in the mean-field picture), here only the 2nd $\Delta$ is of that form. Hence, it is a priory less clear what makes this term negligible. 

In this section, we study the effect of this term in perturbation theory and when it can be neglected. 

Specifically, we treat the above term as a perturbation over the mean-field limit and calculate its leading order effect on the mean-field average of, $(\bm{f}-\bm{y})/\sigma^2$ which coincides with the mean-field average of $\bm{t}$ which is directly related to the MSE. To expose this matter in its simplest form, we shall assume that the penultimate layer is linear. Consequently $Q_f= \sigma^2_{L} K^{(L-1)}$ and in addition, VGA becomes exact (see Eq. \ref{eq:S_j} and definition of $Q_f(\bm{h}^{(L-1)})$). 

Turning to an action formulation, we focus on the following mean-field action of the $\bm{h}^{(L-1)},\bm{f}$ and $\bm{t}$ variables together with the perturbation namely, 
\begin{align}
\mathcal{S}_f = \mathcal{S}_{f,0}+\Delta \mathcal{S}_f 
\end{align}
where
\begin{align}
\mathcal{S}_{f,0}&=i\sum_{\mu}t_{\mu}f_{\mu}+\frac{1}{2\sigma^{2}}\sum_{\mu}(f_{\mu}-y_{\mu})^{2}+\frac{1}{2}\sum_{\mu\nu}\langle t_{\mu}t_{\nu}\rangle_{\mathrm{MF}}\left[\tilde{Q}_{f}(\bm{h}^{(L-1)})\right]_{\mu\nu} \\ \nonumber &+\frac{1}{2}\sum_{\mu\nu} t_{\mu}t_{\nu}\left[\langle\tilde{Q}_{f}(\bm{h}^{(L-1)})\rangle_{\mathrm{MF}}\right]_{\mu\nu}\\
\Delta \mathcal{S}_f&= \frac{1}{2}\sum_{\mu\nu}\Delta[ t_{\mu}t_{\nu}]\left[\Delta \tilde{Q}_{f}(\bm{h}^{(L-1)})\right]_{\mu\nu}. 
\end{align}
Perturbation in the $\Delta \mathcal{S}_f$ term yields the following zeroth and first-order contributions 
\begin{align}
\langle t_{\nu} \rangle &= \langle t_{\nu}\rangle_{
\mathrm{MF}} \\ \nonumber&+\frac{1}{2} \sum_{a,b}\langle t_{\nu}
\Delta [t_{a} t_{b}]\Delta\tilde{Q}_{f}(\bm{h}^{(L-1)})]_{ab} \rangle_{\mathrm{MF,con}}+ \text{Higher order terms}
\end{align}
where $\langle...\rangle_{\mathrm{MF},\mathrm{con}}$ means connected average (i.e. one in which, following Wick's theorem, the perturbation is never contracted/averaged just with itself), and $K_f=\left[Q_{f}+\sigma^{2}I_{n}\right]$. Clearly, the first order contribution is strictly zero, since by definition $\langle [\Delta\tilde{Q}_{f}(\bm{h}^{(L-1)})]_{ab}  \rangle_{\mathrm{MF}}=0$ and due to the mean-field decoupling under $\langle .. \rangle_{\rm{MF}}$, $h$'s can only contract with themselves. We thus turn to the first non-zero contribution which occurs in second order namely 
\begin{align}
\frac{1}{2!4}\langle t_{\nu} \sum_{a b d e} \Delta[tt]_{ab}\Delta [tt]_{d e}  [\Delta\tilde{Q}_{f}(\bm{h}^{(L-1)})]_{ab} [\Delta\tilde{Q}_{f}(\bm{h}^{(L-1)})]_{de}  \rangle_{\mathrm{MF,con}}.
\end{align}
Noting that we are restricted to connected averages, the fact that within the mean-field picture $\bm{h}^{(L-1)}$'s and $\bm{t}$'s are uncorrelated as well as ${h}^{(L-1)}_i$'s between different neurons, and the fact that $\tilde{Q}_f = \sigma_L^2 N_{L-1}^{-1} \sum_{j=1}^{N_{L-1}} \bm{h}^{(L-1)}_j (\bm{h}_j^{(L-1)})^{\transpose}$ we obtain  the following correction to $\bar{t}_{\mu}$ 
\begin{align}
\frac{2}{N_{L-1}} [K_f^{-1} Q_f K_f^{-1} Q_f \bar{\bm{t}}]_{
\nu} \\ \nonumber 
&+\frac{1}{N_{L-1}} [K_f^{-1} Q_f \bar{\bm{t}}]_{\nu}\mathrm{Tr}[K_f^{-1} Q_f]
\end{align}
Recalling that $K_f = Q_f + \sigma^2 I_n$, the first term in the correction to $\bar{t}_{\nu}$ is upper bounded by $\frac{2}{N_{L-1}} \bar{t}_{\nu}$ and hence negligible. The first piece of the second term is similarly upper bounded by $\frac{1}{N_{L-1}} \bar{t}_{\nu}$, however it also contains a factor of $\mathrm{Tr}[K_f^{-1} Q_f] = \sum_{\lambda} \sigma_L^2\lambda/(\sigma_L^2\lambda+\sigma^2)$, where $\sigma_L^2\lambda$'s are eigenvalues of $Q_f$. 

Let us discuss how these factors behave in various settings. The first settings are similar to that in the main text, only with standard rather than mean-field scaling for the top layer weight decay. Here, the $\mathrm{Tr}[K_f^{-1} Q_f]$ term at $\sigma^2 \rightarrow 0$ is equal to $n$. Similarly, our upper bound for the first factor becomes exact. Together, this leads to an over $n/N_{L-1}$ ratio between the correction to the mean-field result and the mean-field result for $\bar{t}_{\mu}$. Hence, when $n$ becomes larger than, $N_{L-1}$ we expect the mean-field treatment to fail. 

Turning to nonzero $\sigma^2$, the spectrum of $Q_f$ starts to play a role. Indeed, since the spectrum of $Q_f$ typically decays quickly, \cite{bordelon2020spectrum} $\mathrm{Tr}[K_f^{-1} Q_f]$ becomes proportional to the number of $Q_f$ eigenvalues which are larger than $\sigma^2$. In addition, upper bounding $[K_f^{-1} Q_f \bar{\bm{t}}]_{\nu}$ by $\bar{t}_{\nu}$ may be a very loose bound if $\bar{t}$ has support on $Q_f$ eigenvalues which are smaller than $\sigma^2$. Hence, estimating the magnitude of these corrections requires a careful case-by-case inspection. Some preliminary numerical checks we perform in teacher-student settings similar to those in the main text for $\sigma^2 \leq 0.01$ showed that the corrections to mean-field became large as soon as feature learning began. We believe that our theory still applies to FCNs away from student-teacher settings, in particular in cases where the teacher induces a target that has support mainly on sub-leading eigenvalues of $Q_f$. However, we leave this matter for future study. We comment that CNNs are different in that aspect-- as shown later below, they appear to obey our mean-field treatment within the standard scaling. 

Last, we consider nonzero $\sigma^2$ with mean-field scaling. Here one can capitalize from the fact that the mean-field scaling, scales down $Q_f$, thereby making both $[K_f^{-1} Q_f \bar{\bm{t}}]_{\nu}$ and $\mathrm{Tr}[K_f^{-1} Q_f]$ smaller. To address this issue most cleanly, let us define a GP limit to the mean-field scaling wherein we fix $\sigma^2_l$ at order $1$, $\sigma_L^2$ at order $\epsilon$ (independent from $N_{L-1}$), and take $N_l \rightarrow \infty$. Here we find that the eigenvalues of $Q_f$ are those of the standard GP limit, with an additional factor of $\epsilon$. Similarly to the emergent scale section, we can now ask when the mean-field correction becomes noticeable as we reduce $N_l$. Assuming $\chi$ is finite but not yet of order $1$, $Q_f$ is not expected to change drastically. Hence, to obtain an estimate of the mean-field correction, we may use the $N_l \rightarrow \infty$ values for $Q_f$ and $\bar{t}_{\mu}$. At small, $\epsilon$ this implies that $\mathrm{Tr}[K_f^{-1} Q_f] \approx \mathrm{Tr}[Q_f]/\sigma^2$ which is of the order of $n \epsilon/\sigma^2$. Similarly, $[K_f^{-1} Q_f \bar{\bm{t}}]_{\nu}$ can be estimated by $\epsilon \lambda_{\max}/\sigma^2$, where $\lambda_{\max}$ is the dominant eigenvalue of $Q_f$ on which $\bm{t}$ has support. Putting this together with the ${N_{L-1}^{-1}}$ factor we find that $n \lambda_{\max} \epsilon^2/(N_{L-1} \sigma^4) \ll 1$ is required which is a much less stringent requirement. Indeed, by taking $\epsilon$ to zero, one can make this as small as needed. 

A key point is that making $\epsilon$ smaller makes $\chi$ bigger. Hence, our mean-field treatment works within the feature learning regime. Indeed, the emergent scale is defined as $\bm{\delta}^{\transpose} Q^{(L-1)} {\bm \delta}/N_{L-1}$ and hence does not involve $\epsilon$ directly. Decreasing $\epsilon$ can only make ${\bm \delta}$ larger since ${\bm \delta} = [Q_f + \sigma^2 I_n]^{-1} {\bm y}$. Thus, decreasing $\epsilon$ increases feature learning while making mean-field corrections more negligible. On this note, we comment that in our FCN experiments we found, similarly to Ref. \cite{Yu2020}, that mean-field scaled FCNs had better test performance compared to those which used standard scaling similar to

\section{Mean-Field Equations for CNNs
\label{Sec:3layerCNN}}

In this section, we provide a sketch of the derivation of the equations of state for deep CNNs highlighting the differences between CNN and FCN. For simplicity, we follow here the derivation of a three-layer CNN. The generalization to any number of layers is straightforward. The model we consider is a three-layer CNN having two activated convolutional layers  and one linear readout layer. Specifically, we consider 
\begin{align}
f(\bm{x}) &= \sum_{j=0}^{N-1} \sum_{c'=1}^{C_2} a_{c'j} \phi\left(h^{(2)}_{c' j}(\bm{x}) \right) \\ \nonumber 
h^{(2)}_{c' j}(\bm{x}) &= \sum_{i=0}^{S_1-1}\sum_{c=1}^{C_1} v_{c' c i}\phi\left(\bm{w}_{c} \cdot \bm{x}_{i+j S_1} \right) 
\end{align}
where $\bm{w}_c,\bm{x}_{i+jS_1}\in\mathbb{R}^{S_0}$, $\bm{a}\in\mathbb{R}^{C_2 \times N}$, $\bm{v}\in\mathbb{R}^{C_2 \times C_1 \times S_1}$, and the input vector $\bm{x}\in\mathbb{R}^d$ with $d=S_0 S_1 N$. Bold letters mark vector or tensor quantities. 
Similar to the FCN case, we analyze the following distribution over the pre-activations  
\begin{align}
p(\bm{\bm{f}}|\mathcal{D}_{n})
&\propto p(\bm{f}|X_n)\mathrm{exp}\left(-\frac{1}{2\sigma^{2}}\sum_{\mu=1}^{n}\left(y_{\mu}-f(\bm{x}_{\mu})\right)^{2}\right),
\label{eq:p(f|D)}
\end{align}
where the matrix $X_n$ represents all the samples, $\{\bm{x}_\mu\}^n_{\mu=1}$, and the first term on the r.h.s. is given by 
\begin{align}
p(\bm{f}|X_n)
&=\left \langle \prod_{\mu}\delta \left[f(\bm{x}_{\mu}) - \sum_{j=0}^{N-1} \sum_{c'=1}^{C_2} a_{c'j} \phi\left(h^{(2)}_{c' j}(\bm{x}_{\mu}) \right)\right]\right\rangle_{\bm{a,v,w}} \label{eq:p(f|X_n)_CNN}
\end{align}
where $\bm{f}=(f_1,...,f_n)$ is viewed now as a random variable following the CNN outputs on all different training points, and $\langle...\rangle_{\bm{a,v,w}}$ denote average over the weights $\bm{a,v,w}$. 
The above probability can be viewed as the prior induced on $\bm{f}$ by a finite random DNN with i.i.d. Gaussian weights and variance $\sigma_\text{w}^2/S_0,\sigma_\text{v}^2/(S_1C_1)$ and $\sigma_\text{a}^2/(NC_2)$ respectively for each layer. At $C_1,C_2\rightarrow \infty$ such priors tend to a GP, however our interest here is at finite $C_1,C_2$.

By conditioning on the pre-activation output ($\bm{h}^{(2)}$), Eq. (\ref{eq:p(f|X_n)_CNN}) can be re-written as 
\begin{align}
p(\bm{f}|X_n)
&=\int p(\boldsymbol{f}|\bm{h}^{(2)},X_{n})p(\bm{h}^{(2)}|\boldsymbol{w},X_{n})p(\bm{w})\mathcal{D}\bm{h}^{(2)}\mathcal{D}\bm{w}
\end{align}
where the hidden layers probabilities used above are 
\begin{align}
p(\bm{f}|\bm{h}^{(2)},X_{n})&=\left\langle \prod_{\mu}\delta\left(f_{\mu}-\sum_{j=0}^{N-1}\sum_{c'=1}^{C_2}a_{c'j}\phi\left(h^{(2)}_{c'j \mu}\right)\right)\right\rangle_{\bm{a}}, \\ \nonumber 
p(\bm{h}^{(2)}|\bm{w},X_{n})&=\left\langle\prod_{\mu c'j}\delta\left(h^{(1)}_{c'j\mu}-\sum_{i=0}^{S_1-1}\sum_{c=1}^{C_1}v_{c'ci}\phi\left(\bm{w}_{c}\cdot\bm{x}_{\mu,i+jS_1}\right)\right)\right\rangle_{\bm{v}}, 
\end{align}
the tensor $\bm{h}^{(2)}$ consists of all $h^{(2)}_{i c \mu}$ the latter being the random variables describing outputs of the second layer at hidden pixel $i$, channel $c$ on the $\bm{x}_{\mu}$ data-point.  

Similar to the FCN, we continue our analysis by using the Fourier identity, which replaces the delta function with auxiliary fields $\bm{t}, \bm{m}$. 	The resulting action ($Z = \int e^{-\cS}$) is then, 
\begin{align}
\label{AppEq:3LayerAction}
\cS &= \frac{S_0}{2 \sigma_\text{w}^2} \sum_c \norm{\bm{w}_c}^2 - i\sum_{\mu c'j}m_{\mu c'j}h^{(2)}_{c'j}(\bm{x}_{\mu})+\frac{1}{2}\sum_{\mu\nu j_{1}j_{2}c'}m_{\mu c'j_{1}}m_{\nu c'j_{2}}\tilde{Q}^{(2)}(\bm{w})_{\mu\nu j_{1}j_{2}}
\\ \nonumber &-i\sum_{\mu}t_{\mu}f_{\mu} + \frac{1}{2}\sum_{\mu\nu}t_{\mu}t_{\nu}\left[\tilde{Q}_{f}(\bm{h}^{(2)})\right]_{\mu\nu}
% \sum_{\mu} 
+\frac{1}{2 \sigma^2} \sum_{\mu} (f_{\mu} - y_{\mu})^2
\end{align}  
Where the ``channel'' \textit{post-kernels} for CNN contain also summation overstrides and are defined as: 
\begin{align}
\tilde{Q}_f(\boldsymbol{h}^{(2)})_{\mu\nu}&=\frac{\sigma_\text{a}^2}{C_2 N}\sum_{jc'}^{N,C_2}\phi\left(h^{(2)}_{c'j\mu}\right)\phi\left(h^{(2)}_{c'j\nu}\right)
\\ \nonumber 
\tilde{Q}^{(2)}(\boldsymbol{w})_{\mu j_{1},\nu j_{2}}&=\frac{\sigma_\text{v}^{2}}{C_1 S_1}\sum^{S_1,C_1}_{ic}\phi\left(\boldsymbol{w}_{c}\cdot\boldsymbol{x}_{\mu,i+j_{1}S_1}\right)\phi\left(\boldsymbol{w}_{c}\cdot\boldsymbol{x}_{\nu,i+j_{2}S_1}\right).
\end{align}

We comment that by averaging over the auxiliary fields, $\bm{t},\bm{m}$, one obtains the following equivalent form, containing only the pre-activations and the outputs 
\begin{multline}
\cS = \frac{S_0}{2 \sigma_\text{w}^2} \sum_c \norm{\bm{w}_c}^2 + \frac{1}{2}\sum_{c'\mu\nu j j'} h^{(2)}_{c' j \mu} \left[\tilde{Q}^{(2)}(\bm{w})\right]^{-1}_{j \mu,j' \nu} {h}^{(2)}_{c' j'\nu} 
\\ \nonumber +\frac{1}{2}f_{\mu} \left[ \tilde{Q}_f(\bm{h}^{(2)})\right]^{-1}_{\mu,\nu} f_{ \nu} +
% \sum_{\mu} 
\frac{1}{2 \sigma^2} \sum_{\mu} (f_{\mu} - y_{\mu})^2
\end{multline}

We continue with the auxiliary variables and derive the inter-layer mean-field decoupling. Where for CNN the number of channels $C_1, C_2$ plays the role of width in FCN. As for FCN, we note that in Eq. (\ref{AppEq:3LayerAction}), the hidden layer and the output layer depend on their respective upstream layers only through the "channel" \textit{post-kernels}. Performing our mean-field decoupling as in Sec. \ref{subsec:MF_decoupling_FCN} we obtain the resulting action   
\begin{align}\label{eq:actionsMF_CNN}
\cS_{\mathrm{MF}} &= \cS^{(1)}_{\mathrm{MF}}+\cS^{(2)}_{\mathrm{MF}}+\cS_{f,\mathrm{MF}}\\ \nonumber 
\cS^{(1)}_{\mathrm{MF}} &= \frac{S_0}{2 \sigma_\text{w}^2} \sum_c \norm{\bm{w}_c}^2 +\frac{1}{2}\sum_{\mu\nu j_{1}j_{2}c'}\langle{\bm{m}\bm{m}^{\transpose}}\rangle_{\mathrm{MF},\mu j_1 \nu j_2}\tilde{Q}^{(2)}(\bm{w})_{\mu j_{1} \nu j_{2}} \\ \nonumber 
\cS^{(2)}_{\mathrm{MF}} &=  \frac{1}{2}\sum_{\mu\nu}\langle t_{\mu}t_{\nu}\rangle_{\mathrm{MF}}  \left[ \tilde{Q}_{f}(\bm{h}^{(2)})\right]_{\mu \nu}
+\frac{1}{2}
\sum_{\mu c'j_1,j_2}h^{(2)}_{c'j_1}(\bm{x}_{\mu})[{Q}^{(2)}]_{\mu j_{1} \nu j_{2}}^{-1} h^{(2)}_{c'j_2}(\bm{x}_{\mu})\\ \nonumber 
\cS_{f,\mathrm{MF}} &=  \frac{1}{2 \sigma^2} \sum_{\mu} (f_{\mu} - y_{\mu})^2 + \frac{1}{2}\sum_{\mu \nu}f_{\mu} [Q_f]_{\mu \nu}^{-1}f_{\nu},
\end{align}
where the \textit{post-kernel} can now be defined as the mean-field average of the ``channel'' \textit{post-kernel} via the above mean-field action distribution: 
\begin{align}
[Q_f]_{\mu \nu}  &= \left\langle\tilde{Q}_{f}(\boldsymbol{h}^{(2)})_{\mu\nu}\right\rangle_{\mathrm{MF}} \\ \nonumber 
[Q^{(2)}]_{\mu j, \nu j'} &= \left\langle\tilde{Q}^{(2)}(\boldsymbol{w})_{\mu j_{1},\nu j_{2}}\right\rangle_{\mathrm{MF}}.  
\end{align}
The correlation functions similar to the FCN are then,
\begin{equation}
\langle{\bm{m}\bm{m}^{\transpose}}\rangle_{\mathrm{MF}} =	\left \langle C_2^{-1} \sum^{C_2}_{c',j=1} \bm{m}_{c'}\bm{m}^{\transpose}_{ c'}\right \rangle_{\mathrm{MF}}= [Q^{(2)}]^{-1}\left(I_{n}-\langle\bm{h}_{1}^{(2)}(\bm{h}_{1}^{(2)})^{\transpose}\rangle_{\mathrm{MF}}[Q^{(2)}]^{-1}\right)
\end{equation}
The last layer of auxiliary field correlation is 
\begin{align}
\langle t_{\mu} t_{\nu} \rangle_{\mathrm{MF}} = -\delta_{\mu} \delta_{\nu} + \left[(Q_{f} + \sigma^{2}I_{n})^{-1}\right]_{\mu \nu}
\end{align}
this is easily derive from Eq. (\ref{eq:actionsMF_CNN}). The average of $\bm{t}$ using the mean-field action is then,  
\begin{align}
i\bar{\bm{t}} =\bm{\delta}= \frac{\boldsymbol{y}-\bar{\bm{f}}}{\sigma^{2}} = \left(Q_{f} + \sigma^{2}I_{n}\right)^{-1}\bm{y},
\end{align}
where we denote by bar the $\langle ...\rangle_{\mathrm{MF}}$ corresponding to 
averages with respect to the mean-field distribution. 
Note that, in the main text, we denote $\tilde{Q}^{(2)}(\bm{w})$ by $\tilde{Q}^{(2)}(\bm{h}^{(1)})$. These are the same quantities, the connection being ${h}_{cij}^{(1)}=\bm{w}_c \cdot \bm{x}_{i+jS_1}$.

Despite reducing the full system into decoupled systems per channel and layer, the resulting action for all but the top layer is still non-Gaussian. Following the justifications discussed in the main text and for the FCN, we approximate the latter using the variational Gaussian approximation (VGA). Assuming for simplicity an antisymmetric activation function such as $\erf$, the CNN has an internal symmetry, making each pre-activation positive output as likely as its negative. Also, at large enough $C_1,C_2$ we do not expect spontaneous symmetry breaking, thus our VGA will involve only a centered Gaussian. Specifically, we denote the optimal covariance of $\bm{h}^{(2)}$, $K^{(2)}_{\mu j \nu j'}$ as the \textit{pre-kernel} of the second layer and the optimal covariance of $\bm{w}$, $\Sigma_{ss'}$, is a connected to the first layer \textit{pre-kernel}, $K^{(1)} = X_n\Sigma X_n^{\transpose} $. 

Following the analysis in subsection. \ref{eq:EquationsOfState_FCN} with the above modification for CNN, we obtain the following closed set of equations determining all \textit{pre-kernels} as well as the outputs namely
\begin{align}
\bar{\bm{f}} &= 
Q_{f}[\sigma^{2}I_{n}+Q_{f}]^{-1}\bm{y}\\ \nonumber 
[Q_f]_{\mu \nu} &= \sigma_\text{a}^2 \frac{1}{N}\sum_j \frac{2}{\pi} \sin^{-1}\left(\frac{2 [K^{(2)}]_{\mu j,\nu j}}{\sqrt{1+2[K^{(2)}]_{\mu j,\mu j}}\sqrt{1+2[K^{(2)}]_{\nu j,\nu j}}} \right) \\ \nonumber
[Q^{(2)}]_{\mu j, \nu j'} &= \sigma_\text{v}^2 \frac{1}{S_1} \sum_{i} \frac{2}{\pi}\sin^{-1}\left(\frac{2 \bm{x}_{\mu j i}\Sigma \bm{x}_{\nu j' i}}{\sqrt{1+2\bm{x}_{\mu j i}\Sigma \bm{x}_{\mu j i}}\sqrt{1+2\bm{x}_{\nu j' i}\Sigma \bm{x}_{\nu j' i}}} \right) \\ \nonumber
[(K^{(2)})^{-1}]_{\mu j,\nu j'} &= [(Q^{(2)})^{-1}]_{\mu j,\nu j'} - \frac{1}{C_2}\Tr{A^{(3)} \frac{\partial Q_f}{\partial [K^{(2)}]_{\mu j,\nu j'}}}  \\ \nonumber 
[\Sigma^{-1}]_{ss'} &= \frac{S_0}{\sigma_\text{w}^2} \delta_{ss'} - \frac{C_2}{C_1} \Tr\left[[(Q^{(2)})^{-1}(K^{(2)}-Q^{(2)}) (Q^{(2)})^{-1}]\partial_{\Sigma_{ss'}}Q^{(2)}\right]\\\nonumber 
A^{(3)} &= (\bm{y}-\bar{\bm{f}})(\bm{y}-\bar{\bm{f}})^{\transpose}\sigma^{-4}-\left[Q_{f}+\sigma^{2}I_{n}\right]^{-1}\nonumber 
\end{align}
\subsection{An Emergent Scale in CNNs}

Following the FCN case, we again examine the EoS for the penultimate layer assuming that it is linear, solve them using perturbation theory in $1/C_l$, and use the magnitude of the correction to estimate the scale at which feature learning becomes important. For our CNNs we have that $[Q_f]_{\mu \nu} = \sigma_L^2 N^{-1} \sum_j K^{(L-1)}_{\mu j,\nu j}$ which yields the following equation of state for $K^{(L-1)}$  

\[
[K^{(L-1)}]_{\mu i,\nu j}^{-1}=[Q^{(L-1)}]_{\mu i,\nu j}^{-1}-\frac{\delta_{ij}}{C_{L-1}N}\left(\delta_{\mu}\delta_{\nu}-\left[Q_{f}+\sigma^{2}I_{n}\right]_{\mu\nu}^{-1}\right)
\]
Next, we perform a leading order perturbation theory in $1/C_l$ (or equivalently in the second term on the r.h.s) yielding 

\begin{multline}
[K^{(L-1)}]_{\mu i, \nu j} = [Q^{(L-1)}]_{\mu i, \nu j} +\frac{1}{C_{L-1} N} \sum_{k a b} [Q^{(L-1)}]_{\mu i, a k} \delta_a \delta_b [Q^{(L-1)}]_{b k, \nu j}\\ \nonumber 
- \frac{1}{C_{L-1} N} \sum_{k a b } [Q^{(L-1)}]_{\mu i, a k} [Q_f+\sigma^2 I_n]^{-1}_{ab} [Q^{(L-1)}]_{b k, \nu j}+O(1/(C_{L-1}^2))
\end{multline}
As justified in the FCN case, we focus on the second term on the r.h.s. and look again at $\bm{\delta}^{\transpose} Q_f \bm{\delta}$ yielding,  
\begin{align}
\bm{\delta}^{\transpose} Q_f \bm{\delta} &= \frac{\sigma^2_L}{N} \sum_{i \mu \nu} \delta_{\mu} Q^{(L-1)}_{\mu i, \nu i} \delta_{\nu} + \frac{\sigma_L^2}{C_{L-1} N^2} \sum_{i kab\mu\nu}\delta_{\mu}  [Q^{(L-1)}]_{\mu i, a k} \delta_a \delta_b [Q^{(L-1)}]_{b k, \nu i} \delta_{\nu} 
\end{align}
We define the ratio of the second to the first term as the emergent scale. Notably for $N=1$ it coincides with the definition for FCNs. In addition to considering the 2-layer CNN studied in the main text and estimating it using the same approximations, it results in the same scale. 

\subsection{Estimating Mean-field corrections - CNNs \label{sec:fluctuationCNN}}
In the FCN case, we found that an additional ingredient, on top of large $N_{l}$, is needed for our mean-field decoupling to hold - either mean-field scaling or a target with support only on weak $Q_f$ eigenvalues. For the CNNs we have studied, and quite possibly for a much larger family of CNNs, this additional ingredient comes naturally from the read-out layer averages over $N$ latent pixels in the penultimate layers. Since these are expected to be somewhat independent, one can hope that summing over these terms is similar to increasing the number of channels by a factor of $N$ (the number of pixels in the penultimate layer). Here we establish this more concretely. Similar to the FCN we calculate the average of the discrepancy $t_\mu$ up to the second order: 
\begin{multline}
\langle t_\mu \rangle = \langle t_\mu \rangle_{\mathrm{MF}} + \frac{1}{2!4}\langle t_{\mu}\sum_{abde}\Delta[tt]_{ab}\Delta[tt]_{de}[\Delta\tilde{Q}_{f}(\bm{h}^{(L-1)})]_{ab}[\Delta\tilde{Q}_{f}(\bm{h}^{(L-1)})]_{de}\rangle_{\mathrm{MF},\mathrm{con}}\\+\text{Higher order terms}
\end{multline}
For CNN second order term can be further simplified by using Wick theorem for pre-activations of the last layer when again we consider the case of linear activation function
\begin{equation}
\frac{\sigma_{L}^{4}}{8N^{2}C_{L-1}}\sum_{ji}\sum_{abde}\langle t_{\mu}\left(t_{a}t_{b}-\langle t_{a}t_{b}\rangle_{\mathrm{MF}}\right)\left(t_{d}t_{e}-\langle t_{d}t_{e}\rangle_{\mathrm{MF}}\right)\left(K_{ajei}^{(L-1)}K_{bjdi}^{(L-1)}+K_{ajdi}^{(L-1)}K_{bjei}^{(L-1)}\right)\rangle_{\mathrm{MF},\mathrm{con}}
\end{equation}
A similar derivation to \ref{subsec:fluctuationMF} yields the following correction to $\bar{t}_{\mu}$ ($(f_{\mu}-y_{\mu})/\sigma^2$) 
\begin{align}
&\frac{2\sigma_{L}^{4}}{C_{L-1} N^2} \sum_{abde i j} [K_f^{-1}]_{ 
\mu a} K^{(L-1)}_{a i,bj} [K_f^{-1} ]_{bd} K^{(L-1)}_{d i, e j} \bar{t}_{e}\\ \nonumber 
&+\frac{\sigma_{L}^{4}}{C_{L-1} N^2} \sum_{i j}\left( \sum_{ab} [K_f^{-1}]_{\mu a} K^{(L-1)}_{a i, b j} \bar{t}_{b}\right) \left( \sum_{ab} [K_f^{-1}]_{ab} K^{(L-1)}_{i a, j b}\right)
\end{align}

Unlike in the FCN case, $K^{(L-1)}\in \mathbb{R}^{Nn\times Nn}$ is not proportional to $[K_f]_{\mu \nu}=\sigma^2\delta_{\mu\nu}+\sigma^{2}_LN^{-1}\sum_{j} K_{\mu j,\nu j}^{(L-1)}$ where $\mu,\nu\in[1,n]$, in the $\sigma^2 \rightarrow 0$ limit. In particular, for $N>1$, the two matrices have different dimensions. Hence, we cannot cancel them together just yet. 
To obtain an order of magnitude estimate of the second (and more dominant) term in this perturbation, we turn to a different route and compare the mean-field value of the norm, $\sum_{\mu} |\langle t_{\mu} \rangle|^2$ which is $\sum_{\mu} \delta^2_{\mu}$ with its leading correction. Following similar arguments to the FCN section, we find that the correction is dominated by   
\begin{align}
\frac{\sigma_{L}^{4}}{C_{L-1} N^2} \sum_{i j}\left( \sum_{ \mu ab} \bar{t}_{\mu} [K_f^{-1}]_{\mu a} K^{(L-1)}_{a i, b j} \bar{t}_{b}\right) \left( \sum_{ab} [K_f^{-1}]_{ab} K^{(L-1)}_{i a, j b}\right)
\end{align}
To simplify this expression we next note that for data-sets in which for each $\bm{x}_{\mu}$ there exists a "symmetry-partner" point wherein all coordinates in the fan-in of the $i$'th latent pixels are flipped - the second, trace-like, term must vanish for $i \neq j$. This is due to the fact that $K_{f}$ is invariant under the action of the associated symmetry, whereas $K^{(L-1)}_{* i,* j}$ receives a minor sign whenever $i \neq j$. As $n \rightarrow \infty$ we expect this symmetry to be approximately realized as it is a symmetry of the underlying measure from which $\bm{x}_{\mu}$ are drawn. 
Following this, we remove $i \neq j$ terms from the summation. 

Next we notice that due to the approximate translation symmetry of the dataset, at large $n$ 
\begin{align}
\sum_{ab} [K_f^{-1}]_{ab} K^{(L-1)}_{i a, i b} 
\end{align}
becomes independent of $i$. We thus replace it by its average and perform the remaining summation over, $i$ which now involves only the first term to obtain 
\begin{align}
\frac{1}{C_{L-1} N} \left( \sum_{ \mu ab} \bar{t}_{\mu} [K_f^{-1}]_{\mu a} [Q_f]_{ab} \bar{t}_{b}\right) \left( \sum_{ab} [K_f^{-1}]_{ab} [Q_{f}]_{ab}\right) 
\end{align}
Recall that $[Q_f]_{\mu\nu} = \sigma_{L}^{2}N^{-1} \sum_j K_{\mu j, \nu j}^{(L-1)}$. This resulting expression is very similar to its FCN version (with standard scaling) with one crucial difference, which is the appearance of the aforementioned $1/N$ factor. More specifically, the first summation is smaller than the zeroth term ($\sum_{\mu} t_{\mu}^2$). The scale controlling the mean-field decoupling is therefore $\frac{1}{C_{L-1} N} \Tr[K_f^{-1} Q_f]$. Thus, at large $N$, we can have a reliable mean-field decoupling even when $n=C_{L-1}$.  

\section{Toy Example - One Hidden Layer}
\label{App:2layer}
In this section, we provide a detailed calculation of the toy example presented in the main text (subsection 2.3). We consider $\bm{x}_{\mu}$'s as a center Gaussian vector of size $NS$ with random i.i.d. entries with zero mean and variance $1$. We choose a target of the form $y(\bm{x})=\sum_i a^*_i \bm{w}^* \cdot \bm{x}_i$. The activation function here is $\phi=\mathrm{erf}$. 
Specifically, we consider a student network of the form  
\begin{align}
f(\bm{x}_{\mu}) &= \sum_{i=1}^N \sum_{c=1}^{C} a_{ic} \erf\left(\bm{w}_c \cdot \bm{x}_{\mu,i}\right). 
\end{align} 
requiring enough data-points to resolve the target sets $n > N+S$ (number of target parameters) while staying within the over-parameterized regime implies $n < CNS$. We further consider the large-scale ``thermodynamic'' limit, where $C,S,N,n \gg 1$. 

Similarly to the 3 layer case, the equations of states here are given by, 
\begin{align}
[\Sigma^{-1}]_{ss'} &= \frac{S}{\sigma_\text{w}^2} \delta_{ss'} + \frac{1}{C} \sum_{\mu \nu} A_{\mu \nu}^{(2)} \frac{\partial [Q_f]_{\mu \nu}}{\partial \Sigma_{ss'}} \\ \nonumber 
\label{Eq:Qf2Layer}
[Q_f]_{\mu \nu} &= \frac{2 \sigma_\text{a}^2}{\pi N} \sum_i\sin^{-1} \left(\frac{2 \bm{x}_{\mu,i} \Sigma \bm{x}_{\mu,i}}{\sqrt{1+2 \bm{x}_{\mu,i} \Sigma \bm{x}_{\mu,i}}\sqrt{1+2 \bm{x}_{\nu,i} \Sigma \bm{x}_{\nu,i}} }  \right)  \\ \nonumber
{\bm{\delta}} &= [Q_f + \sigma^2 I_n]^{-1} \bm{y} \\ \nonumber 
A_{\mu \nu}^{(2)} &=- {\delta}_{\mu} {\delta}_{\nu} + \left[(Q_f + \sigma^2 I_n)^{-1}\right]_{\mu \nu}
\end{align}
These can be viewed as non-linear equations in the $S(S-1)/2$ variables making up the symmetric matrix $\Sigma$. Having these variables determines the $t_{\mu}$ variables directly.

We begin with approximating the GP inference appearing in the last equation. This can be represented as ${\bm{\delta}}=[\bm{y}-\bar{\bm{f}}]/\sigma^2$ where, $\bar{\bm{f}} = Q_f [Q_f + \sigma^2 I_n]^{-1}\bm{y}$ following the standard GP prediction formula (for the training set). At large $n$, $\bar{\bm{f}}$ can be approximated using the equivalence kernel (EK) approximation \cite{Rasmussen2005} together with its perturbative corrections \cite{Cohen2019,naveh2021predicting} (non-perturbative approaches in $1/n$ could also be considered \cite{bordelon2021,Simon2021}). Within this approximation scheme, one considers $[Q_f]_{\mu \nu}=Q_f(\bm{x}_{\mu},\bm{x}_{\nu})$ as the continuum operator $Q_f(\bm{x},\bm{y})$, diagonalized on the data-set measure ($d\mu$), leading to the following formula for $\bar{f}(\bm{x})$ in the strict EK limit 
\begin{align}
\bar{f}(\bm{x}) &= \sum_{\lambda} \frac{\lambda}{\lambda+\sigma^2/n} y_{\lambda} \phi_{\lambda}(\bm{x}),
\end{align}
where $\lambda,\phi_{\lambda}(\bm{x})$ are the eigenvalues and eigenfunctions of $Q_f$ and $y_{\lambda}=\int d \mu_x y(\bm{x}) \phi_{\lambda}(\bm{x})$. To obtain an explicit formula for, $\bar{t}(\bm{x}) = [y(\bm{x}) - \bar{f}(\bm{x})]/\sigma^2$, we proceed by solving the eigenvalue problem. 
To this end, we first consider the kernel action on a general linear function ($\bm{w'} \cdot \bm{z}_j$), where $\bm{z}_j,\bm{w'}$ are vectors of size $S$ for all $j$, and $\bm{z}_j$ is drawn from the dataset measure, 
\begin{align}\label{eq:OfLinearprojection}
\int d\mu_zQ_f(\bm{x},\bm{z})\bm{w}'\cdot \bm{z}_j =&\int d\mu_z  \frac{\sigma_\text{a}^2}{N}\sum_i \int d\bm{w} p(\bm{w}) \erf(\bm{w} \cdot \bm{x}_i) \erf(\bm{w} \cdot \bm{z}_i) (\bm{w}' \cdot \bm{z}_j). 
\end{align}
where $p(\bm{w})$ is a centered Gaussian with covariance matrix $\Sigma$. In the second transition, we  undo the kernel integral, where $d\mu_z$ is a Gaussian measure.  We then exchange the order of integration and sum over the weights and the data. We now do the integration over the data, 
\begin{align}
\label{AppEq:MFforErf}
\int d\mu_z \erf(\bm{w} \cdot \bm{z}_i) \bm{w}' \cdot \bm{z}_j &= \delta_{ij} \sqrt{\frac{4}{\pi}} \frac{\bm{w}' \cdot \bm{w}}{\sqrt{1+2\norm{\bm{w}}^2}} &\approx  \delta_{ij} \sqrt{\frac{4}{\pi}} \frac{\bm{w}' \cdot \bm{w}}{\sqrt{1+2\langle \norm{\bm{w}}^2 \rangle_{\Sigma}}}
\end{align}
where on the r.h.s. we noted that as $S \gg 1$, $\norm{\bm{w}}^2$ is weakly fluctuating and close to its mean. Next, we perform the $\int d\bm{w} p(\bm{w})$ integral which, following this mean-field replacement, is now of the same type as the previous one. Overall this yields 
\begin{align}
\label{AppEq:EigsForQ}
&\int d\mu_z  \frac{\sigma_\text{a}^2}{N}\sum_i \int d\bm{w} p(\bm{w}) \erf(\bm{w} \cdot \bm{x}_i) \erf(\bm{w} \cdot \bm{z}_i) (\bm{w'} \cdot \bm{z}_j) \\ \nonumber &=  \frac{\sigma_\text{a}^2}{N}\sum_i \sqrt{\frac{4}{\pi}} \frac{1}{\sqrt{1+2\langle \norm{\bm{w}}^2 \rangle_{\Sigma}}} \int d\bm{w} p(\bm{w}) \erf(\bm{w} \cdot \bm{x}_j) (\bm{w}' \cdot \bm{w}) \\ \nonumber
&= \frac{\sigma_\text{a}^2}{N} \sqrt{\frac{4}{\pi}} \frac{1}{\sqrt{1+2\langle \norm{\bm{w}}^2 \rangle_{\Sigma}}}  \sqrt{\frac{4}{\pi}} \frac{\bm{w}' \Sigma \bm{x}_j}{\sqrt{1+2\bm{x}^\transpose_j \Sigma \bm{x}_j}}
\end{align}
Using again $S \gg 1$ we replace $2\bm{x}_j^\transpose \Sigma \bm{x}_j$ by its mean under $\int d\mu_x$. Following this, we obtain that the action of $Q_f$ preserves the space of linear function. Furthermore, we see that diagonalizing $Q_f$ in this subspace reduces to diagonalizing $\Sigma$. We thus reach the conclusion that since $y(\bm{x})$ is a linear function, so must be $\bar{f}(\bm{x})$ and hence $\bar{t}(\bm{x})$. 

We turn to the quantity $\sum_{\mu \nu} A_{\mu \nu}^{(2)} \frac{\partial [Q_f]_{\mu \nu}}{\partial \Sigma_{ss'}}$ appearing in the equation of state for $\Sigma$. For the moment, we omit fluctuation piece of $A_{\mu \nu}^{(2)}$ and replace it by $-{\delta}_{\mu}{\delta}_{\nu}$. Below, we will show that the contribution of this fluctuation term is negligible at large, $n$ which is our focus here. Following this, we approximate the two summations with the following expression   
\begin{align}
\label{AppEq:ttQ2}
\sum_{\mu \nu} \delta_{\nu} \delta_{\mu} [Q_f]_{\mu \nu} &\approx n^2 \int \int d\mu_x d \mu_y \delta(\bm{x}) \delta(\bm{y}) Q_f(\bm{x},\bm{y}) 
\end{align}

Next, we argue that ${\delta}(\bm{x}) = \sum_i b_i (\bm{w}^* \cdot \bm{x}_i)$ at large $n$, where $b_i$'s are some real numbers. Indeed, at large $n$ when replacing summations by integrals and all matrices by their continuum kernels, the full symmetry of the measure from which $\bm{x}_{\mu}$'s are drawn becomes manifest in the equations. The latter amounts to an independent orthogonal rotation ($O(S)$)  of each, $\bm{x}_i$ which leaves $\bm{w}^*$ invariant. Recalling the previous result, that $\bar{f}(\bm{x})$ is linear, together with this symmetry, implies that $\bar{f}(\bm{x})=\sum_i c_i (\bm{w}^* \cdot \bm{x}_i)$ where $c_i$'s are some real numbers. Consequently, $\delta(\bm{x}) = \sigma^{-2}[y(\bm{x})-\bar{f}(\bm{x})]$ is of the same form. 

Using the above ansatz for, $\bar{t}(\bm{x})$ we can solve for the r.h.s. of Eq. (\ref{AppEq:ttQ2}). To this end, we again rewrite the r.h.s. by undoing the kernel integral and exchanging the order of the integration and the sum,
\begin{align}
\label{AppEq:MeanFieldView}
&n^2 \int \int d\mu_x d \mu_y \delta(\bm{x}) \delta(\bm{y}) \frac{\sigma_\text{a}^2}{N}\sum_i \int d\bm{w} p(\bm{w}) \erf(\bm{w} \cdot \bm{x}_i) \erf(\bm{w} \cdot \bm{y}_i) \\ \nonumber &= n^2 \frac{\sigma_\text{a}^2 \norm{\bm{b}}^2}{N}  \frac{4}{\pi} \int d\bm{w} p(\bm{w}) \frac{(\bm{w}^* \cdot \bm{w})^2}{1+2\norm{\bm{w}}^2} 
\end{align}
where we used the fact that
\begin{align}
\left[\int d\mu_x \delta(\bm{x}) \erf(\bm{w} \cdot \bm{x}_i)\right]^2 &= \frac{4}{\pi} \frac{(\bm{u}_i \cdot \bm{w})^2}{1+2\norm{\bm{w}}^2},
\end{align}
where $\bm{u}_i=b_i\bm{w}^*$. 

Next we use again the fact that $\norm{\bm{w}}^2$ is weakly fluctuating at large $S$ to perform the remaining integration over $\bm{w}$ and obtain \begin{align}
n^2 \frac{\sigma_\text{a}^2 \norm{\bm{b}}^2}{N}  \frac{4}{\pi} \frac{(\bm{w}^*)^{\transpose} \Sigma \bm{w}^*}{1+2\langle \norm{\bm{w}}^2 \rangle_\Sigma} 
\end{align}
Here one can also see a different justification for the VGA underlying our equations of state. Indeed, Eq. (\ref{AppEq:ttQ2}) is exactly the non-linear term in the mean-field-decoupled-action for the input layer. At large, $n$ it leads to Eq. \ref{AppEq:MeanFieldView} where replacing $\norm{\bm{w}}^2$ by its expectation value makes the term quadratic. As the first term in that action is quadratic in, $\bm{w}$ the overall action becomes Gaussian, as the VGA assumes. 

Following the above simplification, the equation for $\Sigma$ becomes 
\begin{align}
\label{AppEq:SigmaInv}
\Sigma^{-1} &= \frac{S}{\sigma_\text{w}^2} I_{S} - \frac{4\norm{\bm{b}}^2 n^2 \sigma_\text{a}^2}{CN(1+2\Tr[\Sigma])\pi} \bm{w}^* (\bm{w}^*)^{\transpose}
\end{align}
revealing that only the eigenvector along $\bm{w}^{*}$ is affected by training. At large, $S$ we may thus replace $\Tr[\Sigma]$ by $\sigma_\text{w}^2$ rendering the above an explicit formula for $\Sigma$. 

Next, we return to the eigenvalue equation for $Q_f$ (Eq. \ref{AppEq:EigsForQ}) and use the fact that $\bm{w}^*$ is an eigenvalue ($l_*$) of $\Sigma$ 
\begin{align}
\int d \mu_z Q_f(\bm{x},\bm{z}) y(\bm{z}) &= \lambda_y y(\bm{x}) \\ \nonumber 
\lambda_y &=  \frac{\sigma_\text{a}^2}{N} \frac{4}{\pi} \frac{1}{1+2 \sigma_\text{w}^2} l_{*} \\ \nonumber 
l_* &= \left[\frac{S}{\sigma_\text{w}^2} - \frac{4 n^2 \sigma_\text{a}^2}{CN(1+2\sigma_\text{w}^2)\pi} \norm{\bm{b}}^2\norm{\bm{w}^*}^2\right]^{-1}
\end{align}
Taking next the EK predictions along with its leading correction yields 
\begin{align}
\label{AppEq:qFacGP}
\bar{f}(\bm{x}) &= q_{\mathrm{train}} \frac{\lambda_y}{\lambda_y + \sigma^2/n} y(\bm{x})
\end{align}
where $q_{\mathrm{train}}$ is $1$ at the strict equivalence kernel limit. Taking the leading order correction as in Refs.  \cite{Cohen2019,naveh2021self} one finds 
\begin{align}
q_{\mathrm{train}} &= \frac{(1-\alpha_{\mathrm{EK}}(1-\sigma^{-2}\bar{C}_n \alpha_{\mathrm{EK}}))}{(1-\alpha_{\mathrm{EK}})} \\ \nonumber
\alpha_{\mathrm{EK}} &= \frac{\sigma^2/n}{\lambda_{y}+\sigma^2/n} 
\end{align}
where $\bar{C}_n$ is the posterior covariance in the EK limit given by $\sum_{\lambda} \frac{1}{\lambda^{-1} + n/\sigma^2}$, where $\lambda$ are $Q_f(\bm{x},\bm{y})$'s eigenvalues in the $C\rightarrow \infty$ limit. In practice, we estimated $\bar{C}_n$ numerically by diagonalizing large kernels. For $N,S=20,64$ found $\bar{C}_{800}=0.13$ and $\bar{C}_{1600}=0.078$.

Notably since $\delta(\bm{x})$ came out proportional to $y(\bm{x})$ we obtained a simplified form for $\delta(\bm{x})$ containing only a single free parameter ($\alpha$) 
\begin{align}
\delta(\bm{x}) &= \alpha y(\bm{x})
\end{align}
In addition we may now replace all the $\norm{\bm{b}}^2$ factor appearing above with $\alpha^2\norm{\bm{a}^*}^2$.

Altogether, this yields the following non-linear equation for the scalar quantity $\alpha$ 
\begin{align}
\label{Eq:AlphaNonLin} 
\sigma^2 \alpha &= 1 - \frac{q_{\mathrm{train}} \lambda_{\infty} \left[1 -  \frac{\norm{\bm{a}^*}^2 \norm{\bm{w}^*}^2\sigma_\text{w}^2 \sigma_\text{a}^2(n \alpha)^2}{NSC} \frac{4}{(1+2\sigma_\text{w}^2)\pi}\right]^{-1}}{\lambda_{\infty} \left[1 -  \frac{\norm{\bm{a}^*}^2 \norm{\bm{w}^*}^2 \sigma_\text{w}^2 \sigma_\text{a}^2(n \alpha)^2}{NSC} \frac{4}{(1+2\sigma_\text{w}^2)\pi}\right]^{-1} + \sigma^2/n} \\ \nonumber 
\lambda_{\infty} &= \frac{4 \sigma_\text{a}^2 \sigma_\text{w}^2}{\pi (1+2 \sigma_\text{w}^2) NS}
\end{align}
Solving the above equation for $\alpha$, one obtains $\Sigma$ and $Q_f$ using the equations of state. Using, $Q_f$ we can calculate the DNNs predictions on the test set using standard GP inference. We note by passing that one can also estimate the result of this GP inference using EK. However, we found that just keeping a leading perturbative correction to the EK result, as done for the training set, resulted in $20-30\%$ discrepancies when compared to exact the GP inference formula. As estimating GP inference was not a main focus of the current work, we instead opted to perform this last GP inference on the test set numerically. In principle, other analytical methods for estimating GP inference could be used here \cite{bordelon2020spectrum}. Finally, we note that when estimating $\alpha_{\text{train}}$ and $\alpha_{\text{test}}$ on a specific dataset they are defined as 
\begin{align}
\alpha_{\text{train}} &= \frac{\sum_{\mu \in {\mathrm{Train}}}{(y_{\mu}-f_{\mu})y_{\mu}}}{\sum_{\mu \in {\mathrm {Train}}} y_{\mu}^2} \\
\alpha_{\mathrm{test}} &= \frac{\sum_{\mu \in {\mathrm{Test}}}{(y_{\mu}-f_{\mu})y_{\mu}}}{\sum_{\mu \in {\mathrm{Test}}} y_{\mu}^2}
\end{align}
where ${\it{Train}}$ and ${\it{Test}}$ refer to samples taken from the train and test datasets, respectively. 

Last we turn to discuss the fluctuation term we omitted given by 
\begin{align}
C^{-1} \sum_{\mu \nu} \left[Q_f + \sigma^2 I_n \right]^{-1}_{\mu \nu} \frac{\partial [Q_f]_{\mu \nu}}{\partial \Sigma_{ss'}}
\end{align}
We wish to compare its contribution to that of $C^{-1}\Tr[{\bm{\delta}} {\bm{\delta}}^{\transpose}\frac{\partial [Q_f]_{\mu \nu}}{\partial \Sigma_{ss'}}]$. To this end, we write $Q_f$ in terms of its eigenvectors ($\bm{v}_k$) and eigenvalues ($\lambda_k$)  
\begin{align}
\label{AppEq:Expand}
C^{-1} \left[Q_f + \sigma^2 I_n \right]^{-1} &=
C^{-1} \sum_k \frac{\bm{v}_k \bm{v}_k^{\transpose}}{\lambda_k + \sigma^2} 
\end{align}
aiming for an order of magnitude estimation, we perform the following two approximations: First, we approximate the eigenvalue by the leading $n$ eigenvalues of the continuum kernel time $n$. Second, we take this continuum kernel to be GP kernel. The latter is justified by the fact that the feature-learning effects we found are large, but still do not correspond to an order of magnitude change. Following this, we obtain $NS$ degenerate eigenvalues equal to $n \lambda_{\infty}$ and the corresponding $\bm{v}_k$'s span all linear functions on input space (sampled on the training set). 

Next, we estimate how adding such terms to the previous computation affects the equation for $\Sigma^{-1}$. First, we note that $n \lambda_{\infty} \sim n/(NS) \sim 1$, in our two experiments. Next, we imagine repeating the computation of the previous section, with these extra terms corresponding to the various different $\bm{v}_k \bm{v}_k^{\transpose}$. Notably each such term would enter the computation in the same exact manner to $\bar{\bm{t}}$ (see Eq. \ref{AppEq:ttQ2}) namely  
\begin{align}
\label{AppEq:vkvk}
\sum_{\mu \nu} [\bm{v}_k]_{\nu} [\bm{v}_k]_{\mu} [Q_f]_{\mu \nu} &\approx n \int \int d\mu_x d \mu_y v_k(\bm{x}) v_k(\bm{y}) Q_f(\bm{x},\bm{y})
\end{align}
the only two differences are that (i) $||\bm{t}||^2=\alpha^2 n$ whereas $|| [\bm{v}_k]||^2 = 1$ (hence a factor $n$ on the r.h.s. was lost compared to Eq. \ref{AppEq:ttQ2}) and (ii) $\bm{v}_k$ can have any dependence on $\bm{x}_i$ and not only through $\bm{w}^* \cdot \bm{x}_i$. For concreteness, let us span the continuum version of these $\bm{v}_{k}$ by $[\bm{x}_i]_s$. Summed together, all these $NS$ eigenvalues will end up augmenting the r.h.s. Eq. \ref{AppEq:SigmaInv} into 
\begin{align}
\label{AppEq:SigmaInv2}
\Sigma^{-1} &= \frac{S}{\sigma_\text{w}^2} I_{S} - \frac{4\norm{\bm{b}}^2 n^2 \sigma_\text{a}^2}{CN(1+2\Tr[\Sigma])\pi} \bm{w}^* (\bm{w}^*)^{\transpose} +\frac{1}{n \lambda_{\infty} + \sigma^2}\frac{4 n  \sigma_\text{a}^2}{C(1+2\Tr[\Sigma])\pi} I_{S} 
\end{align}
comparing the first and last term on the r.h.s we find it is negligible for $CS (n \lambda_{\infty}+\sigma^2) \gg n$. Notably, even for our $n=800,S=64$ experiment at $C=80$, we find this last term is negligible.  

Last, we note that we solved the equations of states numerically both with and without the extra fluctuation ($[Q_f + \sigma^2 I]^{-1}$) term and found a negligible difference. For instance, $\alpha_{\mathrm{train}}$ for the $n=800$ experiment with $C=80$ came out to be $0.384$ ($0.377$) without (with) the fluctuation term. Similarly, for, $n=1600,C=640$ we found $0.375$ ($0.370$) without (with) the fluctuation term.

\subsection{Continuum Limit of Summations for CNNs and Fully-Connected DNNs}
Here, we argue that the replacement involved in Eq. (\ref{AppEq:ttQ2}) is valid for $n \gg \sqrt{N}S$. We further comment on some implications this has for the fully-connected case ($N=1$).
To show this, we consider a summation of the form 
\begin{align}
\frac{1}{n^2}\sum_{\mu \nu }\delta_\mu \delta_\nu [Q_{f}]_{\mu \nu} =\frac{1}{n^2}\sum_{\mu\nu, ii'} (\bm{u}_i \cdot [\bm{x}_{\mu}]_i)(\bm{u}_{i'} \cdot [\bm{x}_{\nu}]_{i'}) [Q_{f}]_{\mu \nu} 
\end{align}
undoing the kernel integral as we have done before (see for example Eq. (\ref{eq:OfLinearprojection})) yields 
\begin{align}
\sum_{\mu\nu}\frac{\sigma_\text{a}^2}{Nn^2} \sum_{jii'} \int d\bm{w} p(\bm{w})(\bm{u}_i \cdot [\bm{x}_{\mu}]_i)(\bm{u}_{i'} \cdot [\bm{x}_{\nu}]_{i'}) \erf(\bm{w}\cdot [\bm{x}_{\mu}]_j) \erf(\bm{w}\cdot [\bm{x}_{\nu}]_j)  
\end{align}
For simplicity, we present the analysis for $N=1$, and report the results for general $N$ by symmetry.  Re-focusing on the relevant summation,
\begin{align}
I &= \frac{1}{n^2}\sum_{\mu\nu } (\bm{u} \cdot \bm{x}_{\mu})(\bm{u} \cdot \bm{x}_{\nu}) \erf(\bm{w}\cdot \bm{x}_{\mu})\erf(\bm{w}\cdot \bm{x}_{\nu}),  
\end{align}
we consider the average (denoted below by a bar)  and variance of $I$ over the dataset, $\{\bm{x}_{\mu}\}_{\mu=1}^n$ where each sample is drawn from the measure $d\mu_x$, conditioning on the values of $\bm{w}$ and $\bm{u}$. This yield,
\begin{align}
\bar{I} &= \frac{1}{n^2}\sum_{\mu\nu}\mathbb{E}\left[ (\bm{u} \cdot \bm{x}_{\mu})(\bm{u} \cdot \bm{x}_{\nu}) \erf(\bm{w}\cdot \bm{x}_{\mu})\erf(\bm{w}\cdot \bm{x}_{\nu})  \right] 
\\ \nonumber
\text{Var}(I) &= \frac{1}{n^4}\mathbb{E}\left[ \left(\sum_{\mu}(\bm{u} \cdot \bm{x}_{\mu}) \erf(\bm{w}\cdot \bm{x}_{\mu}) \right)^4 \right]- {\bar{I}}^2.
\end{align}
To estimate the scale of these quantities, we focus simplicity on the regime where the error function is linear. This can be generalized by taking into account perturbative corrections, but these do not change the scale. Following this approximation and taking into account centered Gaussian i.i.d. measure with variance $1$ for, $d\mu_x$ one finds, 
\begin{align}
\bar{I} &= (\bm{u}\cdot \bm{w})^2+  \frac{1}{n}\left[(\bm{u} \cdot \bm{w})^2 + \norm{\bm{u}}^2\norm{\bm{w}}^2\right]
\\ \nonumber
\mathrm{Var}(I)& = \frac{3}{n^2}\norm{\bm{u}}^4\norm{\bm{w}}^4 + O((n^2S)^{-1})
\end{align}
Since $\bm{u}$ and $\bm{w}$ are high- dimensional vectors where $\bm{w}\sim\mathcal{N}(0,I_S\sigma^2/S)$ and $\bm{u}$ can be taken to be fixed with $O(1)$ norm in our setting. Therefore, the norm  of $\bm{w}$ concentrates on its average value, $\sigma_\text{w}^2$, which is of order one, with fluctuation of order $S^{-1/2}$. Hence, the variance fluctuation are of order $1/n^2$ and mean fluctuation are of order $\max(1/{S},1/n)$. Thus, $n \gg {S}$ is required for replacing the summation by an integral for $N=1$ (the fully connected case). Similar analysis can be done for general $N$ leading to $Var(I)=O(1/(Nn^2))$, and $\bar{I}=O(1/(SN))$ which then requires  $n \gg {\sqrt{N}S}$, where we assumed, without loss of generality, that $\bm{u}_j = a^*_j\bm{w}^*$ and $\norm{\bm{a}^*}^2=1$, as in the previous section. For our CNN model $N=20,S=64$ and $n=800,1600$ hence this approximation is reasonable.

Let us consider the implications this has on the fully connected case. In the regime where $n \gg S$, the behavior of the MSE will change drastically compared to $N \gg 1$. Indeed, as our previous results show, for our CNN experiments, the train MSE (over $\sigma^4$) reaches $0.4^2$ of the corresponding at $n=1600 \gg S$, and for the GP this quantity is order $1$. Thus, while being small, this MSE is far from negligible. Specifically, in our CNN experiments, the emergent scale ($\alpha^2 n^2/(CNS)$) is order $1$. 

In contrast, at $N=1$, much like in experiments \cite{lee2017deep}, the self-consistent equation predicts a negligible GP-DNN performance gap down to $C$ or order $1$. Moreover, for $n \gg S$ the GP (which has a uniform prior over all the $S$ possible linear functions) actually performs very well. Specifically, the EK approximation yields a train MSE (over $\sigma^4$) of order $10^{-5}$. The emergent scale ($\alpha^2 n^2/(CS)$) is of the same order at $1/C$. 

Several conclusions could be drawn here: (i) Taking $N=1$ in our experiments, there is no separation of scales between the emergent scale and $1/C$. (ii) In a related manner, no appreciable label/target-aware feature learning will take place down to the scale where our inter-layer mean-field breaks down ($C \sim 1$).

\section{Validity of the Variational Gaussian Approximation} 
\label{Sec:validityVGA}
Here, we provide some analytical support to the validity of the Gaussian variational approximation, used to obtain the equation of state. We present the variational treatment from a perturbation theory approach, as a partial summation of a subset of all perturbative corrections. We identify a qualitative difference between this subset and other perturbative corrections that we neglect.
We apply our analysis to one of the typical hidden layers in the mean-field limit. This is easily generalized to all layers due to the recursive structure of the problem. 

In Eq. (\ref{eq:pi_MF}), we introduce the mean-field probability distribution:
\begin{equation}
\pi^{(l)}\propto\prod_{c}e^{-\cS^{(l)}_{c,\mathrm{MF}}}
\label{appEq: prob prod over c}    
\end{equation}
where here we separate the action into two parts $\cS^{(l)}_{c,\mathrm{MF}} = \cS^{(l)}_{0c,\mathrm{MF}}+\frac{1}{C}\Delta \cS^{(l)}_{c,\mathrm{MF}}$. Here we denote by $\cS^{(l)}_{0c,\mathrm{MF}}=\frac{1}{2}\left(\boldsymbol{h}_{c}^{(l)}\right)^{\transpose}\left[Q^{(l)}\right]^{-1}\boldsymbol{h}_{c}^{(l)}$,
the action of the free theory and by $\Delta \cS^{(l)}_{c,\mathrm{MF}}=-\frac{1}{2}\frac{\sigma_{l}^{2}}{N_{l-1}}\sum_{\mu\nu}A_{\mu\nu}^{(l+1)}\phi\left(h_{c\mu}^{(l)}\right)\phi\left(h_{c\nu}^{(l)}\right)$
where $A^{(l+1)}=\langle{\boldsymbol{m}^{(l+1)}\left(\boldsymbol{m}^{(l+1)}\right)^{\transpose}}\rangle_\mathrm{MF}$
the action coming from the interaction between layers. $\sigma_l^2$ is the variance of the weights of the $l$th layer. Since all the channels are independent, in the following, we drop the channel index ($c$). For compactness of notation, we also drop the index of the layer ($l$) and the indices $\mu$ and $\nu$
represents the multi-index over the strides and the training data samples. In the variational Gaussian approximation,
we replace this probability with a Gaussian one with covariance matrix $K$ which minimizes the KL divergence between the two distributions. The  KL divergence between the two distributions is as follows: 
\begin{multline}
D_{KL}(\mathcal{N}(0,K) || \pi) = -\langle\frac{1}{2}\boldsymbol{h}^{\transpose}Q^{-1}\boldsymbol{h}\rangle_{K} \\ + \langle\frac{1}{2}\frac{\sigma_{l}^{2}}{CN}\sum_{\mu\nu}A_{\mu\nu}\left[\phi\left(h_{c\mu}\right)\phi\left(h_{c\nu}\right)\right]\rangle_{K} + \frac{1}{2}\log\det(K) + C.
\end{multline}
where $\langle ... \rangle_K$ is averaging with respect to the Gaussian measures induced by the kernel $K$. For the layer below, This yields the following self-consistent equation for the \textit{post-kernel}, $Q$ and the \textit{pre-kernel}, $K$:
\begin{equation}
Q^{-1}-\frac{1}{C}\sum_{\mu\nu}A_{\mu\nu}\left[\partial_{K}Q\right]_{\mu\nu}=K^{-1}\label{eq:div D for_K-1}.
\end{equation}
We now show in what sense this approximation is valid, i.e. when can we approximate the mean-field distribution by a Gaussian distribution.
We start by calculating the interacting Green function of the process
(second moment of the process). 

\begin{align}
\langle h_{\mu}h_{\nu}\rangle = \langle h_{\mu}h_{\nu}\rangle_{0} + \sum^{\infty}_{j=1} \frac{1}{C^{j}j!} \langle h_{\mu}h_{\nu}\left(\Delta \cS\right)^{j}\rangle_{0}\label{eq:perK},
\end{align}
where $\langle\ldots\rangle$ is the connected expectation with respect
to $\cS_{\text{MF}}$ and $\langle\ldots\rangle_{0}$ is the connected expectation
with respect to $\cS_{0,\text{MF}}.$ Here, connected mean cumulant moments w.r.t the variables $hh$ and $\Delta \cS_{\text{MF}}$.
We perform our perturbation analysis, for simplicity, for the monomial activation function
$\phi(x)=x^{k}$, with $k$ finite, as a characteristic example. This can be generalized to other smooth activation functions. The perturbative correction term, to the free moment, $\langle h_{\mu}h_{\nu}\left(\Delta S_{\text{MF}}\right)^{j}\rangle_{0}$ is as follows: 
\begin{multline}
\langle h_{\mu}h_{\nu}\left(\sum_{\alpha_{1}\beta_{1}}A_{\alpha_{1}\beta_{1}}h_{\alpha_{1}}^{k}h_{\beta_{1}}^{k}\right)^{j}\rangle_{0}\\
=\sum_{\alpha_{1}\beta_{1}...\alpha_{j}\beta_{j}}A_{\alpha_{1}\beta_{1}}\ldots A_{\alpha_{j}\beta_{j}}\langle h_{\mu}h_{\nu}h_{\alpha_{1}}^{k}h_{\beta_{1}}^{k}\ldots h_{\alpha_{j}}^{k}h_{\beta_{j}}^{k}\rangle_{0},
\end{multline}
where $\langle h_{\mu}h_{\nu}\rangle_{0}=Q_{\mu\nu}$. In order to
evaluate the above expression, we need to introduce some additional assumption
on the structure of the matrix $Q$. Based on the numerical simulation and since this
matrix represents the correlation between samples in high dimension, we consider the following estimate on the elements of the matrix $Q$,
off-diagonal elements, $Q_{\mu\nu}=O(m^{-1/2})$, and on-diagonal elements,
$Q_{\mu\mu}=O(1)$, for $Q\in\mathbb{R}^{m\times m}$, where for FCN $m=n$ number of samples for the 3-layer CNN $m=nS_1$ where $S_1$ is the size of the stride.
Under this structure of the matrix $Q$, the leading terms in our perturbation expansions for large $m$ are of the following forms (connected diagrams only), other terms will be of higher powers of $m^{-1/2}$, 
\begin{multline}
\sum_{\alpha_{1}\beta_{1}...\alpha_{j}\beta_{j}}k^{2j}A_{\alpha_{1}\beta_{1}}\ldots A_{\alpha_{j}\beta_{j}}\langle h_{\mu}h_{\alpha_{1}}\langle h_{\alpha_{1}}^{k-1}h_{\beta_{1}}^{k-1}\rangle_{0}h_{\beta_{1}}\ldots h_{\alpha_{j}}\langle h_{\alpha_{j}}^{k-1}h_{\beta_{j}}^{k-1}\rangle_{0}h_{\beta_{j}}h_{\nu}\rangle_{0}\\
=\sum_{\alpha_{1}\beta_{1}...\alpha_{j}\beta_{j}}k^{2j}j!A_{\alpha_{1}\beta_{1}}\ldots A_{\alpha_{j}\beta_{j}}\langle h_{\mu}h_{\alpha_{1}}\rangle_{0}\langle h_{\alpha_{1}}^{k-1}h_{\beta_{1}}^{k-1}\rangle_{0}\langle h_{\beta_{1}}h_{\alpha_{2}}\rangle_{0}\ldots\langle h_{\beta_{j-1}}h_{\alpha_{j}}\rangle_{0}\langle h_{\alpha_{j}}^{k-1}h_{\beta_{j}}^{k-1}\rangle_{0}\langle h_{\beta_{j}}h_{\nu}\rangle_{0}\\
=\sum_{\alpha_{1}\beta_{1}...\alpha_{j}\beta_{j}}k^{2j}j!2^{j}Q_{\mu\alpha_{1}}\langle h_{\alpha_{1}}^{k-1}h_{\beta_{1}}^{k-1}\rangle_{0}A_{\alpha_{1}\beta_{1}}Q_{\beta_{1}\alpha_{2}}\ldots Q_{\beta_{j-1}\alpha_{j}}\langle h_{\alpha_{j}}^{k-1}h_{\beta_{j}}^{k-1}\rangle_{0}A_{\alpha_{j}\beta_{j}}Q_{\beta_{j}\nu}\\
=\sum_{\beta_{1}...\beta_{j}}k^{2j}j!\left[QV(k)\right]_{\mu\beta_{1}}\left[QV(k)\right]_{\beta_{1}\beta_{2}}\ldots\left[QV(k)\right]_{\beta_{j-1}\beta_{j}}Q_{\beta_{j}\nu}\\
=\sum_{\beta}k^{2j}j!\left[\left(QV(k)\right)^{j}\right]_{\mu\beta}Q_{\beta\nu},\label{eq:Qv}
\end{multline}
where $k^{2j}$ is due to the choice of $h_{\alpha}$ from each monomial activation function,
the $j!$ is due to the arrangement in pairs of the remaining fields.  We denote by $V(k)_{\alpha_{1}\beta_{1}}=\langle h_{\alpha_{1}}^{k-1}h_{\beta_{1}}^{k-1}\rangle_{0}A_{\alpha_{1}\beta_{1}}$.
Plugging back in Eq. (\ref{eq:perK}) leads to the following self-consistent equation: 
\begin{multline}
K_{\mu\nu}=Q_{\mu\nu}+\sum_{j=1}^{\infty}\sum_{\beta}\left[\left(\frac{2k^{2}}{C}QV(k)\right)^{j}\right]_{\mu\beta}Q_{\beta\nu}+O(m^{-1/2})\\=\sum_{\beta}\left[\left(I_{m}-\frac{2k^{2}}{C}QV(k)\right)^{-1}\right]_{\mu\beta}Q_{\beta\nu}+O(m^{-1/2})
\end{multline}
In the first transition, we sum over the geometric series in the regime, where $\left\Vert QV(k)\right\Vert _{\mathrm{op}}<C/2k^{2}$,
which is valid for large enough $C$, and $m$. We now invert the equation in
order to compare to our mean-field equation (Eq. (\ref{eq:div D for_K-1})): 
\[
K^{-1}=Q^{-1}-\frac{2k^{2}}{C}QV(k)+O(\left(C\sqrt{m}\right)^{-1})
\]
Plugging the definition of the matrix V, we have that, 
\begin{multline*}
\left[K^{-1}\right]_{\mu\nu}=\left[Q^{-1}\right]_{\mu\nu}-\frac{k^{2}}{C}\sum_{\rho}Q_{\mu\rho}\langle\;h_{\rho}^{k-1}h_{\nu}^{k-1}\rangle_{0}A_{\rho\nu}+O(\left(C\sqrt{m}\right)^{-1})\\
=\left[Q^{-1}\right]_{\mu\nu}-\frac{1}{C}\sum_{\rho}\partial_{Q_{\mu\rho}}\langle\;h_{\rho}^{k}h_{\nu}^{k}\rangle_{0}A_{\rho\nu}+O(\left(C\sqrt{m}\right)^{-1})
\end{multline*}
The second transition is using Gaussian integration by parts.
The resulting equation is very similar to the mean-field equation we find.
The difference is that here instead of derivative by $K$, the \textit{pre-kernel}, we have a derivative of the \textit{post-kernel}. In addition, the expectation
is also with respect to the free theory. Indeed, the full variational treatment is self-consistent or, equivalently stated, it takes into account a larger set of diagrams (terms in perturbation theory) which amount to renormalizing the 2-point function from $Q$ to $K$. We argue however that doing so only improves the overall accuracy. Indeed, the expansion of Eq. (\ref{eq:Qv}), is the same as one would get from a Gaussian action consisting of $\cS_{0,\text{MF}}$ plus a quadratic term of the form 
$h_{\alpha_j}h_{\beta_j}k\langle h_{\alpha_{j}}^{k-1}h_{\beta_{j}}^{k-1}\rangle A_{\alpha_{j}\beta_{j}}$.  The variational Gaussian approximation essentially finds the closest Gaussian distribution. Therefore, it can only improve upon this simpler approximation we took here.

\section{Variational Gaussian Approximation for ReLU Activation \label{sec:VGA_ReLU}}
Here, we extend the previous VGA treatment, which assumed centered distributions, to non-centered ones. Indeed, for antisymmetric activation functions, the pre-activations appear schematically as $\bm{h}^{(l+1)} = \bm{v} \phi(\bm{h}^{(l)})$, thus $\bm{v}\phi(\bm{h}^{(l)})=-\bm{v}\phi(-\bm{h}^{(l)})$. Since $\bm{v}$ appears only in quadratic order in the action, we find that $\bm{h}^{(l)}$ is as likely as $-\bm{h}^{(l)}$ and its ensemble average is strictly zero. However, for $\phi=\text{ReLU}$, it is not the case. This requires us to extend the previous treatment by including extra variational parameters for the mean. 

Concretely, let us focus on the VGA for the input layer of 3 layers, CNN. Our VGA for the probability is now defined by the variance of the Gaussian in weight space, $\Sigma_{ss'}$ as well as the mean $\chi_{s}$. Repeating the previous analysis one obtains
\begin{multline}
D(\mathcal{N}(\bm{\chi},\Sigma)|\pi(\bm{w}_c|X_{n};\langle \bm{m}\bm{m}^{\transpose}\rangle_{\mathrm{MF}}))
=-\frac{S_0}{2\sigma_\text{w}^{2}}\left[\mathrm{Tr}\left(\Sigma\right)+\norm{\bm{\chi}}^2\right]\\ \nonumber
-\frac{C_2}{2C_1}\sum_{\mu\nu j_{1}j_{2}}\left(\langle \boldsymbol{m}\boldsymbol{m}^{\transpose}\rangle_{\mathrm{MF}}\right)_{\mu\nu j_{1}j_{2}}\left[Q^{(2)}\right]{}_{\mu\nu j_{1}j_{2}} +\frac{1}{2}\log\det(\Sigma)+\text{Const},
\end{multline}
where $\bm{w}_c\in R^{S_0}$ are the weights of channel, $c$ and $Q^{(2)}_{\mu j_1 \nu j_2}$ is now defined by 
\begin{align}
Q^{(2)}_{\mu j_1 \nu j_2}&=\left \langle \frac{\sigma_\text{v}^{2}}{ S_1C_1}\sum^{S_1}_{ic}\phi\left(\boldsymbol{w}_c\cdot\boldsymbol{x}_{\mu,i+j_{1}S_1}\right)\phi\left(\boldsymbol{w}_c\cdot\boldsymbol{x}_{\nu,i+j_{2}S_1}\right)\right \rangle_{\Sigma,\bm{\chi}},
\end{align}
which one can reduce to a one-dimensional integral following Ref. \cite{Cho}. Obtaining an explicit expression, or potentially a perturbation expansion in $\bm{\chi}\cdot \bm{x}_{\mu,i}$, is left for future work.  

Taking the derivative of the KL-divergence with respect to $\Sigma$ and equating it to zero, one obtains 
\begin{align}
\Sigma^{-1}=\frac{S_0}{\sigma_\text{w}^{2}}I_{S_0}+\frac{C_2}{C_1}\sum_{\mu\nu j_{1}j_{2}}\left(\langle\boldsymbol{m}\boldsymbol{m}^{\transpose}\rangle_{\mathrm{MF}}\right)_{\mu j_1 \nu j_{2}}\left[\partial_{\Sigma}Q^{(2)}\right]{}_{\mu j_{1},\nu j_{2}}.
\end{align} 
Similarly as $\bm{\chi}$ one obtains the additional equation 
\begin{align}
\chi_{\mu j} &= -\frac{C_2 \sigma_\text{w}^2}{2C_1 S_0}\sum_{\mu j_{1},\nu j_{2}}\left(\langle\boldsymbol{m}\boldsymbol{m}^{\transpose}\rangle_{\mathrm{MF}}\right)_{\mu j_{1},\nu j_{2}}\left[\partial_{\chi_{\mu j}}Q^{(2)}\right]{}_{\mu j_{1},\nu j_{2}}
\end{align} 

We turn to obtain the quantity $\langle \boldsymbol{m}\boldsymbol{m}^{\transpose}\rangle_{\mathrm{MF}}$. Repeating the derivation in subsection \ref{subsec:Auxiliary-field-correlation_mmT} noticing that now $\langle {\bm{h}}^{(2)}\rangle_{\mathrm{MF}}$ is non-zero, we obtain 
\begin{align}
\langle \boldsymbol{m}\boldsymbol{m}^{\transpose} \rangle_{\mathrm{MF}} &= Q^{(2)}-[Q^{(2)}]^{-1} \left[K^{(2)}+\bm{\chi}^{(2)} (\bm{\chi}^{(2)})^{\transpose}\right][Q^{(2)}]^{-1}
\end{align}
where $
\chi^{(2)}_{\mu j}$ is the average of $h_{\mu j c'}^{(2)}$ (for any $c'$).

\section{Further details on the numerical experiments}

\subsection{FCN experiments}

\begin{figure}[h!]
\vspace*{-0.1in}
\begin{center}
\includegraphics[height=7cm,trim={0cm 0cm 0cm 0cm},clip]{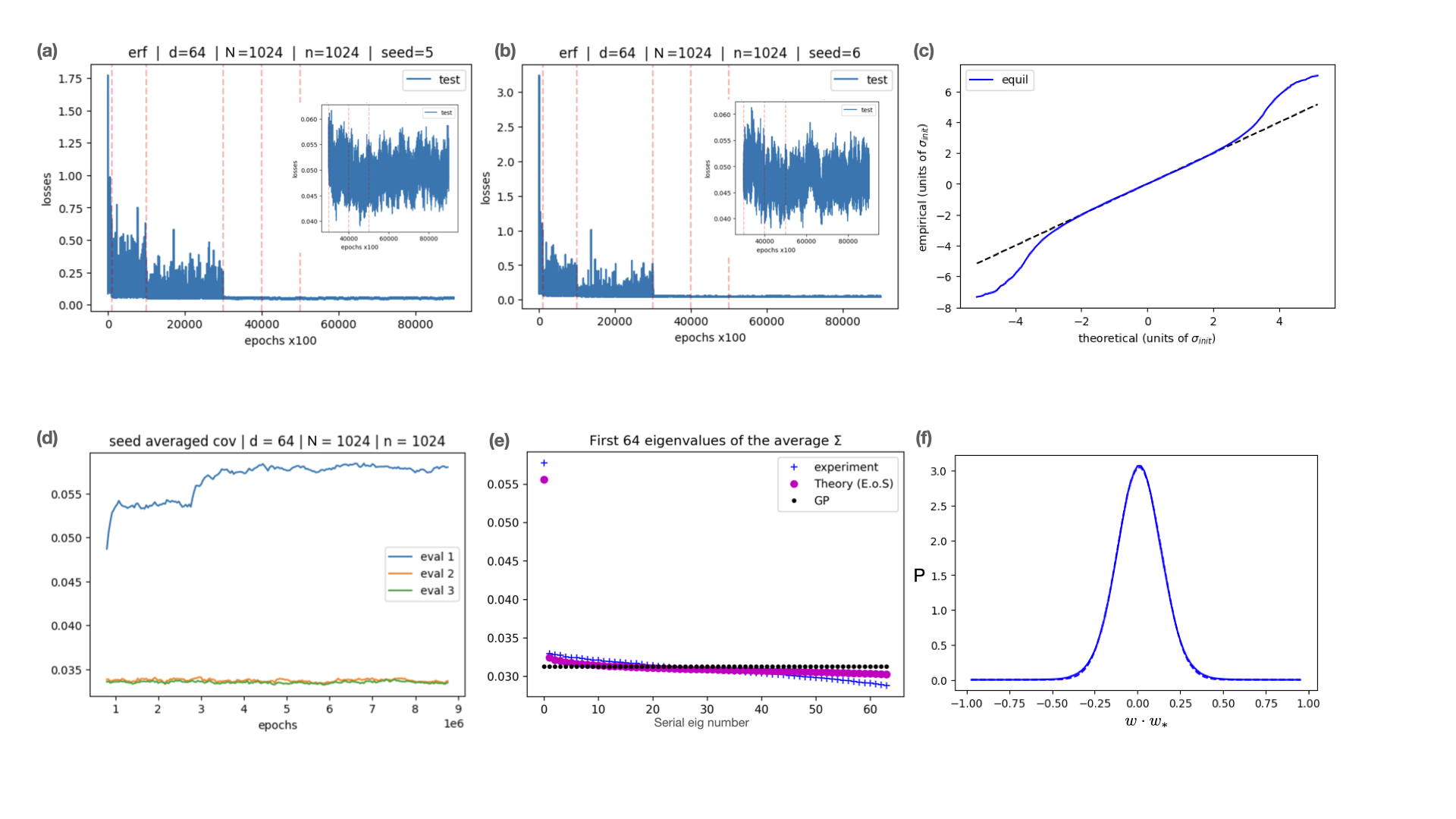}
\end{center}
\caption{\label{fig:fcn_theory_exp64}
{\bf FCN "MF". Theory versus Experiment.} $N_1=N_2=1024,\sigma^2=0.001$, 26 training seeds. Further experimental data on $d=64,n=1024$ FCN with MF scaling is reported in Fig. 1. panel (b) of the main text. Panels ({\bf a},{\bf b}) show the test loss as a function of the number of epochs, with insets focusing on later times. Panels ({\bf c},{\bf f}) study the distribution $P(\bm{w} \cdot {\bm w}^*)$ as a QQ-plot against a Gaussian and as a histogram with the least square fit to a Gaussian (the latter in dashed lines). Panel ({\bf d}) Shows the equilibration $\Sigma$ averaged over the different training-seed, via its 3 leading eigenvalue. Panel ({\bf e}) shows the empirical, GP, and theoretical eigenvalues of the average $\Sigma$. As before, by GP here we mean taking $N_1, N_2 \rightarrow \infty$ at fixed $\sigma_\text{a}^2$. 
}
\end{figure}

Here we report on several additional numerical results for FCNs. In particular, we provide details about the full spectrum of $\Sigma$, the equilibration process, and the Gaussianity measures. We also report on some numerical experiments with FCN in the standard scaling. 
\begin{figure}[h!]
\vspace*{-0.1in}
\begin{center}
\includegraphics[height=7cm,trim={0cm 0cm 0cm 0cm},clip]{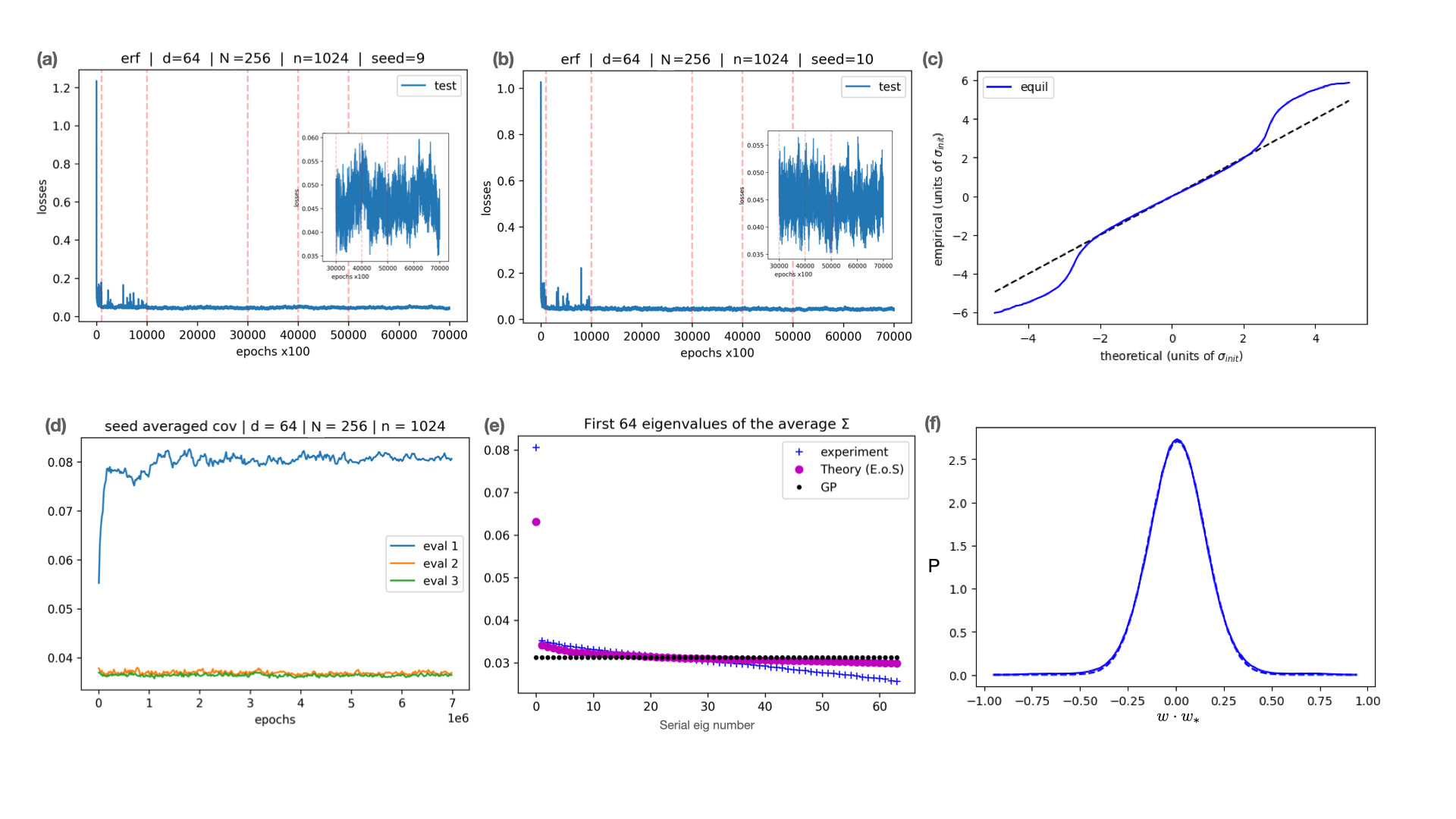}
\end{center}
\caption{ \label{fig:fcn_theory_exp256}{\bf FCN "MF". Theory versus Experiment.} $N_1=N_2=256$. 27 training seeds. $\sigma^2=0.001$. See the previous caption for the panel description.  
}
\end{figure}

We conducted further experiments with the above FCNs ($d=64,n=1024,\sigma^2 = 0.001$, teacher-student with $N_l=1$ for the teacher) however with standard scaling ($\sigma_\text{a}^2=2$, regardless of $N_l$) rather than "MF" scaling ($\sigma_\text{a}^2=2/N_2$). Here, we expect feature learning to diminish \cite{yang2019scaling}. Considering the numerical solution of our EoS, we found that they essentially remain close to the GP limit for $N_1=N_2=64$. Specifically, at $N_1=N_2=64$ leading $\Sigma$ eigenvalue came out $0.03257$, whereas in the GP limit we obtain $2/d=0.03125$. This is consistent with the fact that the emergent scale here (Figure 1. panel (b) main text) is of the order of $1/N_2$. Figure \ref{fig:fcn_theory_exp64} and Figure \ref{fig:fcn_theory_exp256} presents the results for different width $N_l$.

\newpage
\subsection{2-layer CNN experiment}
\begin{figure}[h!]
\vspace*{-0.1in}
\begin{center}
\includegraphics[height=7cm,trim={0cm 0cm 0cm 0cm},clip]{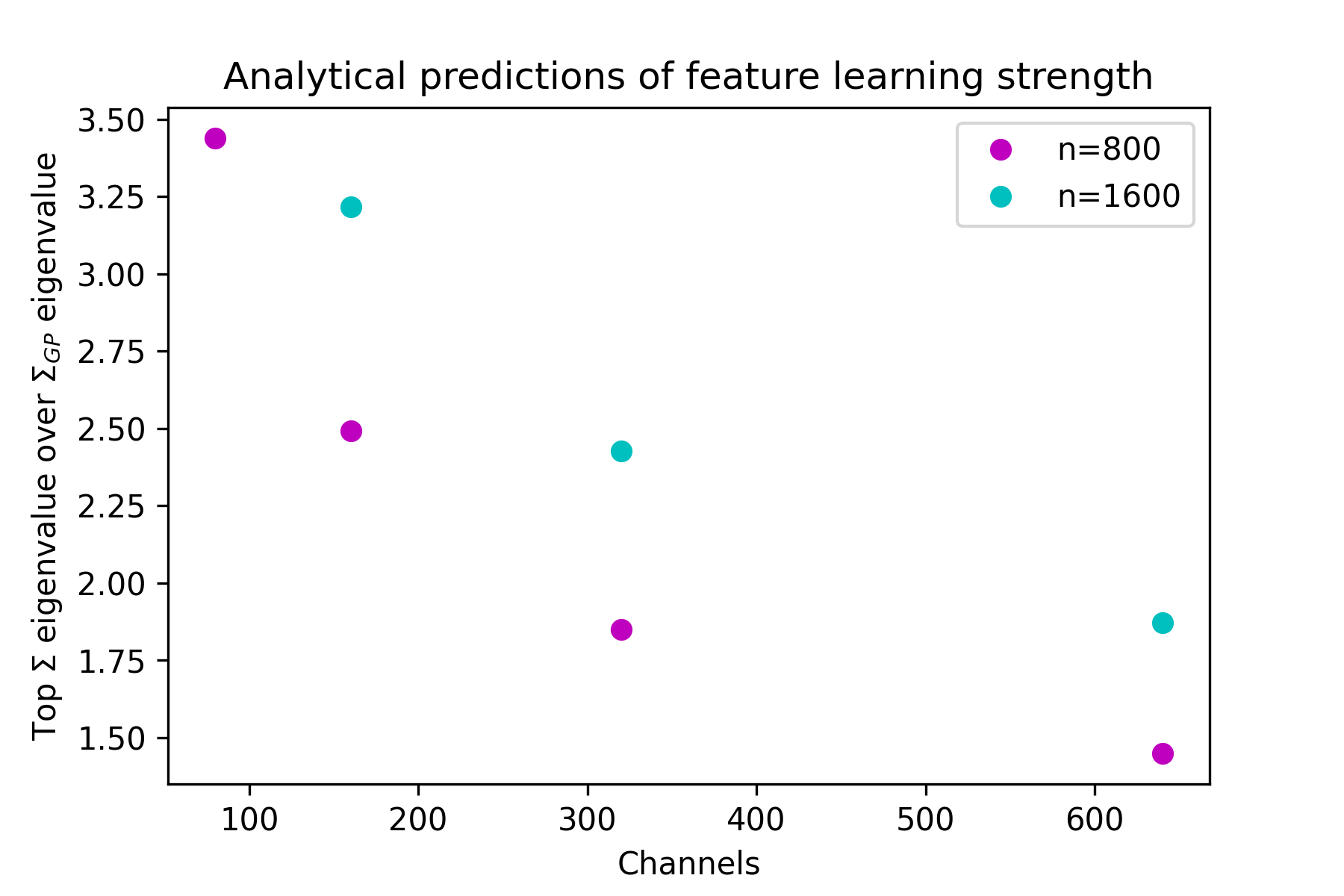}
\end{center}
\caption{ \label{fig:2layerSigmaEig}{\bf 2-Layer CNN, Sigma eigenvalues.} Top $\Sigma$ eigenvalue according to theory over $\sigma_\text{w}^2/d$ which is the (degenerate) $\Sigma$ eigenvalue in the GP limit, for the experiment shown for the 2-layer CNN in the main text. 
}
\end{figure}

Figure \ref{fig:2layerSigmaEig} shows the top $\Sigma$ eigenvalues normalized by $\sigma_\text{w}^2/d$, this is a complementary figure to Figure 1(b) introduced in the main text. Clearly, as the width increase, the eigenvalues of $\Sigma$ are getting closer to the GP kennel eigenvalues. 

\newpage
\subsection{Myrtle-5 CNN on subsets of CIFAR-10 experiment}
\begin{figure}[h!]
\vspace*{-0.1in}
\begin{center}
\includegraphics[width=12cm]{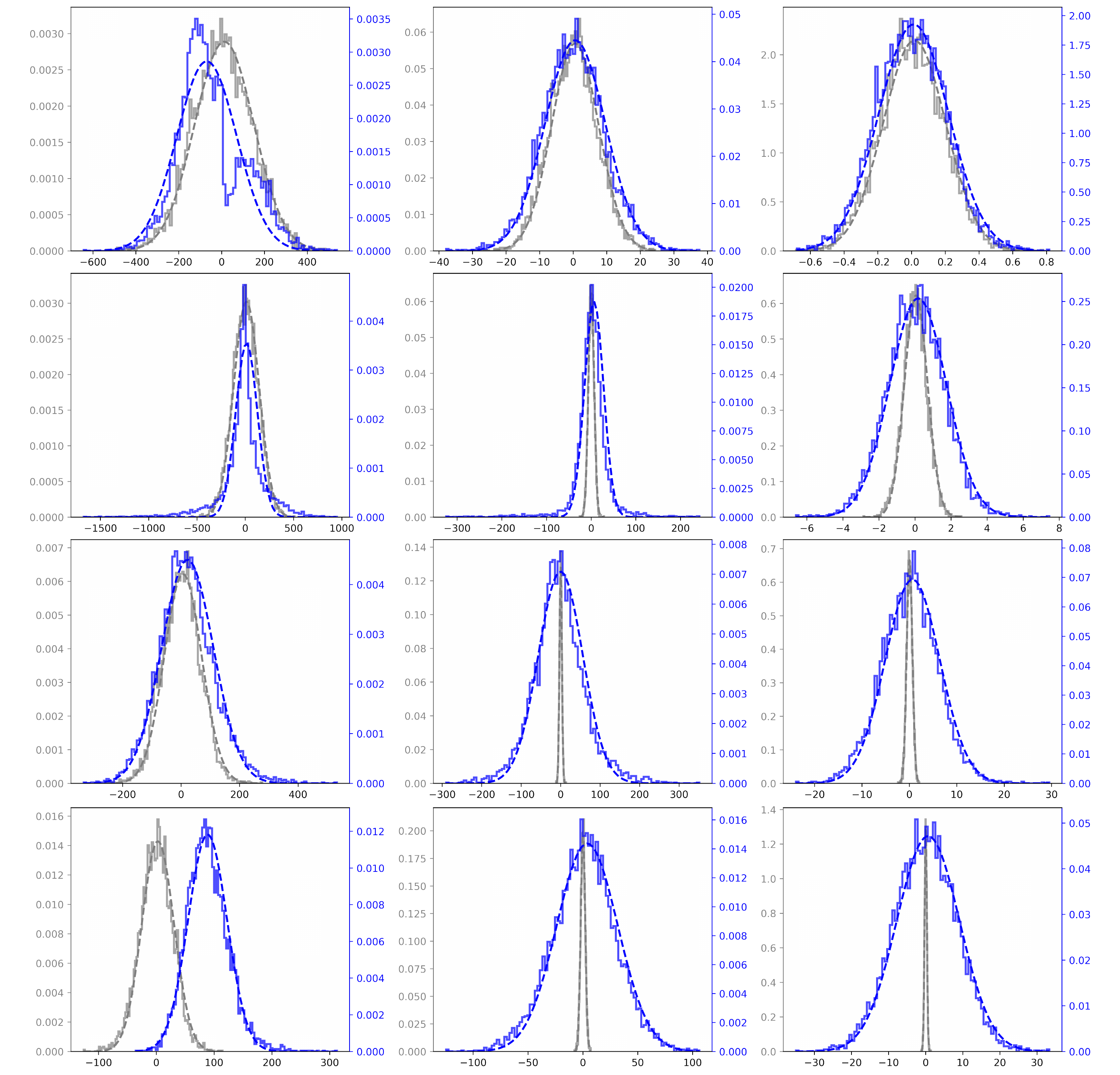}
\end{center}
\caption{ {\bf Pre-activation statistics of Myrtle-5 CNN:} 
Grey and blue lines denote the statistics of untrained nets with random initialization and nets at equilibrium, respectively. Dashed lines are Gaussian fits with non-linear least squares to the empirical distributions. 
Each row corresponds to a specific layer in the network, and each column corresponds to a projection to a different eigenvector of the covariance matrix in Fourier space: the 1st, 3rd and 10th eigenvectors, respectively.
}
\label{suppFig: hists}
\end{figure}

\begin{figure}[h!]
\vspace*{-0.1in}
\begin{center}
\includegraphics[width=12cm]{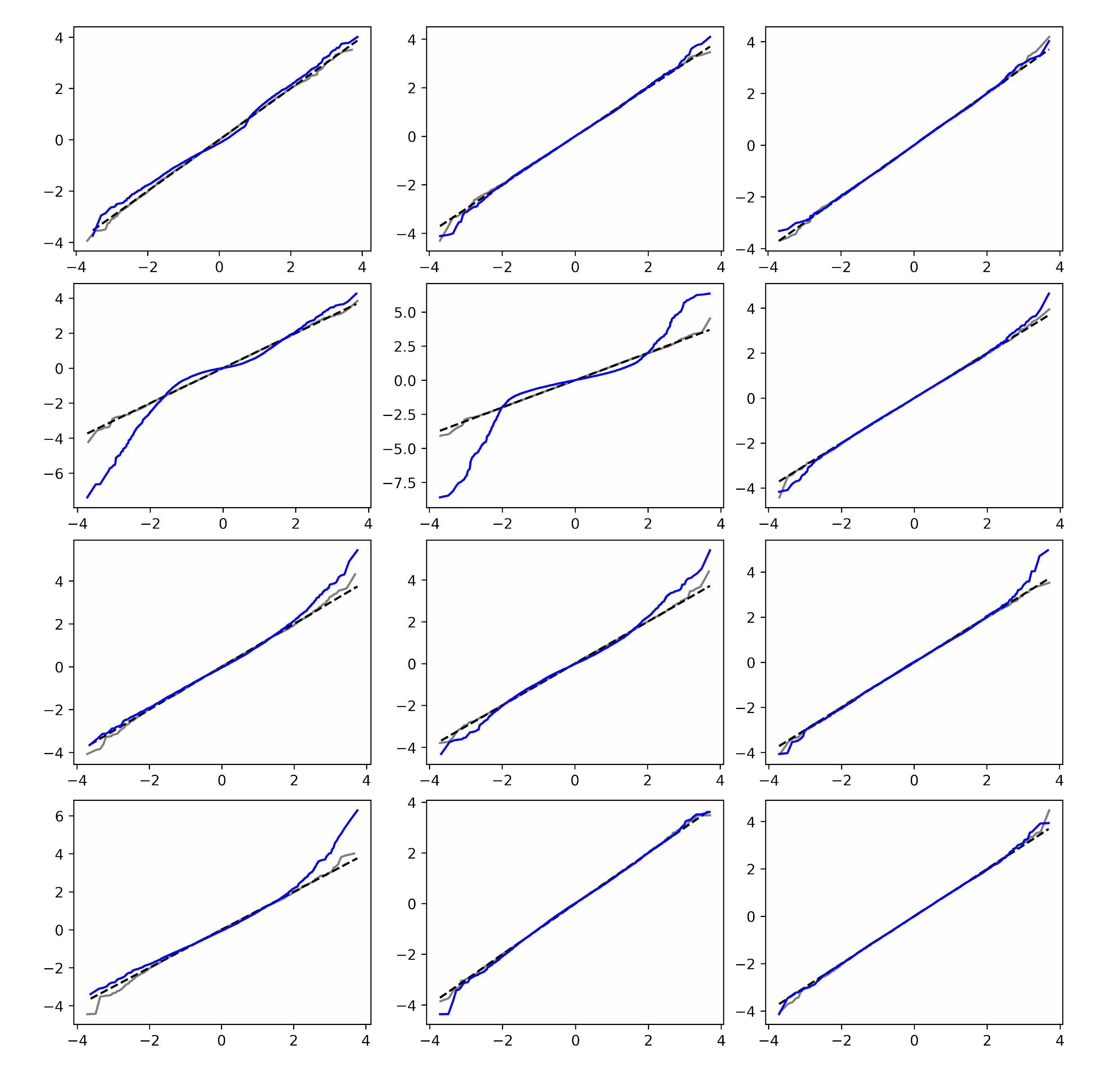}
\end{center}
\caption{ {\bf Pre-activation statistics of Myrtle-5 CNN - QQ plots:} 
Here we show the QQ plots corresponding to the same data appearing in the previous figure. The x-axis corresponds to the best fit Gaussian distribution and the y-axis corresponds to the empirical distribution (both in units of the std at initialization). 
The black dashed line corresponds to the identity line, and the closer the curve is to it the better the fit to a Gaussian. 
Grey and blue lines denote the statistics of untrained nets with random initialization and trained nets at equilibrium, respectively. Dashed lines are Gaussian fits with non-linear least squares to the empirical distributions. 
Each row corresponds to a specific layer in the network, and each column corresponds to a projection to a different eigenvector of the covariance matrix in Fourier space: the 1st, 3rd and 10th eigenvectors, respectively.
}
\label{suppFig: qqplot}
\end{figure}

Here, we report the statistics of the pre-activations in Fourier space similar to Figure 6 in the main text, but now for all layers and projected on the 1st, 3rd and 10th eigenvectors of the Fourier space covariance matrix.
Let us begin by giving some more details on the procedure for deriving these quantities. The pre-activations at some layer is of the form $h_{\mu,c,x,y}$ where $\mu$ is a data point index, $c$ is a channel index, and $x,y$ are pixel coordinates. We choose some wavenumber $k$ and transform these to Fourier space to yield $h^k_{\mu,c}$ which summarizes contributions from all pixels. We then compute the covariance matrix of these $h^k_{\mu,c}$ (an $n \times n$ matrix), averaging across channels and seeds. Finally, we project $h^k_{\mu,c}$ on some eigenvector of the covariance matrix, and these are the quantities whose statistics we report in figures  \ref{suppFig: hists}, \ref{suppFig: qqplot}.  

There are several empirical observations to be made here:
\begin{enumerate}
	\item 
	Gaussianity generally increases as we go deeper into the network from the input to the output. 
	\item 
	Gaussianity generally increases as we project on higher index eigenvectors (going from left to right across the columns). 
	\item 
	Deviations from Gaussianity can appear in several ways: e.g. as multi-modality (e.g. top left panel), or as excessive kurtosis (e.g. 2nd-row left column). 
\end{enumerate}

\newpage

We further report the statistics of the inter-layer and inter-channel correlations in Table \ref{table: inter-layer corr} and Table \ref{table: inter-channel corr}, respectively. 
We report the value of the estimator of correlations together with the error due to finite samples (i.e. mean $\pm$ std of the estimator). 
We see that the correlations across different layers, as well as across different channels within the same layer are mostly on the order of $10^{-3}$ (apart from the inter-channel correlations in layer 4 which are slightly larger). 
The larger error we see for the estimator of the inter-channel correlations is a result of having a smaller sample size since for the inter-layer correlations the number of samples is $n_{\mathrm{seeds}} \cdot C$ while for the inter-channel correlations the number of samples is only $n_{\mathrm{seeds}}$ (where the number of seeds used for this experiment is $n_{\mathrm{seeds}} = 25$, and recall that the number of channels here is $C=256$). 
These empirical findings justify our approximations which neglect inter-channel and inter-layer correlations. 

\begin{table}[h!]
	\centering
	\begin{tabular}{|c| c c c|} 
		\hline
		& layer 2 & layer 3 & layer 4 \\ [0.5ex] 
		\hline\hline
		layer 1    & -9.58e-03 $\pm$ 5.02e-03 & -8.73e-03 $\pm$ 6.77e-03 & 4.51e-03 $\pm$ 8.53e-03  \\
		layer 2 &       & -1.50e-02 $\pm$ 8.94e-03 & -6.85e-03 $\pm$ 6.82e-03 \\
		layer 3 &  &       & -9.27e-05 $\pm$ 1.02e-02 \\ [1ex] 
		\hline
	\end{tabular}
	\caption{Inter-layer Pearson correlations}
	\label{table: inter-layer corr}
\end{table}

\begin{table}[h!]
	\centering
	\begin{tabular}{|c c c c|} 
		\hline
		layer 1 & layer 2 & layer 3 & layer 4 \\ [0.5ex] 
		\hline\hline
		-0.0041 $\pm$ 0.2057 &  0.0079 $\pm$ 0.2057 & -0.0031 $\pm$ 0.2051 & 0.0185 $\pm$ 0.2045 \\ [1ex] 
		
		\hline
	\end{tabular}
	\caption{Inter-channel Pearson correlations}
	\label{table: inter-channel corr}
\end{table}

\bibliography{refs}

\end{document}